\PassOptionsToPackage{prologue,dvipsnames, table}{xcolor}
\documentclass[10pt,twocolumn,letterpaper]{article}

\usepackage[pagenumbers]{wacv} 

\definecolor{wacvblue}{rgb}{0.21,0.49,0.74}
\usepackage[pagebackref,breaklinks,colorlinks,allcolors=wacvblue]{hyperref}

\usepackage{graphicx}
\usepackage{amsmath}
\usepackage{amssymb}
\usepackage{booktabs}
\usepackage{comment}

\usepackage{times}
\usepackage{epsfig}
\usepackage{float}
\usepackage{multirow}
\usepackage{amsfonts}
\usepackage{graphics}
\usepackage{witharrows}
\usepackage{arydshln}
\usepackage{amsfonts} 
\usepackage{mathtools}
\usepackage{enumitem}

\usepackage{algorithm}
\usepackage{algpseudocode}

\newsavebox{\subfloatbox}

\newcommand{\topfloat}[2][\empty]
 {\savebox\subfloatbox{#2}%
  \begin{minipage}[t]{\wd\subfloatbox}
    \usebox\subfloatbox
    \subcaption{#1}
  \end{minipage}}

\newcommand{\bottomfloat}[2][\empty]
 {\savebox\subfloatbox{#2}%
  \begin{minipage}[b]{\wd\subfloatbox}
    \captionsetup{position=top}%
    \subcaption{#1}
    \usebox\subfloatbox
  \end{minipage}}

\usepackage[pagebackref,breaklinks,colorlinks]{hyperref}

\usepackage[capitalize]{cleveref}
\crefname{section}{Sec.}{Secs.}
\Crefname{section}{Section}{Sections}
\Crefname{table}{Table}{Tables}
\crefname{table}{Tab.}{Tabs.}

\newcommand{\fref}[1]{Figure \ref{#1}}
\newcommand{\tref}[1]{Table \ref{#1}}

\def\wacvPaperID{1549} 
\def\confName{WACV}
\def\confYear{2026}

\title{WiSE-OD: Benchmarking Robustness in Infrared Object Detection}

\author{Heitor R. Medeiros \qquad Atif Belal \qquad Masih Aminbeidokhti \\ Eric Granger \qquad Marco Pedersoli \\
Laboratoire d’imagerie, de vision et d’intelligence artificielle (LIVIA) \\
International Laboratory on Learning Systems (ILLS) \\
Dept. of Systems Engineering, ETS Montreal, Canada}

\begin{document}

\twocolumn[{%
\renewcommand\twocolumn[1][]{#1}%
\maketitle
\begin{center}
    \centering
    \captionsetup{type=figure}
    \label{figure:initial_fig}
    \begin{subfigure}[!htp]{0.22\textwidth}
        \caption{GT}
    
        \makebox[0pt][r]{\makebox[15pt]{\raisebox{50pt}{\rotatebox[origin=c]{90}{LLVIP-C}}}}%
        \includegraphics[width=\textwidth]{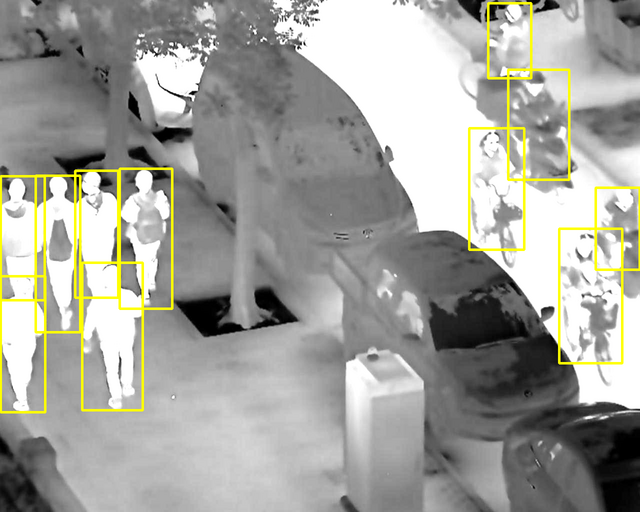}

        \makebox[0pt][r]{\makebox[15pt]{\raisebox{50pt}{\rotatebox[origin=c]{90}{FLIR-C}}}}%
        \includegraphics[width=\textwidth]{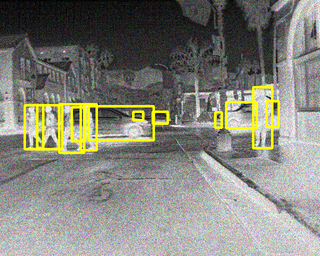}


    \end{subfigure}
    \begin{subfigure}[!htp]{0.22\textwidth}
        \caption{Zero-Shot}
    
        \includegraphics[width=\textwidth]{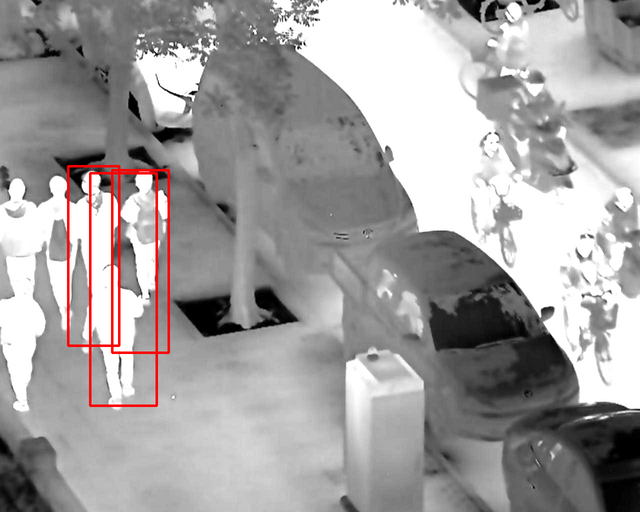}
    
        \includegraphics[width=\textwidth]{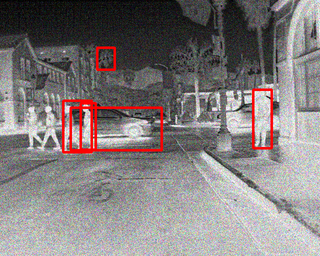}

    \end{subfigure}
    \begin{subfigure}[!htp]{0.22\textwidth}
        \caption{FT}
        
        \includegraphics[width=\textwidth]{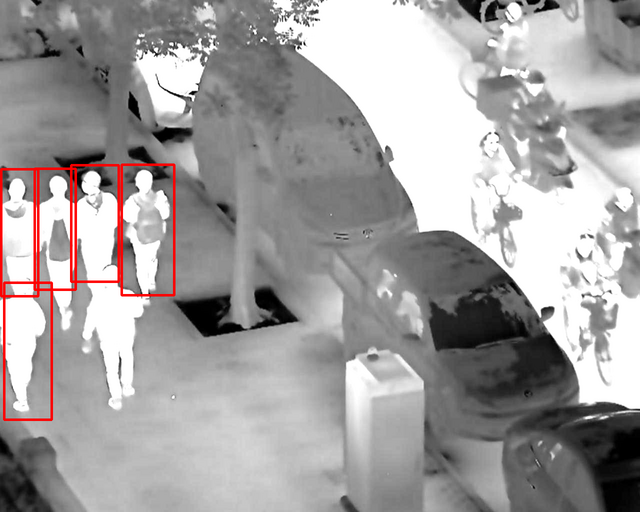}
        
        \includegraphics[width=\textwidth]{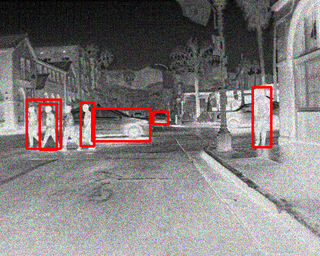}

    \end{subfigure}
    \begin{subfigure}[!htp]{0.22\textwidth}
        \caption{WiSE-OD}
    
        \includegraphics[width=\textwidth]{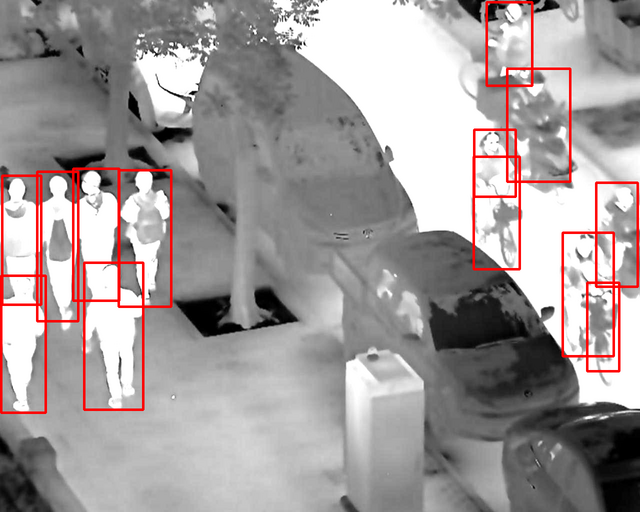}
        \includegraphics[width=\textwidth]{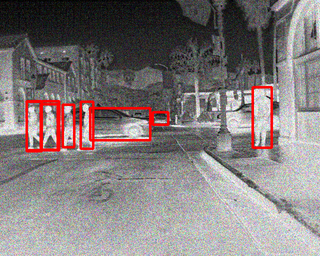}

    \end{subfigure}

    \setcounter{figure}{0}
    \captionof{figure}{\textbf{Robustness of Infrared Object Detection on LLVIP-C and FLIR-C datasets.} In the first row, LLVIP-C has a brightness corruption severity level of $5$; in the second row, FLIR-C shot noise corruption has a severity level of $2$. In (a) ground-truth boxes (yellow); (b) zero-shot COCO; (c) fine-tuning (FT); (d) WiSE-OD with Faster R-CNN.}
\end{center}%
}]

\begin{abstract}
Object detection (OD) in infrared (IR) imagery is critical for low-light and nighttime applications. However, the scarcity of large-scale IR datasets forces models to rely on weights pre-trained on RGB images. While fine-tuning on IR improves accuracy, it often compromises robustness under distribution shifts due to the inherent modality gap between RGB and IR. 
To address this, we introduce LLVIP-C and FLIR-C, two cross-modality out‑of‑distribution (OOD) benchmarks built by applying corruptions to standard IR datasets. Additionally, to fully leverage the complementary knowledge from RGB and infrared-trained models, we propose WiSE-OD, a weight-space ensembling method with two variants: WiSE-OD$_{ZS}$, which combines RGB zero-shot and IR fine-tuned weights, and WiSE-OD$_{LP}$, which blends zero-shot and linear probing. Evaluated using four RGB-pretrained detectors and two robust baselines on our benchmark and in the real-world out-of-distribution M3FD dataset, our WiSE-OD improves robustness across modalities and to corruption in synthetic and real-world distribution shifts without any additional training or inference costs. Our code is available at: \url{https://github.com/heitorrapela/wiseod}.
\end{abstract}

\section{Introduction}
\label{sec:intro}

In recent years, deep learning (DL) has achieved significant success across various computer vision tasks, including object detection (OD)~\cite{zou2023object} using thermal infrared (IR) imaging~\cite{medeiros2024hallucidet, modprompt}. Unlike visible spectrum imaging (RGB), which relies on reflected light, thermal IR imaging captures the heat emitted by objects, allowing it to function independently of lighting conditions. This makes IR-based OD highly effective in challenging environments with limited or absent visible light, such as night-time surveillance and autonomous driving cars~\cite{teutsch2021computer}. Despite these advantages, IR OD models must maintain consistent performance and reliable predictions under variations in input due to occlusions, viewpoint shifts, or image degradation. Ensuring such robustness is therefore essential for real-world applications in surveillance~\cite{ramachandran2021review,dubail2022privacy}, autonomous vehicles~\cite{michaelis2019benchmarking}, and defense~\cite{liu2024benchmarking}, where fluctuating environmental conditions can impact sensor inputs and compromise system reliability.

Without large-scale IR pre-training datasets, OD models for IR typically initialize from powerful models pre-trained on large-scale RGB datasets (e.g., COCO~\cite{lin2014microsoft}), followed by fine-tuning on IR data. While this pipeline yields strong in-domain (ID) performance, where ID refers to test samples similar to the training data, it often compromises robustness against out-of-domain (OOD) samples~\cite{hendrycks2019benchmarking}. OOD samples differ significantly from the training data, leading to performance degradation. This deterioration stems from the fact that the fine-tuning process tends to cause the model to prioritize task-specific information at the expense of broader knowledge acquired during pre-training. As a result, the model struggles to generalize to new or diverse scenarios~\cite{wortsman2022robust}. This issue is further amplified by the already substantial modality shift between RGB and IR, making robust transfer learning even more difficult~\cite{medeiros2024modality}. In classification, several techniques have been proposed to improve robustness under distribution shifts, including linear probing (LP), LP followed by fine-tuning (LP-FT)~\citep{kumar2022fine}, and weight-space ensembling (WiSE-FT)~\citep{wortsman2022robust}. While these methods effectively enhance robustness in classification, they either are not directly applicable to object detection or remain under-explored~\citep{wortsman2022robust}. This gap arises from the complexity of the detection architectures and objectives, which require both precise localization and accurate classification. Moreover, cross-modality adaptation from RGB to infrared (IR) introduces additional challenges due to the modality shift and the scarcity of large-scale IR data~\citep{medeiros2024hallucidet, medeiros2024modality, modprompt}.

To address these challenges, this paper introduces two key components: (i) a novel cross-modality RGB/IR corruption benchmark and (ii) two efficient approaches to improve average IR performance under OOD scenarios without additional training or inference cost. Our benchmark, LLVIP-C and FLIR-C, applies common corruption transforms to the original LLVIP and FLIR datasets to evaluate cross-modality OOD performance. Using this benchmark, we assess three families of IR detectors fine-tuned from RGB pre-trained models against standard robust fine-tuning baselines. Our analysis shows that traditional methods underperform under corruption. Therefore, we introduce WiSE-OD, a simple approach that preserves the original detection head to combine zero-shot and fine-tuned weights, yielding WiSE-OD$_{ZS}$ and its linear probing variation WiSE-OD$_{LP}$. Results show that these weight-space ensembling methods exhibit significant robustness. Additional analysis across various levels of corruption demonstrates that these methods improve average IR model performance by preserving ID accuracy from fine-tuning and OOD robustness from zero-shot weights, which explains their effectiveness. Our method also performs well under real-world shifts such as IR images captured at night, indoors, and under fog or rain, when the original models were trained solely on daytime IR images. \\

\noindent \textbf{Our main contributions can be summarized as follows:} 

\begin{itemize}
\item A new benchmark, LLVIP-C and FLIR-C, is introduced to advance the evaluation of robust cross-modality OD between RGB and IR. This benchmark is essential for measuring detector performance across diverse, real-world conditions. Within this framework, we comprehensively evaluate three widely used object detection models: Faster R-CNN, FCOS, and RetinaNet, each initialized with COCO pre-trained weights.

\item We propose WiSE-OD with two variants: WiSE-OD$_{ZS}$ and WiSE-OD$_{LP}$, an efficient OD technique that combines zero-shot and fine-tuned weights to enhance robustness under real-world and synthetic distribution shifts.

\item Extensive experiments on the proposed benchmark and a real-world IR OOD dataset demonstrate significant gains for WiSE-OD over four OD frameworks (Faster R-CNN, FCOS, RetinaNet, and YOLOv8).
\end{itemize}

\section{Related Works}
\label{sec:formatting}

\noindent \textbf{Object detection.} OD is one of the most challenging computer vision tasks~\citep{liu2020deep}, especially due to many different environmental conditions~\citep{michaelis2019benchmarking}. The objective of OD is to localize with a bounding box and provide labels for all objects in an image~\citep{zhang2021dive}. Commonly, detectors can be categorized into two groups: one-stage and two-stage detectors. The most famous two-stage detector is Faster R-CNN~\citep{ren2015faster}, which first generates regions of interest and then uses a second classifier to confirm object presence within those regions. In contrast, one-stage detectors eliminate the proposal generation stage, targeting real-time inference. Among one-stage detectors, RetinaNet~\citep{lin2017focal} uses focal loss to address the class imbalance. Also, models such as FCOS~\citep{tian2019fcos} have emerged in this category, eliminating predefined anchor boxes to enhance inference efficiency. The YOLO family is a prominent line of one-stage detectors, with YOLOv8 adopting an anchor-free design with a decoupled head and offering a strong speed–accuracy trade-off. Our work focuses on these four detectors: Faster R-CNN, RetinaNet, FCOS, and YOLOv8. \\

\noindent \textbf{Robustness in Object Detection.} Robustness in OD refers to the model's ability to maintain performance despite variations in input conditions. Hendrycks $\&$ Dietterich~\cite{hendrycks2019benchmarking} proposed diverse corruptions for classification datasets, resulting in ImageNet-C and CIFAR10-C.~\citet{michaelis2019benchmarking} extended this to OD, proposing Pascal-C, COCO-C, and Cityscapes-C with a study on corruption severity and detector performance.~\citet{beghdadi2022benchmarking} introduced additional local transformations for RGB OD on COCO, and ~\citet{mao2023coco} proposed COCO-O with six types of natural distribution shifts. Despite growing efforts for RGB OD robustness, IR OD still lacks such benchmarks. In this direction, \citet{josi2023multimodal} applied classification corruptions to IR for person ReID. Given the widespread use of IR in surveillance and autonomous driving, a robustness benchmark for IR OD is essential. \\

\noindent \textbf{Robust Fine-Tuning.} The deep learning community has explored various fine-tuning (FT) strategies to improve robustness in classification tasks. A common approach is linear probing (LP), where the backbone is frozen and only the head is trained.~\citet{kumar2022fine} extended this with LP-FT, which first trains a linear head before unfreezing the backbone for full fine-tuning.~\citet{wortsman2022robust} proposed WiSE-FT, which ensembles the weights of a zero-shot pre-trained model and its fine-tuned counterpart in weight space, showing strong performance under distribution shifts on ImageNet. \\

\noindent In this work, we adapt these robustness techniques, which were originally developed for classification, to the more challenging cross-modality object detection setting, offering simple yet effective strategies to mitigate corruption effects.

\section{Background} 
\label{sec:proposed_method}

In this section, we introduce preliminary definitions that are necessary to understand this work, and subsequently, we define our proposed benchmark.

\noindent \textbf{Object Detection.} Consider a set of training samples $\mathcal{D} = \{(x_{i}, B_{i})\}$, where $x_{i} \in \mathbb{R}^{W \times H \times C}$ are images with spatial resolution $W \times H$ and $C$ channels, and $B_{i} = \{{b_{0}, b_{1},...,b_{N}\}}$ is a set of bounding boxes corresponding to the image $x_{i}$. Each bounding box can be represented as $b = (c_{x}, c_{y}, w, h, o)$ where $c_{x}$ and $c_{y}$ are the center coordinates of the bounding box with size $w \times h$ and $o$ is the class label. During training we aim to learn a parameterized function $f_{\theta}: \mathbb{R}^{W \times H \times C} \rightarrow \mathcal{B}$, with $\mathcal{B}$ being the family of sets $B_{i}$ and $\theta$ the model's parameter vector. The optimization of $f_{\theta}$ is guided by a combination of a regression $\mathcal{L}_{r}$ and classification $\mathcal{L}_{c}$ loss, i.e., $l_{2}$ loss and binary cross-entropy, respectively. The loss function for object detection can be represented as:
\begin{equation}
   \mathcal{L}_{d}(\theta)=\frac{1}{|\mathcal{D}|} \sum_{(x, B) \in \mathcal{D}} \mathcal{L}_{c}(f_{\theta}(x), B)+\lambda \mathcal{L}_{\text {r}}(f_{\theta}(x), B).
    \label{eq:detection_loss}
\end{equation}

\noindent \textbf{Robustness to Corruption.} Corruption robustness measures a classifier’s average performance under classifier‑agnostic input distortions ~\cite{hendrycks2019benchmarking}. 
Let $f : \mathcal{X} \to \mathcal{Y}$ be a classifier trained on samples drawn from a distribution $\mathcal{Q}$, and let 
$C = \{\,c:\mathcal{X}\to\mathcal{X}\}$
be a set of corruption functions (e.g.,\ noise, blur, contrast) and 
$\mathcal{E} = \{\,e:\mathcal{X}\to\mathcal{X}\}$
a set of additional perturbation functions. The classifier’s clean accuracy is $\mathbb{P}_{(x,y)\sim\mathcal{Q}}\bigl(f(x)=y\bigr)$.
Its corruption robustness, i.e., its expected accuracy under all compositions of one corruption and one perturbation, can be represented as:
\[
\mathbb{E}_{c\sim C}\,\mathbb{E}_{e\sim\mathcal{E}}\Bigl[\,
\mathbb{P}_{(x,y)\sim\mathcal{Q}}\bigl(f\bigl(e(c(x))\bigr)=y\bigr)\Bigr].
\]

\noindent \textbf{Weight-space Ensembling.} Given a mixing coefficient $\lambda \in [0,1]$, weight-space ensembling can be defined as the following function: 
\begin{equation}
f_{\mathrm{wse}}(\theta_i, \theta_j; \lambda) \;=\; (1-\lambda)\,\theta_i \;+\; \lambda\,\theta_j, 
\label{eq:wiseft}
\end{equation}
which computes the element‑wise convex combination of two parameter vectors $\theta_i$ and $\theta_j$. The resulting ensemble parameter $\theta_{\mathrm{ens}} = f_{\mathrm{wse}}(\theta_i,\theta_j;\lambda)$ is then used to initialize the model for prediction. A notable example of this technique is WiSE‑FT~\cite{wortsman2022robust}. Moreover, weight‑space ensembling builds on principles of output‑space ensemble averaging~\cite{izmailov2018averaging} and has demonstrated improved OOD robustness on classification benchmarks~\cite{wortsman2022model}.

\section{OD IR Robustness Benchmark}

\begin{figure}[ht]
\captionsetup[subfigure]{labelformat=empty}
\centering
\resizebox{0.70\columnwidth}{!}{%

\begin{subfigure}[t]{0.60\columnwidth}
    \makebox[0pt][r]{\makebox[15pt]{\raisebox{50pt}{\rotatebox[origin=c]{90}{LLVIP-C}}}}%
    \bottomfloat[Shot Noise]{\includegraphics[width=\textwidth]{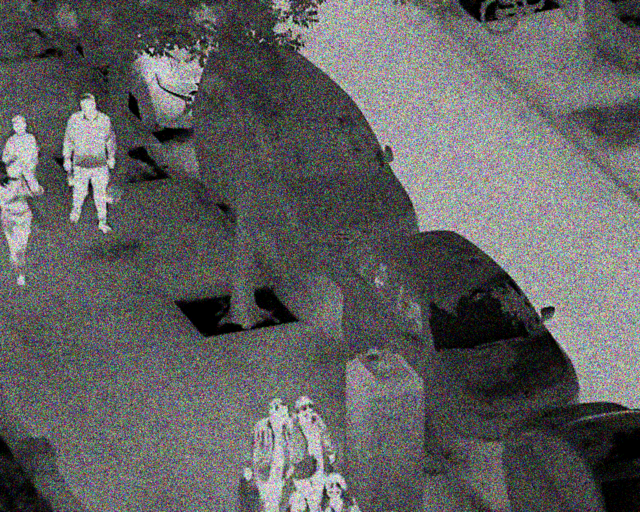}}\hspace{0.1\textwidth}
    
    \makebox[0pt][r]{\makebox[15pt]{\raisebox{50pt}{\rotatebox[origin=c]{90}{FLIR-C}}}}%
    \topfloat[Motion Blur]{\includegraphics[width=\textwidth]{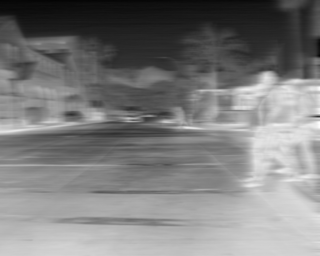}}\hspace{0.1\textwidth}
    
\end{subfigure}

\begin{subfigure}[t]{0.60\columnwidth}

    \bottomfloat[Impulse Noise]{\includegraphics[width=\textwidth]{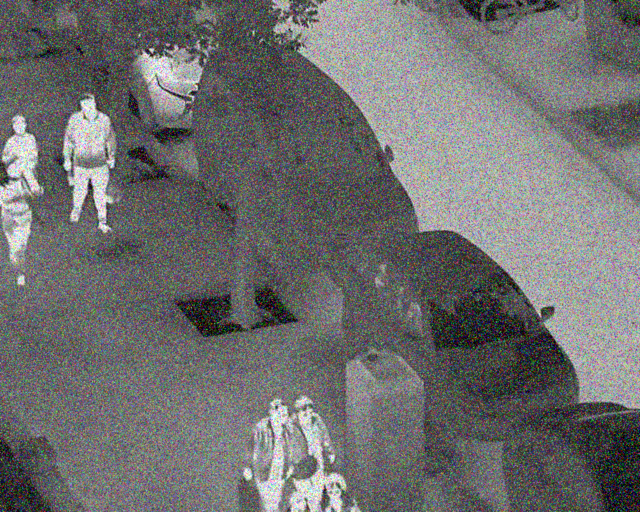}}\hspace{0.1\textwidth} 

    \topfloat[Zoom Blur]{\includegraphics[width=\textwidth]{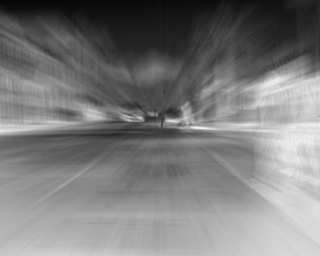}}\hspace{0.1\textwidth}
    
\end{subfigure}

}
\caption{\textbf{LLVIP-C and FLIR-C examples.} First row, we have one example from the LLVIP-C test set with two different corruptions: Shot Noise, and Impulse Noise with a severity level of $5$. In the second row, we have one example from the FLIR-C test set with Motion Blur and Zoom Blur with a severity level of $5$.}
\label{fig:llvip_flir_corruptions_samples}
\end{figure}

\begin{figure}[h]
    \centering
    \setlength{\tabcolsep}{1pt} 
    \renewcommand{\arraystretch}{0.1}
    
    \begin{tabular}{cccccc}
        & \textbf{\footnotesize{Sev. 1}} & \textbf{\footnotesize{Sev. 2}} & \textbf{\footnotesize{Sev. 3}} & \textbf{\footnotesize{Sev. 4}} & \textbf{\footnotesize{Sev. 5}} \\
        
        \rotatebox{90}{\textbf{\;\;\footnotesize{Fog}}} &
        \includegraphics[width=0.17\linewidth]{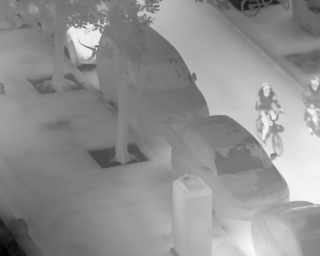} &
        \includegraphics[width=0.17\linewidth]{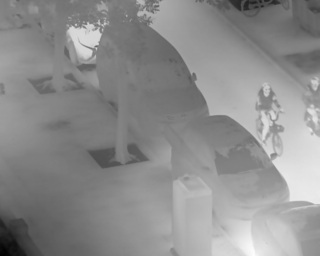} &
        \includegraphics[width=0.17\linewidth]{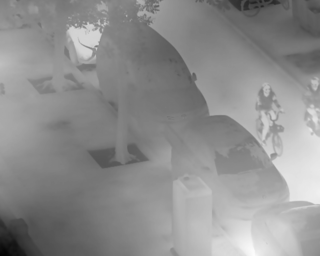} &
        \includegraphics[width=0.17\linewidth]{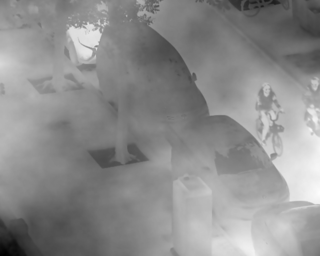} &
        \includegraphics[width=0.17\linewidth]{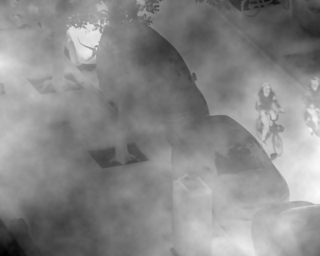} \\

    \end{tabular}

    \begin{tabular}{cccccc}

        \rotatebox{90}{\textbf{\;\;\footnotesize{Brig.}}} &
        \includegraphics[width=0.17\linewidth]{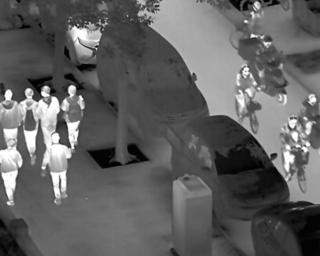} &
        \includegraphics[width=0.17\linewidth]{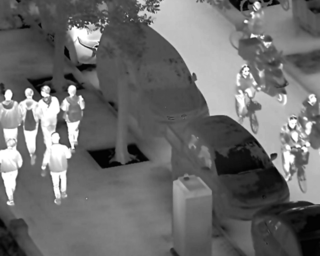} &
        \includegraphics[width=0.17\linewidth]{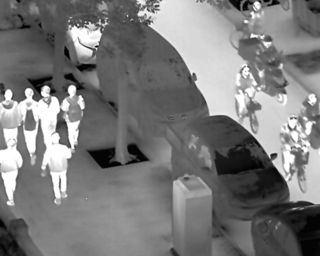} &
        \includegraphics[width=0.17\linewidth]{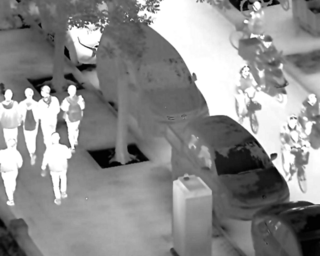} &
        \includegraphics[width=0.17\linewidth]{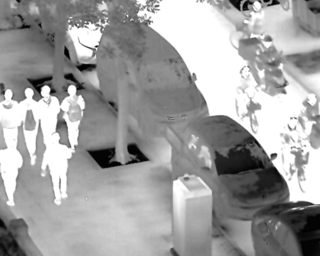} \\

    \end{tabular}

    \begin{tabular}{cccccc}

        \rotatebox{90}{\textbf{\;\;\footnotesize{Contr.}}} &
        \includegraphics[width=0.17\linewidth]{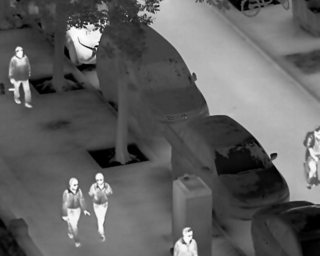} &
        \includegraphics[width=0.17\linewidth]{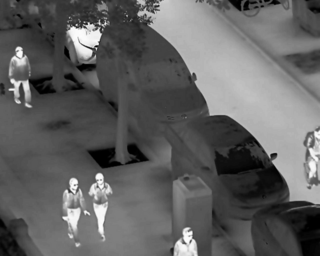} &
        \includegraphics[width=0.17\linewidth]{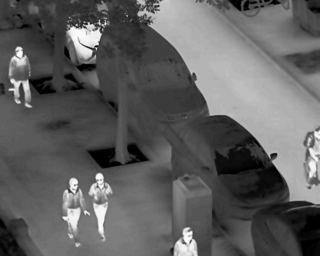} &
        \includegraphics[width=0.17\linewidth]{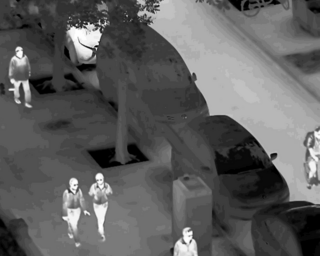} &
        \includegraphics[width=0.17\linewidth]{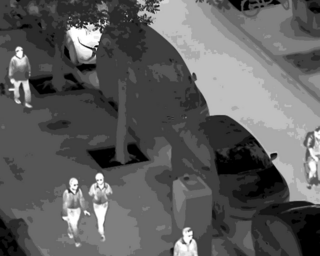} \\

    \end{tabular}
  
    \caption{\textbf{Examples of fog perturbations at different severity levels for LLVIP.} Each column shows the effect of increasing corruption severity (1–5) on infrared images. Rows: top-fog, middle-brightness, and bottom-contrast. Higher severities introduce stronger degradations, simulating real-world challenging conditions.}

    \label{fig:corruptions_different_severities}
    \vspace{-.5cm}

\end{figure}

\subsection{Benchmark Datasets} For our proposed robust IR OD benchmark with corruptions, we explore two classical datasets containing paired RGB and infrared images: LLVIP and FLIR. Additionally, we explore the M3FD dataset, which has real-world shifts.

\noindent\textbf{LLVIP:} LLVIP is a surveillance dataset composed of $12,025$ paired IR and RGB images for training and $3,463$ paired IR and RGB images for testing. The resolution of images is $1280 \times 1024$ pixels, and annotations consist of bounding boxes around pedestrians. \textbf{FLIR ALIGNED:} For the FLIR dataset, we used the sanitized and aligned paired sets provided ~\cite{zhang2020multispectral}, which contains $4,129$ paired IR and RGB images for training, and $1,013$ paired IR and RGB images for testing. The FLIR images are captured by a front‑mounted car camera at a resolution of $640\times512$ pixels, and annotations contain bicycles, dogs, cars, and people. \textbf{M3FD (Real-World OOD):} M3FD~\cite{liu2022target} is a paired RGB-IR benchmark with well-aligned images captured by a calibrated dual-sensor rig. It contains 4{,}200 aligned pairs at 1024$\times$768 resolution and six classes (person, car, bus, motorcycle, truck, lamp). We considered the day IR images as training data, and for testing, we considered fog, night, rain, and indoor images. The high resolution, alignment quality, and scenario diversity make M3FD a strong testbed for robustness in multi-modality detection and fusion.

\noindent\textbf{LLVIP-C and FLIR-C:} In this section, we present our two corrupted benchmarks: LLVIP-C and FLIR-C, derived from the LLVIP and FLIR datasets.
In~\fref{fig:llvip_flir_corruptions_samples}, the first row shows an LLVIP-C test example corrupted with Shot Noise and Impulse Noise at severity level $5$. The second row shows an FLIR-C test example corrupted with Motion Blur and Zoom Blur at severity level~5. As illustrated qualitatively, severity level~5 is too strong for the FLIR images, already compressed JPEGs, and both zero-shot and fine-tuned models perform worse on FLIR-C than on LLVIP-C at this level. Therefore, we recommend a maximum corruption severity of $2$ for FLIR-C based on qualitative and quantitative results. For the following experiments, we use severity level $5$ for LLVIP-C and severity level $2$ for FLIR-C. In~\fref{fig:corruptions_different_severities}, we show the impact of different severity levels (pre-defined levels) defined by~\citep{michaelis2019benchmarking} on the IR images going from severity 1 to severity 5.

\begin{figure}[t]
    \centering
    \includegraphics[width=0.90\linewidth]{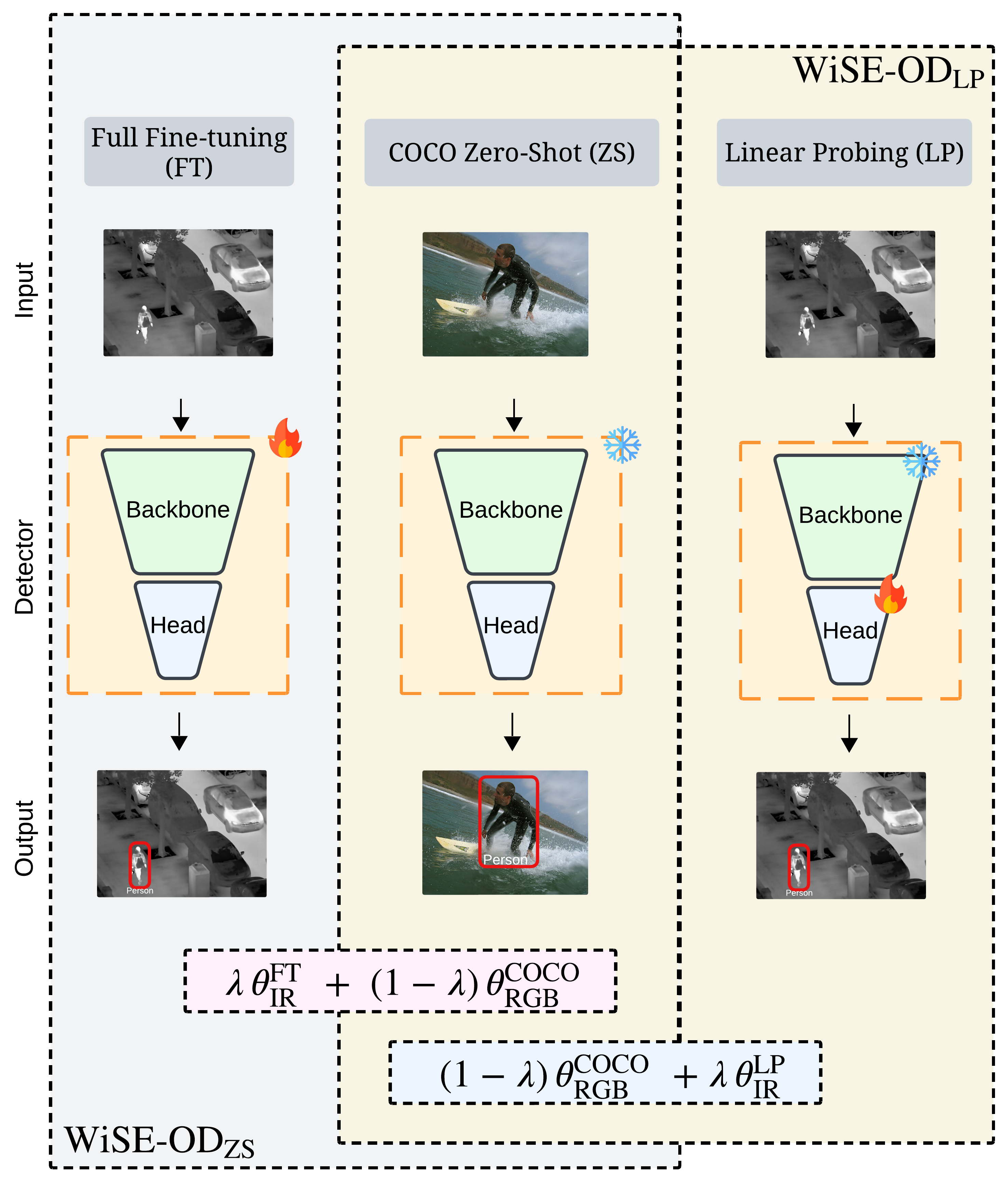}

\caption{\textbf{Our proposed method: WiSE-OD and its variants.} In the large grey box, we have WiSE-OD$_\text{ZS}$ with the equation inside the pink square, and WiSE-OD$_\text{LP}$ in the yellow large box with the equation inside the blue square.}
\label{fig:wiseod}
\vspace{-.5cm}
\end{figure}

\section{WiSE-OD} Our proposed method, WiSE-OD ($f_{\mathrm{wod}}$) in~\fref{fig:wiseod}, extends the idea of WiSE-FT to object detection setting. Let $\theta_{\text{RGB}}^{\text{COCO}}$ and $\theta_{\text{IR}}^{\text{FT}}$ denote the parameters of the RGB pre-trained COCO detector and the fully fine-tuned IR detection models, respectively. WiSE-OD constructs a new detector by interpolating these parameter vectors in weight space:
\begin{equation}
f_{\mathrm{wod}}(\theta_{\text{RGB}}^{\text{COCO}}, \theta_{\text{IR}}^{\text{FT}}; \lambda) \;=\; (1-\lambda)\,\theta_{\text{RGB}}^{\text{COCO}} \;+\; \lambda\,\theta_{\text{IR}}^{\text{FT}}.
\label{eq:wiseod}
\end{equation}
\noindent The resulting interpolated model inherits both the broad generalization of large‑scale COCO pre-training and the modality‑specific accuracy of IR fine‑tuning, yet requires no extra modules or change to the inference pipeline, only a one‑time weight merge. We evaluate two variants:  WiSE-OD\(_{ZS}\) uses \(\theta_{\text{IR}}^{\text{FT}}\) (full fine‑tuning), and  WiSE-OD\(_{LP}\) uses \(\theta_{\text{IR}}^{\text{LP}}\) (linear probing on the detection head with a frozen backbone). Both variants consistently improve robustness under domain shift and common corruptions, while maintaining the same inference cost as a single detector. This weight‑space ensembling is model‑agnostic and can be extended to fuse multiple checkpoints or modalities by hierarchical interpolation.

\noindent \textbf{Metrics:} Following the methodology for benchmarking robustness in OD~\cite{michaelis2019benchmarking}, we select AP$_{50}$ as our detection performance metric for both LLVIP-C and FLIR-C evaluations. We also report the dataset-specific performance ($P$), defined as AP$_{50}$ on the original target dataset (infrared), and mean performance under corruption, mPC, defined as:
\begin{equation} 
    mPC = \frac{1}{N_{c}}\sum_{c=1}^{N_{c}}P_{c},
    \label{eq:mpc}
\end{equation}
\noindent where $P_{c}$ is the AP$_{50}$ under corruption, and $mPC$ is the average over all the corruptions. In our case, $N_{c}$ = $14$ since we decided to remove the glass blur corruption because it lacks a fast implementation for our benchmark. \\

\noindent \textbf{Baseline models:} In our study, we utilize four OD architectures: Faster R-CNN, FCOS, RetinaNet, and YOLOv8, all initialized with COCO pre-trained weights. These models are trained on the COCO dataset, which contains 80 object categories, providing strong initial performance for most detection tasks and facilitating subsequent fine‑tuning. We evaluate the following robust fine-tuning methods using our proposed benchmark:
\begin{enumerate}
    \item \textbf{Zero-Shot (ZS) --} Unmodified detectors used directly for deployment without any fine‑tuning.
    \item \textbf{Linear probing (LP) --} Train the classification and regression heads on top of a frozen backbone by minimizing the detection loss.
    \item \textbf{Full fine-tuning (FT) --} Update both the detection heads and the backbone parameters by minimizing the detection loss. 
    \item \textbf{LP-FT --} A two‑stage process in which we first apply linear probing and then perform full fine‑tuning initialized from the LP stage.
    \item \textbf{Weight-ensembling --} Two variants, WiSE‑OD$_{ZS}$ and WiSE‑OD$_{LP}$, which interpolate parameters between the zero‑shot or linear‑probing models and the fully fine‑tuned models, respectively.
\end{enumerate}
\begin{table*}[t]
\caption{\textbf{AP$_{50}$ performance over the perturbations on different datasets.} For LLVIP-C with severity level $5$, and FLIR-C with severity level $2$ for Faster R-CNN.}
\label{tab:fasterrcnn_llvip_per_corruption}
    \centering
    \resizebox{0.78\textwidth}{!}{%
    \begin{tabular}{lcccccccc}

        \toprule

        {} & {} & \multicolumn{4}{c}{\multirow{2}{*}[-0.1em]{\textbf{LLVIP-C}}} \\

        \multirow{2}{*}[-1em]{} & \multirow{2}{*}[-1em]{Zero-Shot} & 
        \multirow{2}{*}[-1em]{FT} &
        \multirow{2}{*}[-1em]{LP} & \multirow{2}{*}[-1em]{LP-FT} & \multirow{2}{*}[-1em]{WiSE-OD$_{ZS}$} & \multirow{2}{*}[-1em]{WiSE-OD$_{LP}$} \\

        \cmidrule(lr){2-7}
        \addlinespace[10pt]     
        \midrule

        Original & 71.21 ± 0.02 & 93.68 ± 0.86 & 91.82 ± 0.15 & 92.18 ± 0.03 & 96.06 ± 0.22 & 96.24 ± 0.03 \\

        \midrule
        
        \rowcolor[HTML]{EFEFEF}
        Gaussian Noise    & 59.24 ± 0.07 & 67.46 ± 7.45  & 75.12 ± 0.12 & 72.51 ± 0.28 & 86.68 ± 0.44 & 85.45 ± 0.76 \\
        Shot Noise        & 51.48 ± 0.14 & 64.83 ± 7.79  & 70.82 ± 0.27 & 69.89 ± 0.26 & 85.26 ± 0.50 & 85.25 ± 0.12 \\
        \rowcolor[HTML]{EFEFEF}
        Impulse Noise     & 56.62 ± 0.07 & 71.32 ± 6.33  & 78.31 ± 1.13 & 75.20 ± 0.86 & 88.54 ± 0.33 & 88.40 ± 0.15 \\
        Defocus Blur      & 47.90 ± 0.08 & 80.48 ± 3.60  & 84.31 ± 0.24 & 83.12 ± 0.05 & 89.74 ± 0.98 & 90.40 ± 0.03 \\
        \rowcolor[HTML]{EFEFEF}
        Motion Blur       & 26.39 ± 0.23 & 78.32 ± 3.18  & 77.15 ± 0.33 & 75.13 ± 0.05 & 86.81 ± 0.71 & 87.02 ± 0.32 \\

        Zoom Blur         & 02.47 ± 0.02 & 11.18 ± 1.56  & 24.65 ± 0.39 & 17.46 ± 0.02 & 22.83 ± 2.44 & 27.08 ± 0.01 \\
        
        \rowcolor[HTML]{EFEFEF}
        Snow              & 33.65 ± 0.01 & 13.46 ± 4.45  & 69.92 ± 0.14 & 69.34 ± 0.13 & 65.97 ± 1.90 & 65.28 ± 2.70 \\

        Frost             & 33.25 ± 0.38 & 47.32 ± 3.45  & 68.00 ± 0.27 & 66.93 ± 0.42 & 75.87 ± 0.39 & 74.85 ± 0.29 \\

        \rowcolor[HTML]{EFEFEF}
        Fog               & 59.60 ± 0.10 & 50.90 ± 10.07 & 87.05 ± 0.07 & 87.33 ± 0.39 & 84.51 ± 3.80 & 88.17 ± 0.06 \\

        Brightness        & 41.77 ± 0.03 & 35.36 ± 6.97  & 71.47 ± 0.34 & 76.45 ± 0.14 & 82.10 ± 1.20 & 82.61 ± 0.92 \\

        \rowcolor[HTML]{EFEFEF}
        Contrast          & 47.48 ± 0.04 & 00.00 ± 0.00  & 51.53 ± 0.03 & 48.93 ± 0.02 & 10.57 ± 3.82 & 14.30 ± 1.82 \\
        
        Elastic transform & 52.42 ± 0.18 & 92.41 ± 0.93  & 86.30 ± 0.25 & 88.98 ± 0.14 & 94.72 ± 0.07 & 94.85 ± 0.14 \\

        \rowcolor[HTML]{EFEFEF}
        Pixelate          & 03.95 ± 0.01 & 87.69 ± 2.67  & 65.35 ± 0.09 & 65.71 ± 0.04 & 85.06 ± 3.32 & 84.33 ± 0.05 \\

        JPEG compression  & 57.22 ± 0.02 & 88.93 ± 1.69  & 83.58 ± 0.07 & 83.32 ± 0.24 & 92.59 ± 1.22 & 93.73 ± 0.03 \\

        \rowcolor[HTML]{EFEFEF}

        \midrule
        mPC  & 40.96 & 56.40 & 70.96 & 70.02 & 75.08 & \textbf{75.83} \\

        \midrule
        {} & {} & \multicolumn{4}{c}{\multirow{2}{*}[-0.1em]{\textbf{FLIR-C}}} \\

        \multirow{2}{*}[-1em]{} & \multirow{2}{*}[-1em]{Zero-Shot} & 
        \multirow{2}{*}[-1em]{FT} &
        \multirow{2}{*}[-1em]{LP} & \multirow{2}{*}[-1em]{LP-FT} & \multirow{2}{*}[-1em]{WiSE-OD$_{ZS}$} & \multirow{2}{*}[-1em]{WiSE-OD$_{LP}$} \\

        \cmidrule(lr){2-7}
        \addlinespace[10pt]     
        \midrule

        \rowcolor[HTML]{EFEFEF}
        Gaussian Noise 		& 31.21 ± 0.29 & 28.07 ± 2.91 & 41.99 ± 0.39 & 39.83 ± 0.44 & 42.49 ± 4.48 & 39.33 ± 0.36 \\

        Shot Noise 			& 25.26 ± 0.12 & 15.73 ± 2.05 & 33.24 ± 0.23 & 33.15 ± 0.43 & 30.45 ± 3.96 & 35.84 ± 0.46 \\

        \rowcolor[HTML]{EFEFEF}
        Impulse Noise 		& 17.69 ± 0.03 & 13.22 ± 2.27 & 26.15 ± 0.46 & 25.58 ± 0.46 & 22.51 ± 2.78 & 26.72 ± 0.18 \\

        Defocus Blur 		& 25.32 ± 0.22 & 52.47 ± 0.99 & 44.57 ± 0.12 & 45.29 ± 0.20 & 54.08 ± 1.74 & 56.50 ± 1.22 \\

        \rowcolor[HTML]{EFEFEF}
        Motion Blur 		& 25.01 ± 0.25 & 51.71 ± 2.12 & 43.01 ± 0.41 & 42.79 ± 0.48 & 51.03 ± 2.16 & 57.24 ± 0.28 \\

        Zoom Blur 			& 08.98 ± 0.05 & 17.97 ± 0.90 & 15.17 ± 0.02 & 14.17 ± 0.07 & 16.93 ± 1.00 & 19.34 ± 0.17 \\

        \rowcolor[HTML]{EFEFEF}
        Snow 				& 09.84 ± 0.14 & 07.86 ± 2.01 & 16.94 ± 0.31 & 19.57 ± 0.58 & 13.94 ± 2.66 & 16.31 ± 0.23 \\

        Frost 				& 21.96 ± 0.50 & 33.87 ± 4.69 & 36.15 ± 0.29 & 37.82 ± 0.43 & 37.97 ± 3.63 & 38.80 ± 2.87 \\

        \rowcolor[HTML]{EFEFEF}
        Fog 				& 56.36 ± 0.28 & 73.61 ± 0.06 & 71.90 ± 0.37 & 71.26 ± 0.15 & 78.68 ± 1.24 & 78.10 ± 1.09 \\

        Brightness 			& 64.41 ± 0.26 & 75.18 ± 0.99 & 74.92 ± 0.20 & 74.24 ± 0.10 & 79.72 ± 0.35 & 78.42 ± 0.17 \\

        \rowcolor[HTML]{EFEFEF}
        Contrast 			& 54.59 ± 0.04 & 75.47 ± 1.29 & 71.11 ± 0.13 & 70.60 ± 0.30 & 78.36 ± 1.06 & 79.72 ± 0.15 \\

        Elastic transform 	& 41.88 ± 0.24 & 69.68 ± 1.15 & 64.49 ± 0.18 & 64.29 ± 0.06 & 73.39 ± 0.40 & 73.83 ± 0.54 \\

        \rowcolor[HTML]{EFEFEF}
        Pixelate 			& 38.67 ± 0.11 & 54.91 ± 6.21 & 55.61 ± 0.09 & 55.53 ± 0.13 & 61.12 ± 3.13 & 56.54 ± 0.01 \\

        JPEG compression 	& 50.24 ± 0.14 & 57.55 ± 3.04 & 64.36 ± 0.30 & 62.82 ± 0.17 & 66.65 ± 0.89 & 65.70 ± 0.27 \\

        \bottomrule
        \rowcolor[HTML]{EFEFEF}
        mPC & 33.67 & 44.80 & 47.11 & 46.92 & 50.52 & \textbf{51.59} \\

        \bottomrule

    \end{tabular}
    }
\vspace{-.1cm}
\end{table*}

\section{Experiments and Results}
\label{sec:experiments}

\subsection{Training protocol}

For this work, we split each training dataset into $80\%$ for training and $20\%$ for validation, reserving the original test set for final evaluation. All models were implemented in PyTorch, optimized with the Adam optimizer, and trained on an NVIDIA A100 GPU. We set a maximum training budget of 200 epochs for all detectors; in practice, fine‑tuning typically converges within 10–20 epochs, depending on the model and dataset. We used a cosine annealing scheduler on the training loss and applied early stopping based on validation AP$_{50}$. Our 14 corruption types and severities follow the ImageNet-C~\citep{hendrycks2019benchmarking}/COCO-C~\citep{michaelis2019benchmarking} protocol (see supplementary material for details). We provide code for reproducibility.

\subsection{Benchmark Quantitative Results}

In this section, we measured the mPC performance of all the proposed baselines for our benchmark on LLVIP-C with a severity level of $5$ and the FLIR-C dataset with a severity level of $2$. Results are shown in~\tref{tab:average_performance} for Faster R-CNN, FCOS, and RetinaNet under zero-shot, FT, LP, LP-FT, WiSE-OD$_{ZS}$, and WiSE-OD$_{LP}$. We see from~\tref{tab:average_performance} that, on average, WiSE-OD$_{ZS}$ with \(\lambda\) fixed at $0.5$, i.e., equal weighting of Zero‑Shot and FT, outperforms all other baselines without the need to tune any hyperparameters. For instance, on LLVIP‑C, WiSE‑OD$_{ZS}$ improved mPC by 18.68 over FT and by 4.12 over LP for Faster R‑CNN. In most cases, our proposed variant WiSE‑OD$_{LP}$ outperformed WiSE‑OD$_{ZS}$ for Faster R‑CNN and RetinaNet.

\subsection{Detection performance per corruption}

In this section, we evaluated the benchmark per corruption. In~\tref{tab:fasterrcnn_llvip_per_corruption}, we show the in-domain performance (evaluation on infrared of the LLVIP dataset), we name ``Original'', which is the original LLVIP infrared test set. Then, we have the corruptions for the LLVIP-C and the mPC metric for the Faster R-CNN detector; the same methodology was used for FLIR and FLIR-C. The original performance is measured in terms of AP$_{50}$ for Faster R-CNN; additional results for FCOS, RetinaNet, and all the detectors are provided in the supplementary material. As described in~\tref{tab:fasterrcnn_llvip_per_corruption}, the in-domain performance for LP ($91.82$) and LP-FT ($92.18$) is lower than the FT ($93.63$), but the mPC is much higher than the zero-shot and FT. The WiSE-OD$_{ZS}$ and WiSE-OD$_{LP}$ were able to outperform the others with in-domain of $96.06$ and $96.24$, respectively, and for the mPC, WiSE-OD$_{LP}$ achieves $75.83$ and WiSE-OD$_{ZS}$ $75.08$. For WiSE-OD$_{ZS}$ this corresponds to an increase of $18.68$ over FT and $4.12$ mPC over LP. For FLIR-C, we also have good improvements compared to the others. It is important to mention that the WiSE-OD$_{ZS}$ is a training-free technique, and for this table, $\lambda$ is fixed at $0.5$, same for WiSE-OD$_{LP}$, but this variation needs the LP model instead of the zero-shot.  In contrast, IR-adapted detectors already degrade substantially, which limits ensemble gains. For example, on LLVIP-C FT collapses (AP$_{50}=0.00$), while WiSE-OD$_{LP}$ recovers to 14.3 (+14.3). On FLIR-C, FT remains strong (75.5) and WiSE-OD$_{LP}$ improves to 79.7 (+4.2). Accordingly, we keep a fixed $\lambda$ to avoid target-data tuning.

\subsection{Performance over different corruption levels}

\begin{figure}[ht]
    \centering
    \begin{subfigure}[t]{0.5\linewidth}
        \centering
        \includegraphics[width=\linewidth]{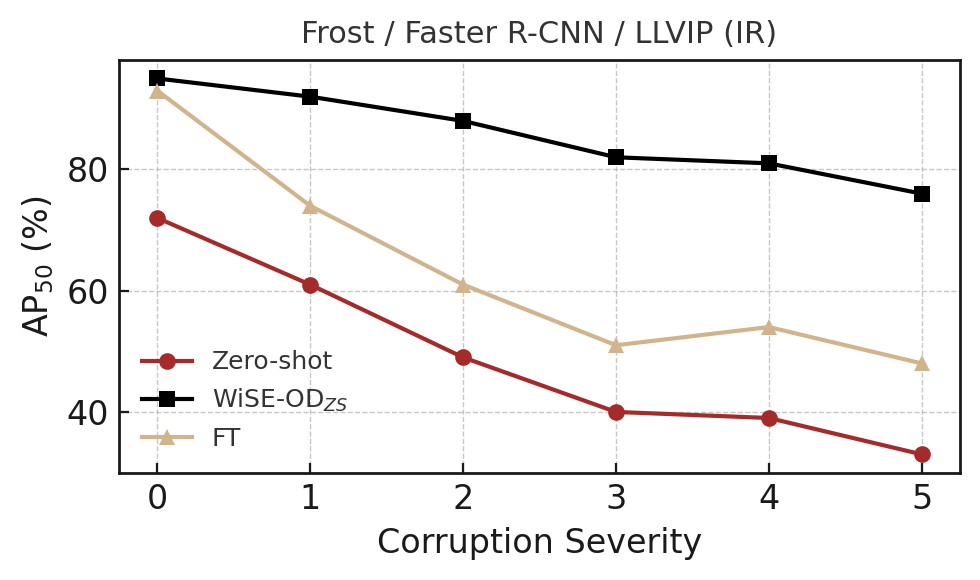}
        \caption{Frost (LLVIP-C)}
    \end{subfigure}%
    \begin{subfigure}[t]{0.5\linewidth}
        \centering
        \includegraphics[width=\linewidth]{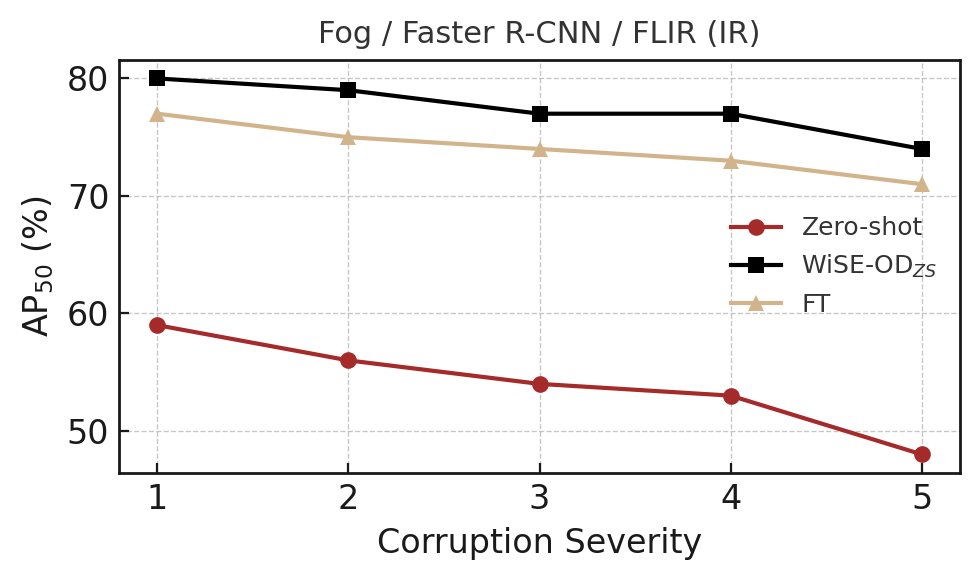}
        \caption{Fog (FLIR-C)}
    \end{subfigure}

    \begin{subfigure}[t]{1.0\linewidth}
        \centering
        \includegraphics[width=\linewidth]{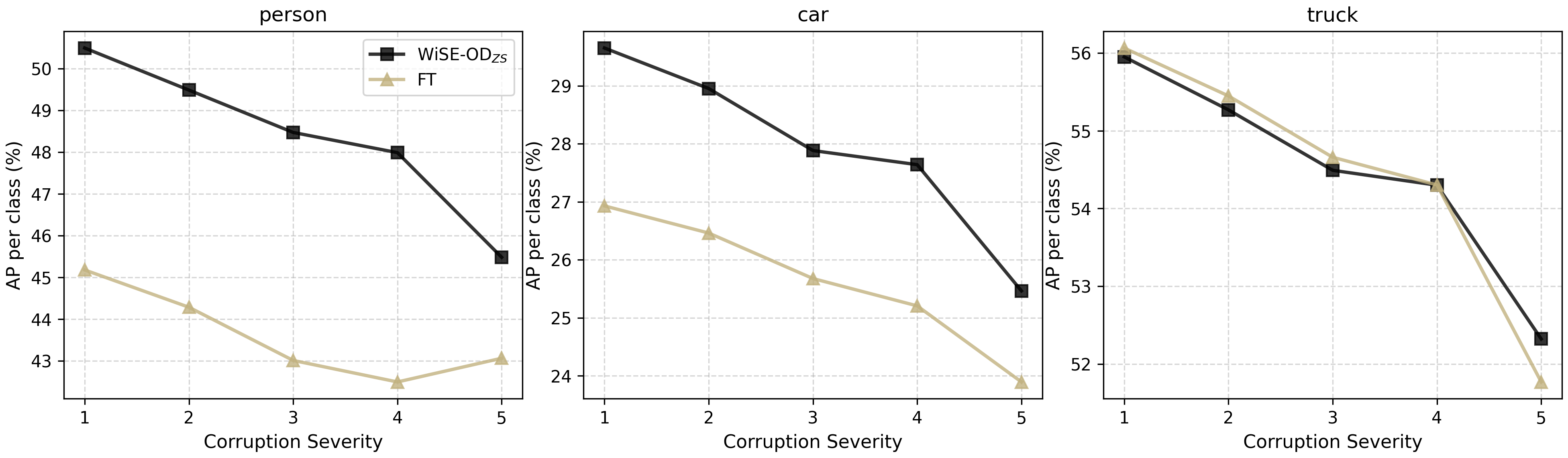}
        \caption{AP per class (FLIR-C)}
    \end{subfigure}
    
\caption{\textbf{AP$_{50}$ versus corruption severity.} 
(a) Frost on LLVIP-C (IR) and (b) Fog on FLIR-C (IR). Curves compare Faster R-CNN in Zero-shot, WiSE-OD$_{ZS}$, and FT settings; y-axis shows AP$_{50}$~(\%). Severity increases left-to-right (0–5 for LLVIP-C, 1–5 for FLIR-C). 
(c) FLIR-C per-class AP$_{50}$ under fog for \emph{person}, \emph{car}, and \emph{truck}. WiSE-OD$_{ZS}$ maintains a higher level of performance across severities.}

\label{fig:performance_over_corruption_levels}
\vspace{-.5cm}
\end{figure}

In this section, we measured the per AP$_{50}$ performance for Faster R-CNN, FCOS, and RetinaNet over different corruption severity levels for the benchmark. Here, in~\fref{fig:performance_over_corruption_levels}, we provided (a) Frost for LLVIP-C, (b) Fog for FLIR-C for Faster R-CNN, and (c) different APs for per-class analysis (person, car, truck) in FLIR-C. When the corruption severity level increases, e.g., from $1$ to $5$ in LLVIP-C, we can see a large drop in the zero-shot and FT, while the WiSE-OD$_{ZS}$ is more stable and can bring more robustness to the final model. Some corruptions have more impact than others for each dataset; for instance, in FLIR-C, the noise corruption affected the performance more due to the original low-quality images. In the IR modality, the contrast corruption affects the detection performance more because IR images already have low contrast when compared to natural RGB images. A similar trend of stability of WiSE-OD$_{ZS}$ for other detectors and corruptions over zero-shot and FT is shown in the supp. materials.

\begin{table}[!h]
\caption{\textbf{Detection performance for the OD IR Robustness Benchmark.} mPC metric for LLVIP-C with severity $5$ and FLIR-C with severity level $2$.}
\label{tab:average_performance}
    \centering
    \resizebox{0.99\columnwidth}{!}{%
    \begin{tabular}{lcccccc}

        \toprule

        \multirow{2}{*}[-1em]{\textbf{Detector}} & {} & \multicolumn{4}{c}{\multirow{2}{*}[-0.1em]{\textbf{LLVIP-C}}} \\

        \multirow{2}{*}[-1em]{} & \multirow{2}{*}[-1em]{Zero-Shot} & 
        \multirow{2}{*}[-1em]{FT} & \multirow{2}{*}[-1em]{LP} &
        \multirow{2}{*}[-1em]{LP-FT} &
        \multirow{2}{*}[-1em]{WiSE-OD$_{ZS}$} & \multirow{2}{*}[-1em]{WiSE-OD$_{LP}$} \\

        \cmidrule(lr){2-7}
        \addlinespace[10pt]        

        \midrule
        \rowcolor[HTML]{EFEFEF}
        
        Faster R-CNN & 40.96 & 56.40 & 70.96 & 70.02 & 75.08 & \textbf{75.83} \\
        \midrule
        
        FCOS  & 36.11 & 61.17 & 63.91 & 60.26 & \textbf{76.50} & 75.95 \\

        \midrule
        \rowcolor[HTML]{EFEFEF}
        RetinaNet & 37.50 & 61.13 & 60.37 & 61.11 & 73.69 & \textbf{74.39} \\

        \midrule

        \multirow{2}{*}[-1em]{\textbf{Detector}} & {} & \multicolumn{4}{c}{\multirow{2}{*}[-0.1em]{\textbf{FLIR-C}}} \\

        \multirow{2}{*}[-1em]{} & \multirow{2}{*}[-1em]{Zero-Shot} & 
        \multirow{2}{*}[-1em]{FT} & \multirow{2}{*}[-1em]{LP} &
        \multirow{2}{*}[-1em]{LP-FT} &
        \multirow{2}{*}[-1em]{WiSE-OD$_{ZS}$} & \multirow{2}{*}[-1em]{WiSE-OD$_{LP}$} \\

        \cmidrule(lr){2-7}
        \addlinespace[10pt]        

        \midrule
        \rowcolor[HTML]{EFEFEF}
        Faster R-CNN & 33.67 & 44.80 & 47.11 & 46.92 & 50.52 & \textbf{51.59} \\
        \midrule
        
        FCOS  & 28.85 & 41.07 & 38.92 & 38.14 & \textbf{47.13} & 46.76 \\

        \midrule
        \rowcolor[HTML]{EFEFEF}
        RetinaNet & 28.27 & 42.71 & 36.71 & 36.45 & 45.35 & \textbf{47.53} \\
        
        \bottomrule
    \end{tabular}
    }
\vspace{-.5cm}
\end{table}

\subsection{Activation map analysis}

\begin{figure}[!h]
    \centering
    \setlength{\tabcolsep}{1pt}  
    \renewcommand{\arraystretch}{0.25} 

    \begin{tabular}{cccc}
        & \multicolumn{1}{c}{\textbf{\footnotesize{ZS}}}
        & \multicolumn{1}{c}{\textbf{\footnotesize{WiSE-OD$_{ZS}$}}}
        & \multicolumn{1}{c}{\textbf{\footnotesize{FT}}} \\

        \rotatebox{90}{\textbf{\quad \footnotesize{Imp.\ N.}}} &
        \includegraphics[width=0.29\linewidth]{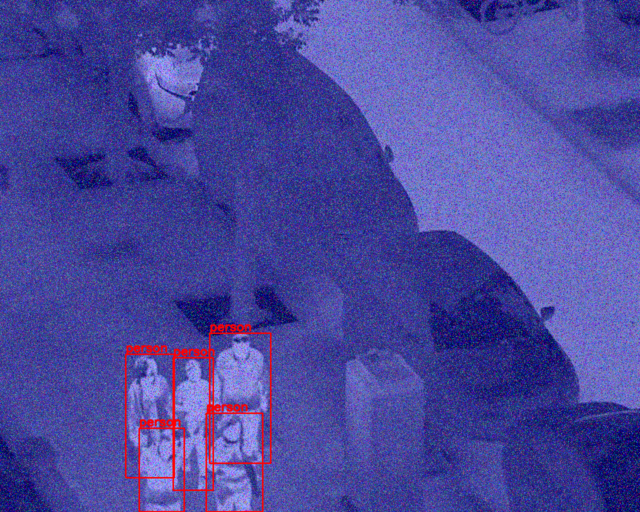} &
        \includegraphics[width=0.29\linewidth]{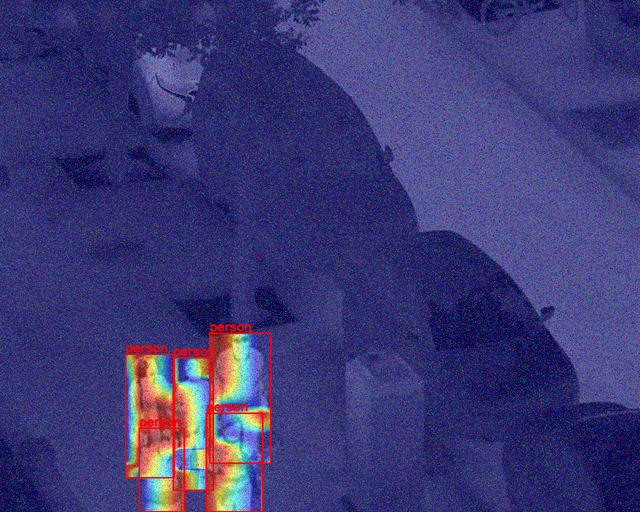} &
        \includegraphics[width=0.29\linewidth]{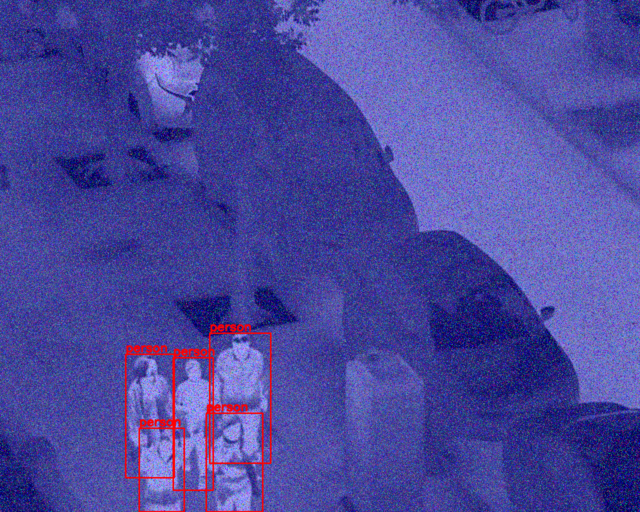} \\

        \rotatebox{90}{\quad \; \textbf{\footnotesize{Z.\ Blur}}} &
        \includegraphics[width=0.29\linewidth]{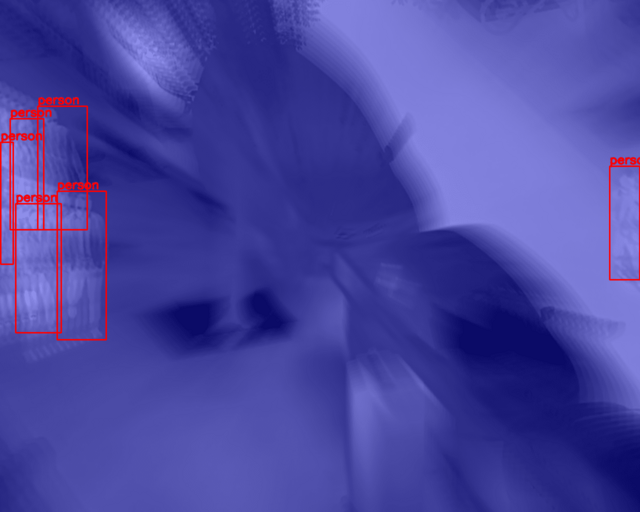} &
        \includegraphics[width=0.29\linewidth]{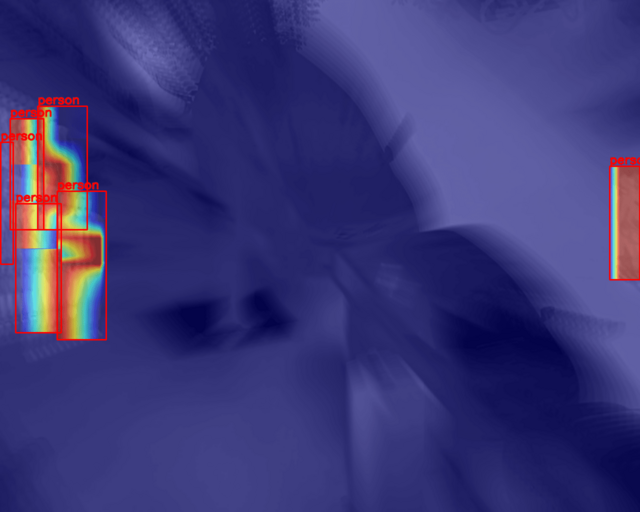} &
        \includegraphics[width=0.29\linewidth]{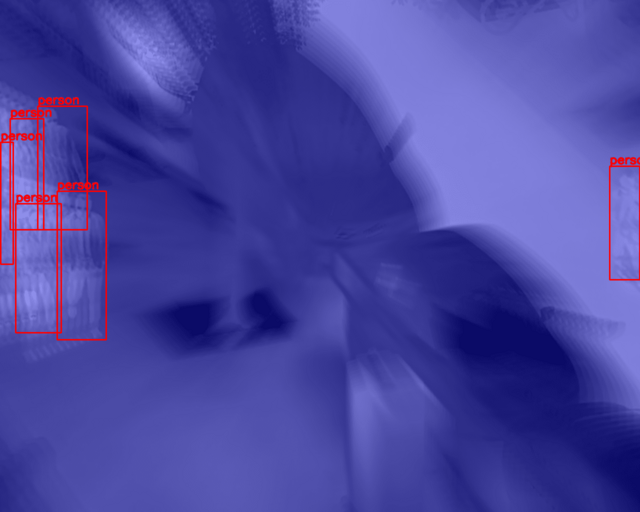} \\
    \end{tabular}

    \caption{\textbf{Activation map analysis on LLVIP-C.}
    Rows: impulse noise (top) and zoom blur (bottom; severity~5).
    Columns: Zero-shot (ZS), WiSE-OD$_{ZS}$, and FT. Ground-truth boxes in red.}
    \label{fig:activation_map}
    \vspace{-.5cm}
\end{figure}

In this section, we qualitatively analyze activation maps of Faster R-CNN under corruptions on LLVIP-C for the zero-shot model, WiSE-OD$_{ZS}$, and FT. 
As shown in~\fref{fig:activation_map}, Grad-CAM~\cite{selvaraju2017grad} highlights impulse noise (top) and zoom blur (bottom), with ground-truth boxes in red (additional examples are in the supplementary). 
While FT and zero-shot often fail to detect the person under heavy corruptions, WiSE-OD$_{ZS}$ activates more strongly on object regions, indicating greater robustness. 
Although performance varies across corruption types, tuning $\lambda$ can balance zero-shot and FT contributions for specific real-world needs. 
Overall, WiSE-OD$_{ZS}$ preserves more complete object regions across LLVIP-C and FLIR-C, especially under fog and low contrast, showing that ensembling retains complementary cues and stabilizes predictions.

\begin{table}[ht]
\caption{\textbf{Ablation of $\lambda$ over LLVIP-C and FLIR-C dataset for Faster R-CNN.} Where $\lambda = 0.0$ represents the zero-shot model, $\lambda = 0.5$ represents default WiSE-OD$_{ZS}$ and $\lambda = 1.0$ represents the fine-tuning model. For LLVIP-C, the severity level is $5$.}
\label{tab:ablation_lambda_wiseod}
    \centering
    \resizebox{1.0\columnwidth}{!}{%
    \begin{tabular}{lcccccc}

        \toprule

        {} & {} &  {} & \multirow{1}{*}[-0.7em]{\textbf{LLVIP-C}} & {} &  {} & {}\\

        \multirow{2}{*}[-1em]{} & \multirow{2}{*}[-1em]{$\theta(\lambda=0.0)$} & 
        \multirow{2}{*}[-1em]{$\theta(\lambda=0.2)$} &
        \multirow{2}{*}[-1em]{$\theta(\lambda=0.5)$} & \multirow{2}{*}[-1em]{$\theta(\lambda=0.8)$} & \multirow{2}{*}[-1em]{$\theta(\lambda=1.0)$} \\
        
        \cmidrule(lr){2-7}
        \addlinespace[10pt]

        \midrule

        Original & 71.21 ± 0.02 & 93.88 ± 0.28 & 96.06 ± 0.22 & 95.41 ± 0.60 & 93.68 ± 0.86 \\

        \midrule
        \rowcolor[HTML]{EFEFEF}
         Gaussian Noise & 59.24 ± 0.07 & 86.52 ± 0.40 & 86.68 ± 0.44 & 78.47 ± 3.43 & 67.46 ± 7.45 \\

        Shot Noise & 51.48 ± 0.14 & 83.86 ± 0.70 & 85.26 ± 0.50 & 76.42 ± 3.74 & 64.83 ± 7.79 \\

        \rowcolor[HTML]{EFEFEF}
        Impulse Noise & 56.62 ± 0.07 & 86.93 ± 0.65 & 88.54 ± 0.33 & 80.94 ± 2.70 & 71.32 ± 6.33 \\

        Defocus Blur & 47.90 ± 0.08 & 88.41 ± 0.31 & 89.74 ± 0.98 & 85.85 ± 2.55 & 80.48 ± 3.60 \\

        \rowcolor[HTML]{EFEFEF}
        Motion Blur & 26.39 ± 0.23 & 81.10 ± 0.44 & 86.81 ± 0.71 & 83.24 ± 1.70 & 78.32 ± 3.18 \\
        
        Zoom Blur & 02.47 ± 0.02 & 27.97 ± 0.62 & 22.83 ± 2.44 & 14.46 ± 1.82 & 11.18 ± 1.56 \\

        \rowcolor[HTML]{EFEFEF}
        Snow & 33.65 ± 0.01 & 67.67 ± 0.77 & 65.97 ± 1.90 & 38.50 ± 2.61 & 13.46 ± 4.45 \\
        
        Frost  & 33.25 ± 0.38 & 72.31 ± 0.23 & 75.87 ± 0.39 & 65.10 ± 0.57 & 47.32 ± 3.45 \\

        \rowcolor[HTML]{EFEFEF}
        Fog  & 59.60 ± 0.10 & 89.79 ± 0.43 & 84.51 ± 3.80 & 64.47 ± 9.62 & 50.90 ± 10.0 \\

        Brightness & 41.77 ± 0.03 & 82.38 ± 0.17 & 82.10 ± 1.20 & 62.81 ± 3.16 & 35.36 ± 6.97 \\

        \rowcolor[HTML]{EFEFEF}
        Contrast & 47.48 ± 0.04 & 50.59 ± 4.48 & 10.57 ± 3.82 & 00.77 ± 0.31 & 00.00 ± 0.00 \\

        Elastic transform  & 52.42 ± 0.18 & 89.92 ± 0.51 & 94.72 ± 0.07 & 94.32 ± 0.34 & 92.41 ± 0.93 \\

        \rowcolor[HTML]{EFEFEF}
        Pixelate & 03.95 ± 0.01 & 66.27 ± 3.14 & 85.06 ± 3.32 & 88.97 ± 2.35 & 87.69 ± 2.67 \\
        
        JPEG compression  & 57.22 ± 0.02 & 87.87 ± 0.89 & 92.59 ± 1.22 & 91.86 ± 1.38 & 88.93 ± 1.69 \\

        \midrule
        \rowcolor[HTML]{EFEFEF}
        mPC & 40.96 & 75.82 & 75.08 & 66.15 & 56.40 \\

        \midrule

        {} & {} &  {} & \multirow{1}{*}[-0.7em]{\textbf{FLIR-C}} & {} &  {} & {}\\

        \multirow{2}{*}[-1em]{} & \multirow{2}{*}[-1em]{$\theta(\lambda=0.0)$} & 
        \multirow{2}{*}[-1em]{$\theta(\lambda=0.2)$} &
        \multirow{2}{*}[-1em]{$\theta(\lambda=0.5)$} & \multirow{2}{*}[-1em]{$\theta(\lambda=0.8)$} & \multirow{2}{*}[-1em]{$\theta(\lambda=1.0)$} \\
        
        \cmidrule(lr){2-7}
        \addlinespace[10pt]

        \midrule

            Original & 65.52 ± 0.07 & 77.49 ± 0.10 & 82.20 ± 0.07 & 80.18 ± 0.11 & 77.57 ± 0.24 \\

		\midrule
		\rowcolor[HTML]{EFEFEF}
		Gaussian Noise & 31.21 ± 0.29 & 42.62 ± 2.92 & 42.49 ± 4.48 & 34.85 ± 3.85 & 28.07 ± 2.91  \\

		Shot Noise & 25.26 ± 0.12 & 33.91 ± 2.98 & 30.45 ± 3.96 & 21.88 ± 2.93 & 15.73 ± 2.05 \\

		\rowcolor[HTML]{EFEFEF}
		Impulse Noise & 17.69 ± 0.03 & 24.85 ± 2.26 & 22.51 ± 2.78 & 16.96 ± 2.17 & 13.22 ± 2.27  \\

		Defocus Blur & 25.32 ± 0.22 & 44.57 ± 2.40 & 54.08 ± 1.74 & 55.00 ± 0.98 & 52.47 ± 0.99 \\

		\rowcolor[HTML]{EFEFEF}
		Motion Blur & 25.01 ± 0.25 & 40.63 ± 2.17 & 51.03 ± 2.16 & 53.85 ± 2.19 & 51.71 ± 2.12  \\

		Zoom Blur & 08.98 ± 0.05 & 13.72 ± 0.97 & 16.93 ± 1.00 & 18.32 ± 1.09 & 17.97 ± 0.90 &  \\

		\rowcolor[HTML]{EFEFEF}
		Snow & 09.84 ± 0.14 & 14.55 ± 2.07 & 13.94 ± 2.66 & 10.36 ± 2.48 & 07.86 ± 2.01 \\

		Frost  & 21.96 ± 0.50 & 33.37 ± 2.39 & 37.97 ± 3.63 & 36.47 ± 4.00 & 33.87 ± 4.69  \\

		\rowcolor[HTML]{EFEFEF}
		Fog  & 56.36 ± 0.28 & 72.17 ± 0.86 & 78.68 ± 1.24 & 78.11 ± 1.49 & 73.61 ± 0.06 \\

		Brightness & 64.41 ± 0.26 & 75.68 ± 0.19 & 79.72 ± 0.35 & 77.79 ± 1.24 & 75.18 ± 0.99 \\

		\rowcolor[HTML]{EFEFEF}
		Contrast & 54.59 ± 0.04 & 71.38 ± 0.95 & 78.36 ± 1.06 & 78.02 ± 1.09 & 75.47 ± 1.29 \\

		Elastic transform  & 41.88 ± 0.24 & 63.89 ± 0.37 & 73.39 ± 0.40 & 72.51 ± 0.58 & 69.68 ± 1.15 &  \\

		\rowcolor[HTML]{EFEFEF}
		Pixelate & 38.67 ± 0.11 & 55.23 ± 2.37 & 61.12 ± 3.13 & 58.87 ± 4.58 & 54.91 ± 6.21  \\

		JPEG compression  & 50.24 ± 0.14 & 63.21 ± 0.56 & 66.65 ± 0.89 & 63.27 ± 2.36 & 57.55 ± 3.04  \\

		\midrule
		\rowcolor[HTML]{EFEFEF}
		mPC & 33.67 & 46.41 & 50.52 & 48.30 & 44.80 \\

    \bottomrule
        
    \end{tabular}
    }
\vspace{-.0cm}
\end{table}

\subsection{Real-world OOD Scenario}

\begin{figure}[!h]
    \centering

    \setlength{\tabcolsep}{1pt}  
    \renewcommand{\arraystretch}{0.25} 
    
    \begin{tabular}{cccccc}
        & \multicolumn{1}{c}{\textbf{\footnotesize{RGB}}} & \multicolumn{1}{c}{\textbf{\footnotesize{IR}}} & \multicolumn{1}{c}{\textbf{\footnotesize{ZS}}} & \multicolumn{1}{c}{\textbf{\footnotesize{FT}}} & \multicolumn{1}{c}{\textbf{\footnotesize{WiSE‑OD}}} \\

        \rotatebox{90}{\textbf{\; \footnotesize{Fog}}} &
        \includegraphics[width=0.19\linewidth]{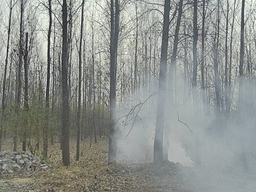} &
        \includegraphics[width=0.19\linewidth]{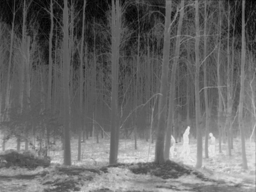} &
        \includegraphics[width=0.19\linewidth]{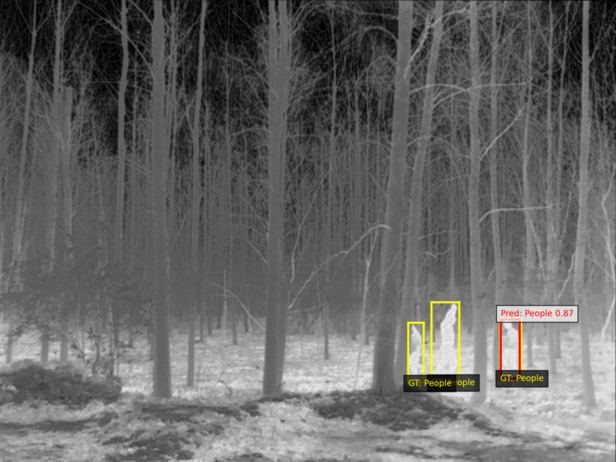} &
        \includegraphics[width=0.19\linewidth]{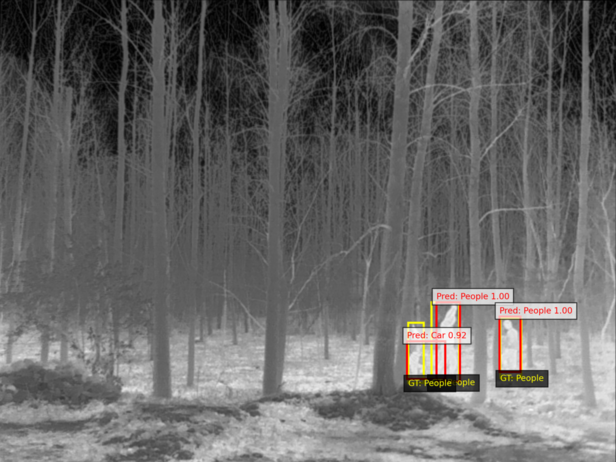} &
        \includegraphics[width=0.19\linewidth]{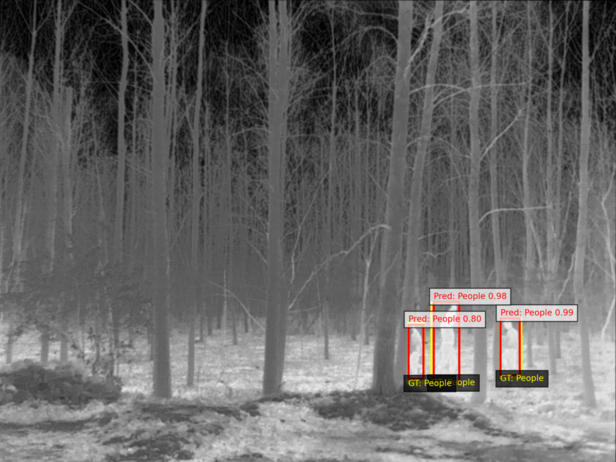} \\

        \rotatebox{90}{\textbf{\; \footnotesize{Night}}} &
        \includegraphics[width=0.19\linewidth]{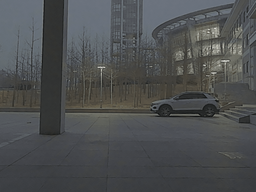} &
        \includegraphics[width=0.19\linewidth]{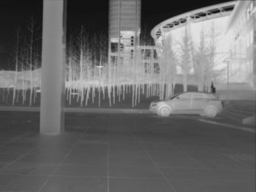} &
        \includegraphics[width=0.19\linewidth]{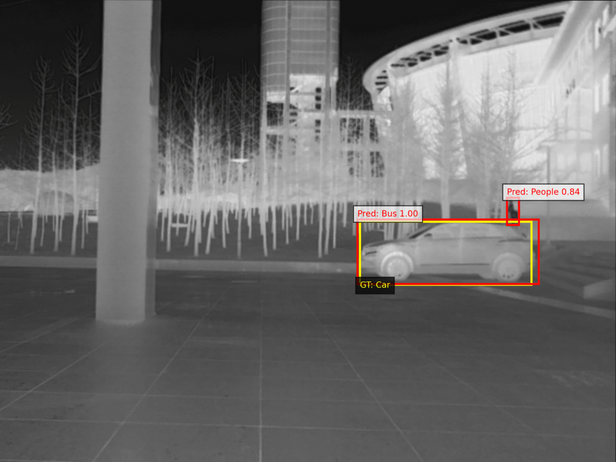} &
        \includegraphics[width=0.19\linewidth]{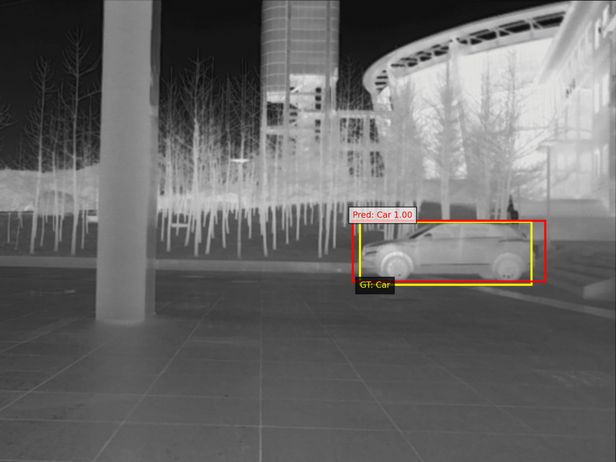} &
        \includegraphics[width=0.19\linewidth]{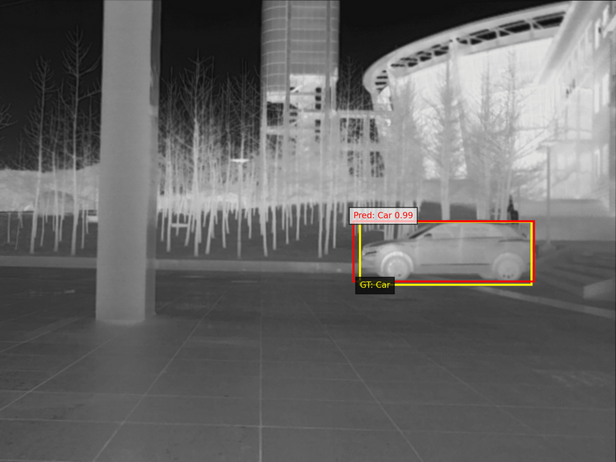} \\
        
        \rotatebox{90}{\textbf{\;\; \footnotesize{Rain}}} &
        \includegraphics[width=0.19\linewidth]{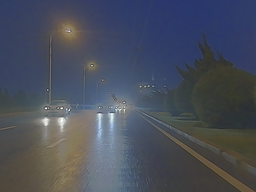} &
        \includegraphics[width=0.19\linewidth]{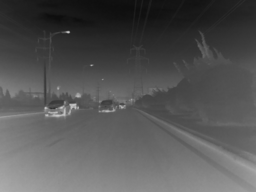} &
        \includegraphics[width=0.19\linewidth]{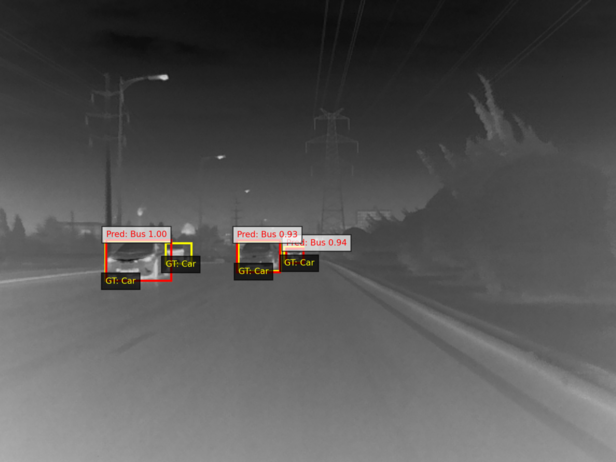} &
        \includegraphics[width=0.19\linewidth]{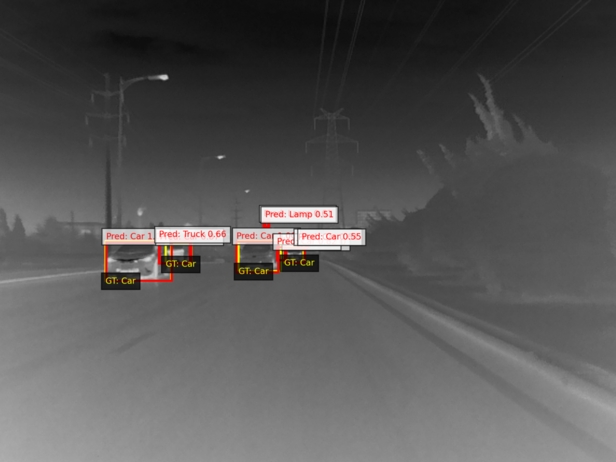} &
        \includegraphics[width=0.19\linewidth]{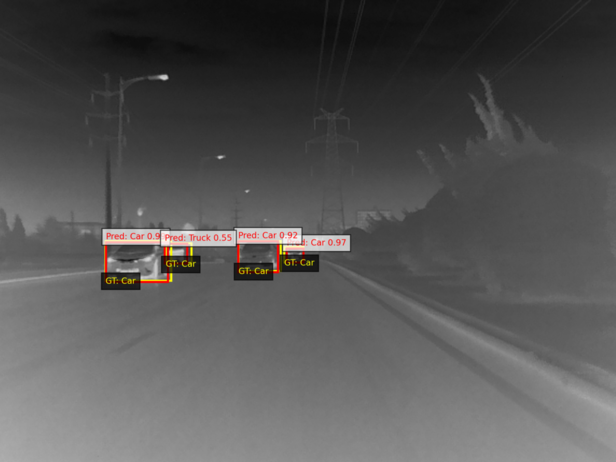} \\
        
        \rotatebox{90}{\textbf{\; \footnotesize{Indoor}}} &
        \includegraphics[width=0.19\linewidth]{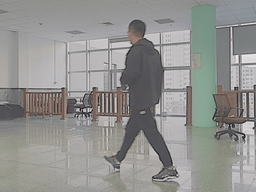} &
        \includegraphics[width=0.19\linewidth]{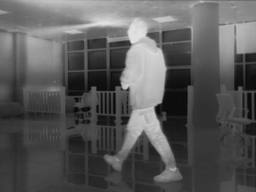} &
        \includegraphics[width=0.19\linewidth]{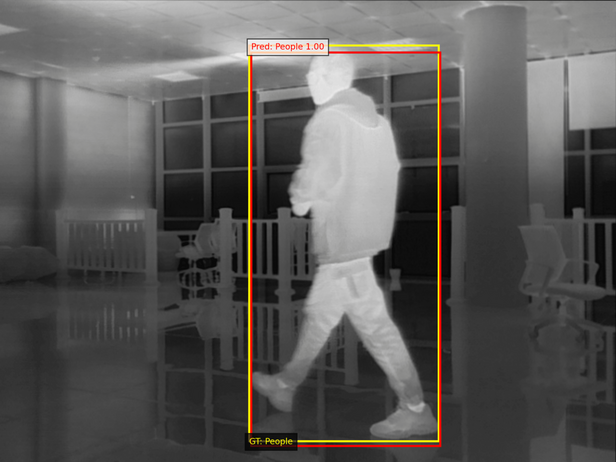} &
        \includegraphics[width=0.19\linewidth]{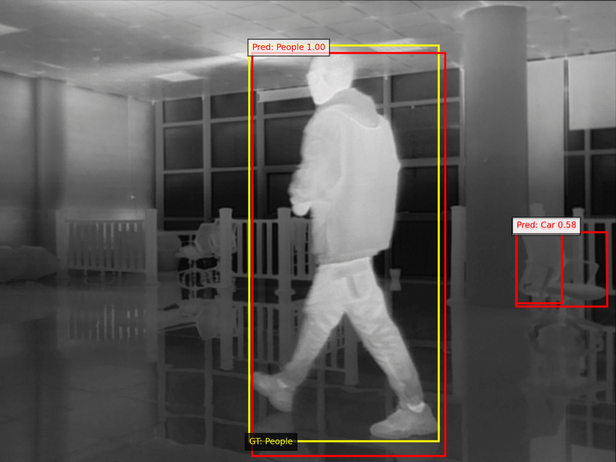} &
        \includegraphics[width=0.19\linewidth]{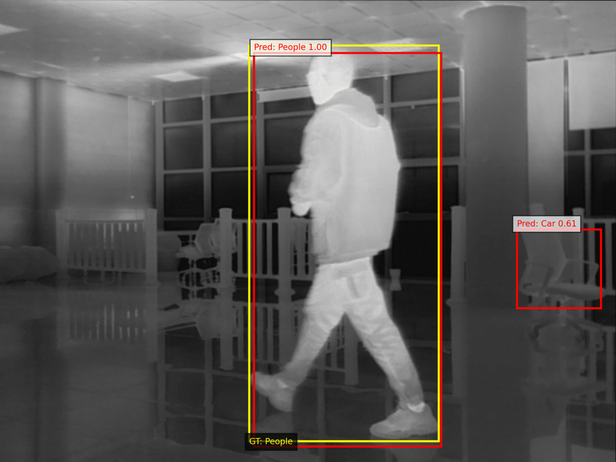} \\
    \end{tabular}

    \caption{\textbf{Qualitative comparison on M3FD under adverse conditions.} 
    Rows: fog, night, rain, indoor. Columns: RGB, IR, Zero-Shot (ZS), Fine-Tuning (FT), and WiSE-OD$_{ZS}$.}

    \label{fig:all_conditions_grid}
    \vspace{-.2cm}
\end{figure}

\noindent In this section, we validate robustness on the M3FD dataset~\cite{liu2022target}, a multi-scenario, multi-modality benchmark designed for fusing infrared and visible modalities in object detection. M3FD contains diverse real-world IR scenes across \textit{fog, rain, indoor, day, and night conditions}, providing challenging evaluation scenarios for robustness studies. Models are trained on day images and evaluated on the other splits, simulating a realistic OOD deployment. As shown in Tab.~\ref{tab:wiseod_splits_reordered_avg_detectors}, WiSE-OD consistently improves over FT and ZS: \textbf{+10.8} mAP (raining), \textbf{+7.2} (fog), \textbf{+7.1} (night), and \textbf{+0.2} (indoor), for an average gain of \textbf{+6.3}. These gains are achieved without retraining or access to target data, highlighting WiSE-OD’s practicality. Importantly, the consistent improvements on M3FD demonstrate that our approach 
is not limited to synthetic benchmarks (LLVIP-C, FLIR-C), 
but generalizes to real-world distribution shifts.


\begin{table}[!h]
\centering
\small
\setlength{\tabcolsep}{2.2pt}
\renewcommand{\arraystretch}{0.9}
\caption{\textbf{Detection performance of four detectors across M3FD splits.}
Models are trained on IR day images; WiSE-OD is applied with $\lambda{=}0.5$.
Results are reported for Rain, Night, Fog, Indoor, and average across conditions.}
\label{tab:wiseod_splits_reordered_avg_detectors}

\resizebox{1.0\linewidth}{!}{
\begin{tabular}{l ccc ccc ccc ccc ccc}
\toprule
\multirow{2}{*}{\textbf{Method}} &
\multicolumn{3}{c}{\textbf{Rain}} &
\multicolumn{3}{c}{\textbf{Night}} &
\multicolumn{3}{c}{\textbf{Fog}} &
\multicolumn{3}{c}{\textbf{Indoor}} &
\multicolumn{3}{c}{\textbf{Avg.}} \\
& AP$_{50}$ & AP$_{75}$ & mAP & AP$_{50}$ & AP$_{75}$ & mAP & AP$_{50}$ & AP$_{75}$ & mAP & AP$_{50}$ & AP$_{75}$ & mAP & AP$_{50}$ & AP$_{75}$ & mAP \\
\cmidrule(lr){2-4}\cmidrule(lr){5-7}\cmidrule(lr){8-10}\cmidrule(lr){11-13}\cmidrule(lr){14-16}
\multicolumn{16}{c}{\textbf{Faster R-CNN}} \\
\midrule
\rowcolor[HTML]{EFEFEF}
Zero-shot   & 13.1 &  6.1 &  6.7 & 14.3 &  8.3 &  8.2 & 87.4 & 81.7 & 68.4 &  5.8 &  5.7 &  4.4 & 30.2 & 25.4 & 22.0 \\
Fine-tuning & 28.0 & 11.8 & 13.8 & \textbf{47.8} & \textbf{30.9} & \textbf{29.4} & 93.0 & 77.5 & 66.2 & \textbf{59.8} & 32.8 & 33.2 & 57.2 & 38.3 & 35.7 \\
\rowcolor[HTML]{EFEFEF}
WiSE-OD     & \textbf{29.4} & \textbf{13.2} & \textbf{14.9} & 43.3 & 30.3 & 27.9 & \textbf{97.2} & \textbf{84.5} & \textbf{74.2} & 59.2 & \textbf{36.2} & \textbf{33.9} & \textbf{57.3} & \textbf{41.1} & \textbf{37.7} \\
\midrule
\multicolumn{16}{c}{\textbf{FCOS}} \\
\midrule
\rowcolor[HTML]{EFEFEF}
Zero-shot   & 10.6 &  5.5 &  5.6 & 12.3 &  7.2 &  7.1 & 83.4 & 74.8 & 64.0 &  4.3 &  4.0 &  3.3 & 27.7 & 22.9 & 20.0 \\
Fine-tuning & 21.7 &  9.5 & 10.7 & \textbf{42.3} & \textbf{28.3} & \textbf{27.9} & 93.5 & 78.7 & 65.9 & \textbf{61.6} & 32.1 & 31.7 & \textbf{54.8} & 37.2 & 34.1 \\
\rowcolor[HTML]{EFEFEF}
WiSE-OD     & \textbf{25.5} & \textbf{11.7} & \textbf{13.0} & 34.7 & 22.0 & 21.6 & \textbf{96.0} & \textbf{86.8} & \textbf{73.0} & 56.4 & \textbf{35.2} & \textbf{33.7} & 53.2 & \textbf{38.9} & \textbf{35.3} \\
\midrule
\multicolumn{16}{c}{\textbf{RetinaNet}} \\
\midrule
\rowcolor[HTML]{EFEFEF}
Zero-shot   & 10.6 &  4.9 &  5.4 & 12.9 &  7.7 &  7.4 & 91.4 & 84.6 & 69.6 &  9.0 &  6.9 &  6.3 & 31.0 & 26.0 & 22.2 \\
Fine-tuning & \textbf{29.3} & 11.5 & 13.5 & \textbf{45.0} & \textbf{28.8} & \textbf{27.3} & 96.5 & 82.8 & 67.7 & 55.3 & \textbf{34.4} & \textbf{32.9} & \textbf{56.5} & 39.4 & 35.3 \\
\rowcolor[HTML]{EFEFEF}
WiSE-OD     & 27.2 & \textbf{12.7} & \textbf{13.7} & 40.1 & 27.1 & 25.8 & \textbf{97.6} & \textbf{89.9} & \textbf{73.6} & \textbf{56.7} & 33.6 & 32.8 & 55.4 & \textbf{40.8} & \textbf{36.5} \\
\midrule
\multicolumn{16}{c}{\textbf{YOLOv8n}} \\
\midrule
\rowcolor[HTML]{EFEFEF}
Zero-shot   & 28.4 & 22.4 & 20.5 & 37.3 & 34.6 & 31.8 & 88.9 & 84.9 & 73.8 & 22.1 & 22.1 & 19.3 & 44.2 & 41.0 & 36.4 \\
Fine-tuning & 29.2 & 23.1 & 21.7 & 49.1 & 37.7 & 36.6 & 92.3 & 83.6 & 70.3 & \textbf{55.1} & \textbf{43.4} & \textbf{38.5} & 56.4 & 46.9 & 41.8 \\
\rowcolor[HTML]{EFEFEF}
WiSE-OD     & \textbf{42.2} & \textbf{35.5} & \textbf{32.5} & \textbf{55.5} & \textbf{46.1} & \textbf{43.7} & \textbf{95.3} & \textbf{90.9} & \textbf{77.5} & 48.8 & 43.6 & 38.7 & \textbf{60.4} & \textbf{54.0} & \textbf{48.1} \\
\bottomrule
\end{tabular}}
\vspace{-.5cm}
\end{table}

\subsection{WiSE-OD\texorpdfstring{$_{ZS}$}{ ZS}: Ablation study on \texorpdfstring{$\lambda$}{lambda}}

In this section, we extensively conducted studies about the $\lambda$ value to combine the zero-shot RGB COCO pre-training weights of Faster R-CNN, FCOS, and RetinaNet with the FT IR under the respective datasets LLVIP-C and FLIR-C. Evaluating the performance of such weight ensembling WiSE-OD$_{ZS}$ under the different corruption settings. Here, in the main manuscript, we show the results for Faster R-CNN in~\tref{tab:ablation_lambda_wiseod} for some values of $\lambda$, and we provide the detailed ablation and additional results in the supplementary material. It is important to mention that in the rest of the main manuscript, $\lambda$ was fixed to $0.5$, while here, we wanted to further investigate the potential of the WiSE-OD$_{ZS}$ over the different corruptions. In~\tref{tab:ablation_lambda_wiseod}, the best results for in-domain performance were with $\lambda = 0.5$, while the best out-of-domain was $\lambda=0.2$, which shows that for Faster R-CNN the zero-shot model can bring robustness for the model, but some corruptions, such as pixelate, are better when the $\lambda$ is higher. 

We observe that $\lambda = 0.5$ consistently provides a favorable trade-off, even for FLIR-C, it was the best ID. For LLVIP-C, WiSE-OD achieved the best mean performance under corruption ($\lambda = 0.2$, mPC $75.82$), outperforming both the pure fine-tuned model ($\lambda = 1.0$, mPC $56.40$) and the zero-shot model ($\lambda = 0.0$, mPC $40.96$). Notably, $\lambda = 0.5$ performs best under heavy corruptions such as Gaussian noise, Fog, and Brightness shifts, scenarios where both FT and ZS individually struggle. For example, under Fog and Brightness, $\lambda = 0.5$ yields $84.51$ and $82.10$, respectively, while $\lambda = 1.0$ achieves only $50.90$ and $35.36$. This highlights the benefit of WiSE-OD in preserving complementary robustness features from both models. Interestingly, $\lambda = 0.8$ performs well in several cases but shows more variability, suggesting that moderate ensembling (rather than heavily biasing toward FT) is more robust under distribution shifts. These results justify the use of $\lambda = 0.5$ as a robust default and motivate future work on adaptive $\lambda$ selection strategies.

\section{Conclusion}
\label{sec:conclusion}

In this work, we presented a new benchmark for IR OD robustness based on the work of~\citet{hendrycks2019benchmarking}, targeting traditional IR datasets such as LLVIP and FLIR, as well as real-world shifts such as M3FD. Our new benchmark is a challenging setting for IR robustness with the introduction of LLVIP-C and FLIR-C. Furthermore, we conducted an extensive study of different robust fine-tuning strategies over our proposed benchmark and in real-world OOD data. Additionally, we presented the WiSE-OD method and its variants WiSE-OD$_{ZS}$ and WiSE-OD$_{LP}$, both of which surpass traditional robustness strategies while also increasing in-domain performance across different detectors, such as Faster R-CNN, FCOS, RetinaNet, and YOLOv8. Our extensive study shows that our simple WiSE-OD strategy can mitigate performance drops without any additional training cost.


\noindent\textbf{Main limitations.} WiSE-OD assumes access to both a zero-shot and a fine-tuned (or linearly probed) model, which can be infeasible in constrained deployments. A fixed mixing coefficient ($\lambda = 0.5$) works well on average but is not uniformly optimal; we deliberately avoid tuning $\lambda$ on held-out target data to reflect deployment realities, which may leave corruption-specific gains unrealized. The method also inherits weaknesses from its base models; for instance, if either the zero-shot or FT model underperforms, ensemble gains are limited.

\noindent \textbf{Failure cases.} Robustness degrades under extreme corruptions, like severe snow, heavy blur, very low brightness, and especially \emph{low contrast} with FLIR-C most affected. In such scenes, both base models often miss or fragment objects, and the ensemble propagates these errors, yielding uncertain activations and unstable boxes. Mitigation likely requires corruption-aware ensembling or adaptive $\lambda$.

\noindent \textbf{Future work.} A natural extension is \emph{adaptive} weight-space ensembling: predict $\lambda$ per image or corruption using lightweight signals (e.g., confidence/entropy, or a small gating network). Integrate WiSE-OD with domain-generalization to better handle unseen IR conditions without target labels. Finally, evaluate additional sensors (depth, multispectral) to assess generality beyond IR.

\section*{Acknowledgments} This work was supported by Distech Controls Inc., the Natural Sciences and Engineering Research Council of Canada, the Digital Research Alliance of Canada, and Mitacs. \\

\maketitlesupplementary

In this supplementary material, we provide additional information to reproduce our work~\footnote{Our code is available at: \url{https://github.com/heitorrapela/wiseod.git}}. This supplementary material is divided into the following sections: Dataset Visualization (Section~\ref{sec:dataset Visualization}), Detection performance per corruption (Section~\ref{sec:detection_performance_per_corruption}), Activation map analysis (Section~\ref{sec:gradcam}), WiSE-OD$_{ZS}$: Ablation study on $\lambda$ (Section~\ref{sec:ablation_lambdas}) and Performance over different corruption levels (Section~\ref{sec:diff_corruption_levels}).

\section{Dataset Visualization}
\label{sec:dataset Visualization}

\begin{figure*}
\captionsetup[subfigure]{labelformat=empty}
\centering
\begin{subfigure}[t]{0.50\columnwidth}
    \caption{Gaussian Noise}
    \includegraphics[width=\columnwidth]{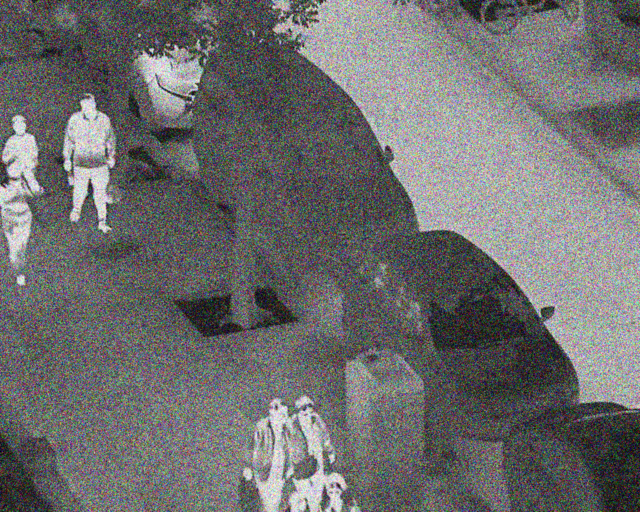}
\end{subfigure}
\begin{subfigure}[t]{0.50\columnwidth}
    \caption{Shot Noise}
    \includegraphics[width=\columnwidth]{imgs/corruptions/llvip/llvip_shot_noise_severity5.png}
\end{subfigure}
\begin{subfigure}[t]{0.50\columnwidth}
    \caption{Impulse Noise}
    \includegraphics[width=\columnwidth]{imgs/corruptions/llvip/llvip_impulse_noise_severity5.png}
\end{subfigure}

\begin{subfigure}[t]{0.50\columnwidth}
    \caption{Defocus Blur}
    \includegraphics[width=\columnwidth]{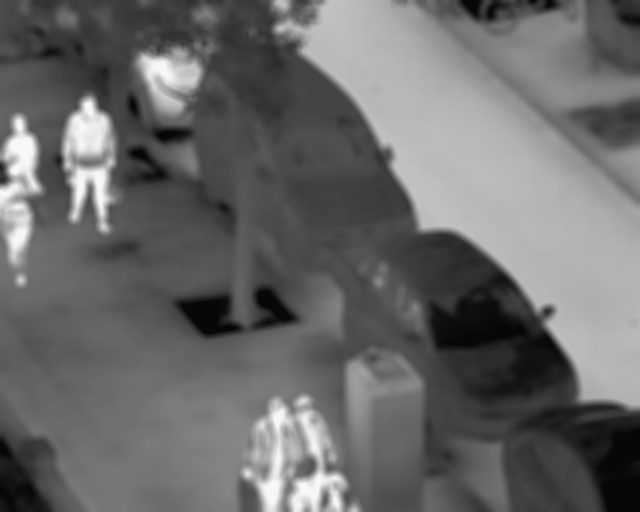}
\end{subfigure}
\begin{subfigure}[t]{0.50\columnwidth}
    \caption{Motion Blur}
    \includegraphics[width=\columnwidth]{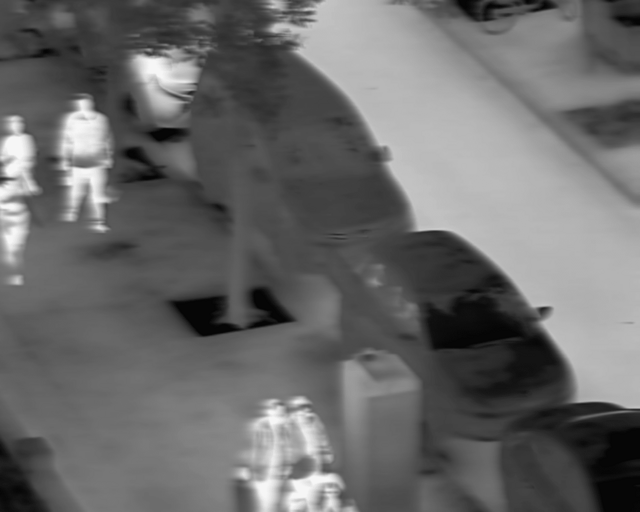}
\end{subfigure}
\begin{subfigure}[t]{0.50\columnwidth}
    \caption{Zoom Blur}
    \includegraphics[width=\columnwidth]{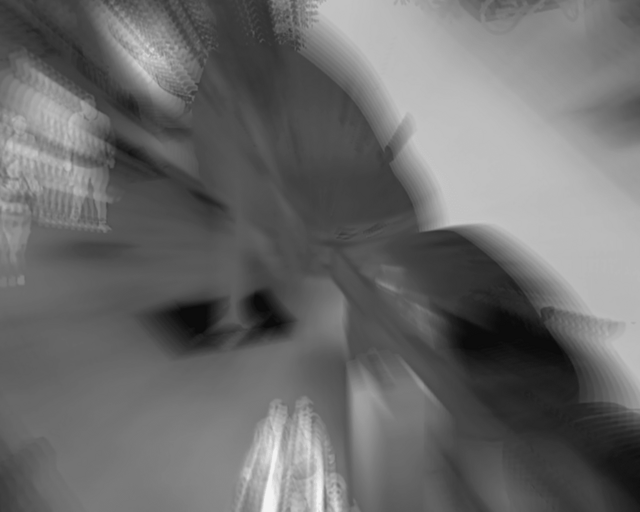}
\end{subfigure}

\begin{subfigure}[t]{0.50\columnwidth}
    \caption{Snow}
    \includegraphics[width=\columnwidth]{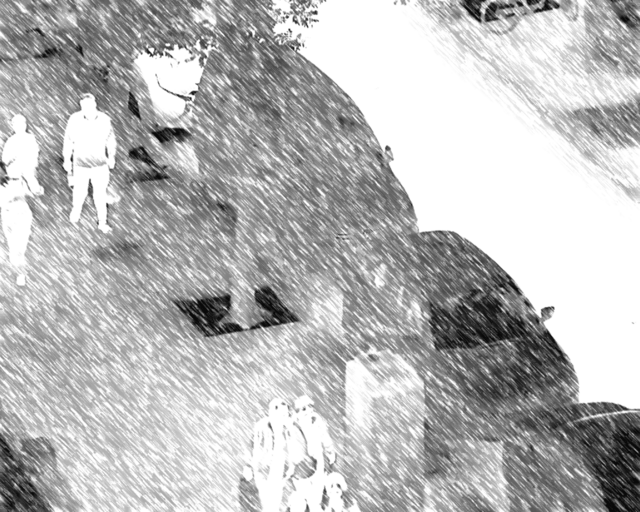}
\end{subfigure}
\begin{subfigure}[t]{0.50\columnwidth}
    \caption{Frost}
    \includegraphics[width=\columnwidth]{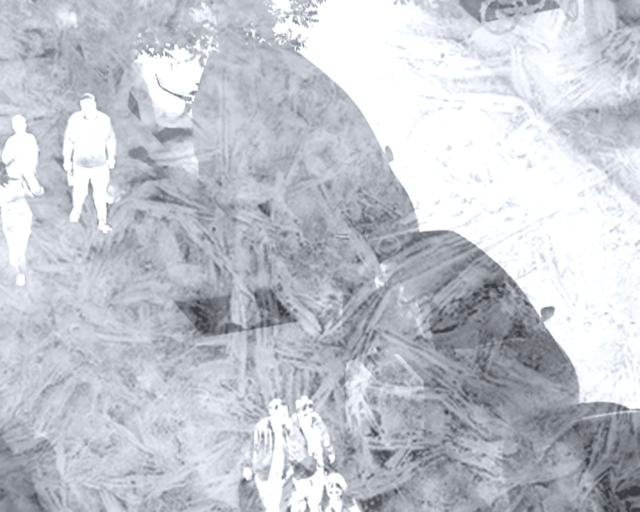}
\end{subfigure}
\begin{subfigure}[t]{0.50\columnwidth}
    \caption{Fog}
    \includegraphics[width=\columnwidth]{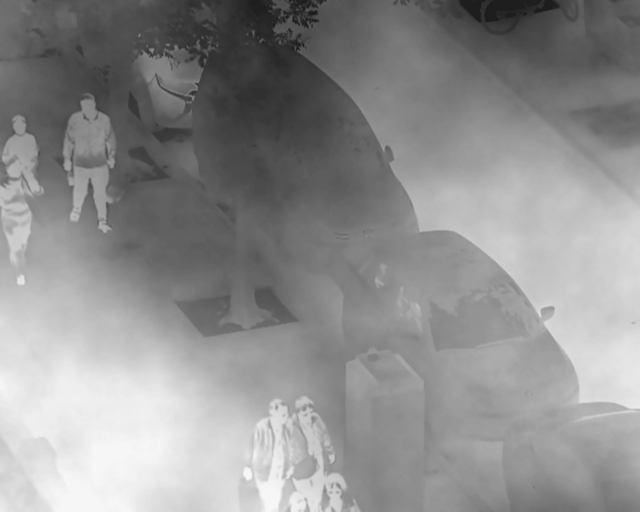}
\end{subfigure}

\begin{subfigure}[t]{0.50\columnwidth}
    \caption{Brightness}
    \includegraphics[width=\columnwidth]{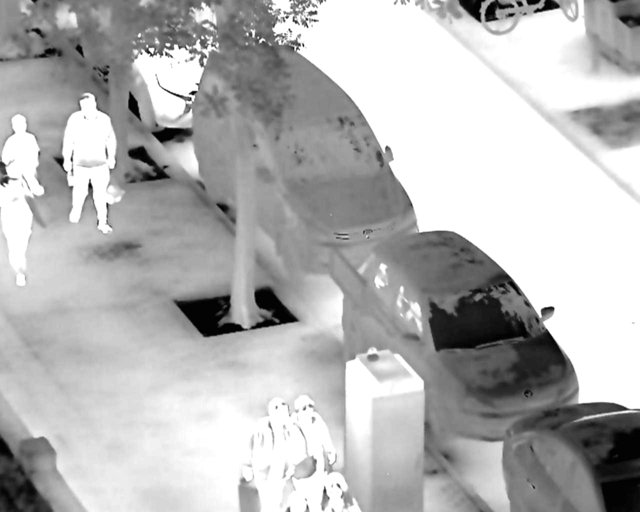}
\end{subfigure}
\begin{subfigure}[t]{0.50\columnwidth}
    \caption{Contrast}
    \includegraphics[width=\columnwidth]{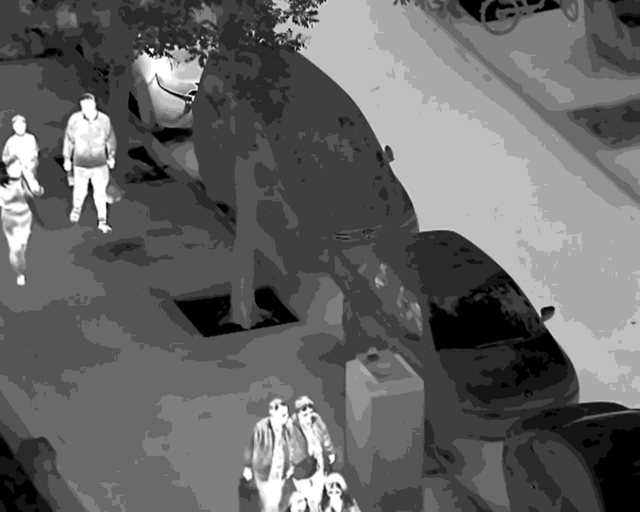}
\end{subfigure}
\begin{subfigure}[t]{0.50\columnwidth}
    \caption{Elastic Transform}
    \includegraphics[width=\columnwidth]{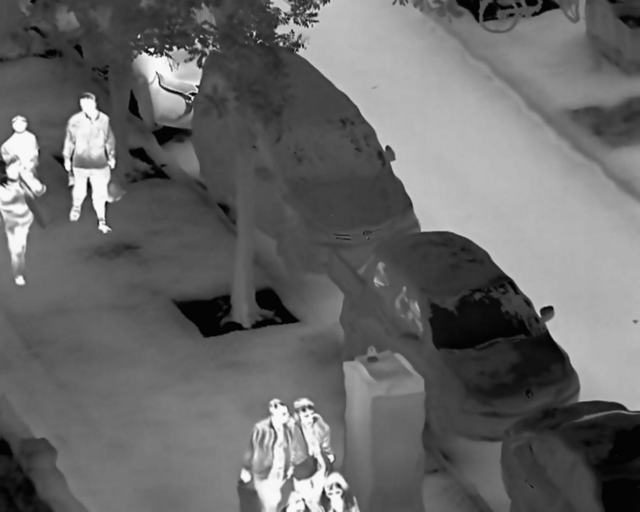}
\end{subfigure}

\begin{subfigure}[t]{0.50\columnwidth}
    \caption{Pixelate}
    \includegraphics[width=\columnwidth]{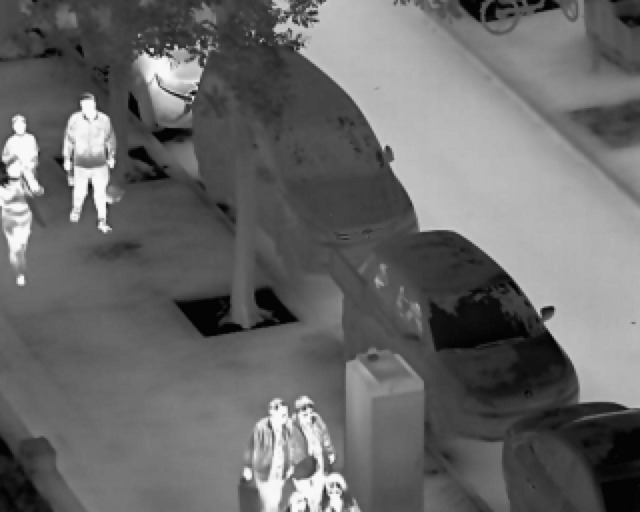}
\end{subfigure}
\begin{subfigure}[t]{0.50\columnwidth}
    \caption{JPEG Compression}
    \includegraphics[width=\columnwidth]{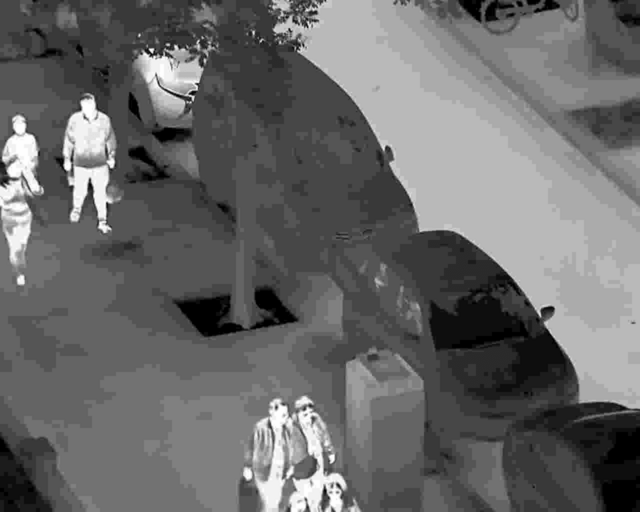}
\end{subfigure}

\caption{\textbf{All the 14 corruptions types from~\cite{hendrycks2019benchmarking}, adapted to our LLVIP-C benchmark with severity of 5.}}
\label{fig:llvip_images}
\end{figure*}

\begin{figure*}
\captionsetup[subfigure]{labelformat=empty}
\centering
\begin{subfigure}[t]{0.50\columnwidth}
    \caption{Gaussian Noise}
    \includegraphics[width=\columnwidth]{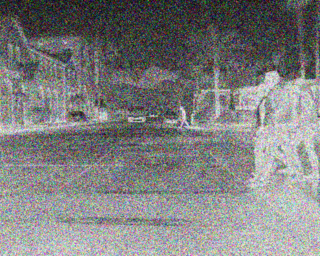}
\end{subfigure}
\begin{subfigure}[t]{0.50\columnwidth}
    \caption{Shot Noise}
    \includegraphics[width=\columnwidth]{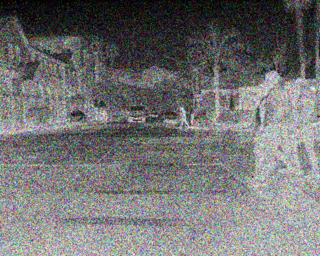}
\end{subfigure}
\begin{subfigure}[t]{0.50\columnwidth}
    \caption{Impulse Noise}
    \includegraphics[width=\columnwidth]{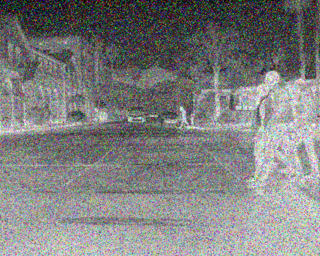}
\end{subfigure}

\begin{subfigure}[t]{0.50\columnwidth}
    \caption{Defocus Blur}
    \includegraphics[width=\columnwidth]{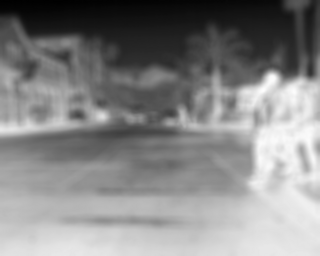}
\end{subfigure}
\begin{subfigure}[t]{0.50\columnwidth}
    \caption{Motion Blur}
    \includegraphics[width=\columnwidth]{imgs/corruptions/flir/flir_motion_blur_severity5.png}
\end{subfigure}
\begin{subfigure}[t]{0.50\columnwidth}
    \caption{Zoom Blur}
    \includegraphics[width=\columnwidth]{imgs/corruptions/flir/flir_zoom_blur_severity5.png}
\end{subfigure}

\begin{subfigure}[t]{0.50\columnwidth}
    \caption{Snow}
    \includegraphics[width=\columnwidth]{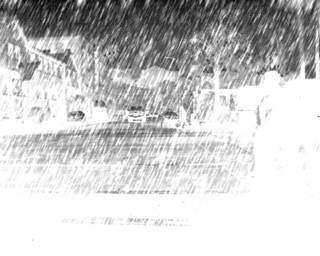}
\end{subfigure}
\begin{subfigure}[t]{0.50\columnwidth}
    \caption{Frost}
    \includegraphics[width=\columnwidth]{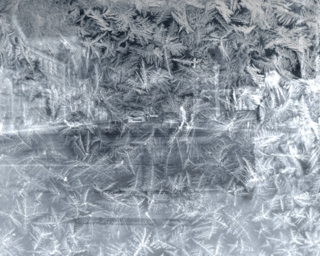}
\end{subfigure}
\begin{subfigure}[t]{0.50\columnwidth}
    \caption{Fog}
    \includegraphics[width=\columnwidth]{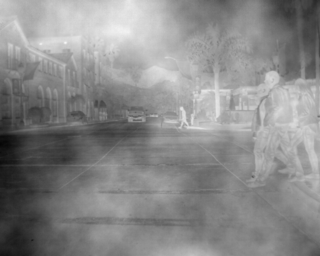}
\end{subfigure}

\begin{subfigure}[t]{0.50\columnwidth}
    \caption{Brightness}
    \includegraphics[width=\columnwidth]{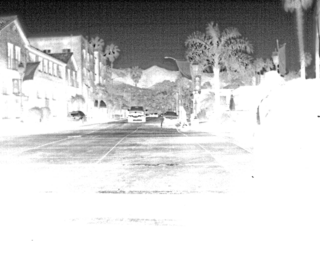}
\end{subfigure}
\begin{subfigure}[t]{0.50\columnwidth}
    \caption{Contrast}
    \includegraphics[width=\columnwidth]{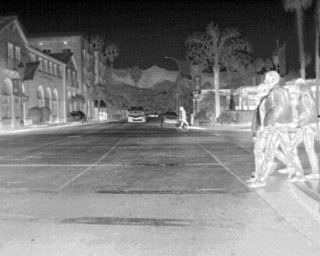}
\end{subfigure}
\begin{subfigure}[t]{0.50\columnwidth}
    \caption{Elastic Transform}
    \includegraphics[width=\columnwidth]{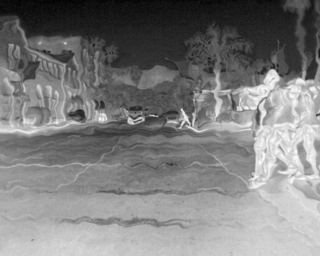}
\end{subfigure}

\begin{subfigure}[t]{0.50\columnwidth}
    \caption{Pixelate}
    \includegraphics[width=\columnwidth]{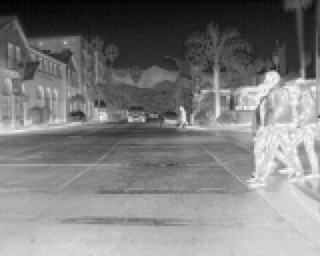}
\end{subfigure}
\begin{subfigure}[t]{0.50\columnwidth}
    \caption{JPEG Compression}
    \includegraphics[width=\columnwidth]{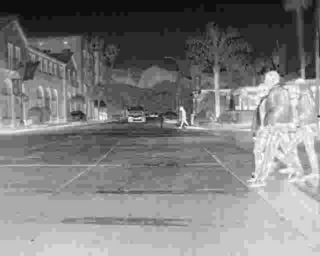}
\end{subfigure}

\caption{\textbf{All the 14 corruptions types from~\cite{hendrycks2019benchmarking}, adapted to our FLIR-C benchmark with severity of 5.}}
\label{fig:flir_images}
\end{figure*}

In this section, we provide additional visualization of each corruption for both datasets: LLVIP-C in~\fref{fig:llvip_images} and FLIR-C in~\fref{fig:flir_images}. Here, we wanted to highlight how strong the severity level of $5$ is for FLIR-C, which can destroy the whole image, for instance, for the Frost corruption.


\section{Detection performance per corruption}
\label{sec:detection_performance_per_corruption}

In this section, we expanded our evaluation of the benchmark per corruption. In the main manuscript, we provided the results per corruption for Faster R-CNN with a severity level of $5$ for LLVIP-C. Here, we provide the additional results for the FCOS (\tref{tab:fcos_llvip_per_corruption}) and RetinaNet (\tref{tab:retinanet_llvip_per_corruption}) for LLVIP-C, FCOS (\tref{tab:fcos_flir_per_corruption}) and RetinaNet (\tref{tab:retinanet_flir_per_corruption}) for FLIR-C with severity level of $2$.

\begin{table}[!h]
\caption{\textbf{AP$_{50}$ performance over the perturbations for LLVIP-C with severity level $5$ for FCOS.}}
\label{tab:fcos_llvip_per_corruption}
    \centering
    \resizebox{\columnwidth}{!}{%
    \begin{tabular}{lcccccccc}

        \toprule

        {} & {} & \multicolumn{4}{c}{\multirow{2}{*}[-0.1em]{\textbf{LLVIP-C}}} \\

        \multirow{2}{*}[-1em]{} & \multirow{2}{*}[-1em]{Zero-Shot} & 
        \multirow{2}{*}[-1em]{FT} &
        \multirow{2}{*}[-1em]{LP} & \multirow{2}{*}[-1em]{LP-FT} & \multirow{2}{*}[-1em]{WiSE-OD$_{ZS}$} & \multirow{2}{*}[-1em]{WiSE-OD$_{LP}$} \\

        \cmidrule(lr){2-7}
        \addlinespace[10pt]     
        \midrule
        
        \rowcolor[HTML]{EFEFEF}
        Gaussian Noise    & 55.03 ± 0.07 & 62.32 ± 5.96 & 73.63 ± 0.09 & 64.46 ± 0.42 & 83.68 ± 0.67 & 81.78 ± 0.08 \\

        Shot Noise        & 43.83 ± 0.36 & 58.75 ± 5.48 & 66.05 ± 1.98 & 59.34 ± 0.08 & 80.55 ± 0.81 & 80.00 ± 0.59 \\

        \rowcolor[HTML]{EFEFEF}
        Impulse Noise     & 51.60 ± 0.17 & 65.26 ± 5.31 & 68.98 ± 0.04 & 65.06 ± 0.21 & 85.86 ± 0.53 & 85.21 ± 0.95 \\

        Defocus Blur      & 33.67 ± 0.00 & 80.30 ± 1.90 & 78.14 ± 0.00 & 76.47 ± 0.00 & 87.74 ± 0.68 & 88.77 ± 0.00 \\

        \rowcolor[HTML]{EFEFEF}
        Motion Blur       & 15.09 ± 0.34 & 78.87 ± 1.09 & 70.67 ± 0.21 & 67.88 ± 0.25 & 84.32 ± 0.76 & 84.06 ± 0.01 \\

        Zoom Blur         & 01.28 ± 0.00 & 20.13 ± 2.95 & 18.75 ± 0.87 & 14.01 ± 0.00 & 31.47 ± 2.96 & 30.27 ± 0.00 \\

        \rowcolor[HTML]{EFEFEF}
        Snow              & 37.37 ± 0.25 & 33.17 ± 2.74 & 45.72 ± 0.31 & 42.42 ± 0.25 & 69.72 ± 1.83 & 70.94 ± 0.38 \\

        Frost             & 35.33 ± 0.07 & 60.18 ± 1.17 & 53.68 ± 1.07 & 48.67 ± 0.37 & 76.33 ± 1.12 & 74.94 ± 1.47 \\

        \rowcolor[HTML]{EFEFEF}
        Fog               & 55.52 ± 0.13 & 69.47 ± 4.86 & 85.08 ± 0.02 & 81.24 ± 0.19 & 89.49 ± 0.56 & 90.45 ± 0.05 \\

        Brightness        & 36.46 ± 0.00 & 55.69 ± 2.98 & 65.16 ± 0.00 & 65.88 ± 0.00 & 82.52 ± 0.90 & 82.84 ± 0.00 \\

        \rowcolor[HTML]{EFEFEF}
        Contrast          & 42.04 ± 0.00 & 00.64 ± 0.74 & 56.61 ± 0.21 & 49.36 ± 0.00 & 23.45 ± 4.15 & 20.26 ± 0.00 \\

        Elastic transform & 43.02 ± 0.09 & 92.96 ± 0.57 & 83.77 ± 0.15 & 83.65 ± 0.22 & 93.97 ± 0.38 & 93.67 ± 0.03 \\

        \rowcolor[HTML]{EFEFEF}
        Pixelate          & 02.14 ± 0.00 & 88.12 ± 0.65 & 55.20 ± 0.00 & 54.17 ± 0.00 & 89.43 ± 0.90 & 88.50 ± 0.00 \\

        JPEG compression  & 53.22 ± 0.00 & 90.53 ± 1.18 & 73.40 ± 0.00 & 71.14 ± 0.00 & 92.59 ± 0.40 & 91.71 ± 0.00 \\

        \rowcolor[HTML]{EFEFEF}
        \midrule
        mPC  & 36.11 & 61.17 & 63.91 & 60.26 & \textbf{76.50} & 75.95 \\
        
        \bottomrule
    \end{tabular}
    }

\end{table}

\begin{table}[!htp]
\caption{\textbf{AP$_{50}$ performance over the perturbations for LLVIP-C with severity level $5$ for RetinaNet.}}
\label{tab:retinanet_llvip_per_corruption}
    \centering
    \resizebox{\columnwidth}{!}{%
    \begin{tabular}{lcccccccc}

        \toprule

        {} & {} & \multicolumn{4}{c}{\multirow{2}{*}[-0.1em]{\textbf{LLVIP-C}}} \\

        \multirow{2}{*}[-1em]{} & \multirow{2}{*}[-1em]{Zero-Shot} & 
        \multirow{2}{*}[-1em]{FT} &
        \multirow{2}{*}[-1em]{LP} & \multirow{2}{*}[-1em]{LP-FT} & \multirow{2}{*}[-1em]{WiSE-OD$_{ZS}$} & \multirow{2}{*}[-1em]{WiSE-OD$_{LP}$} \\

        \cmidrule(lr){2-7}
        \addlinespace[10pt]     
        \midrule
        
              \rowcolor[HTML]{EFEFEF}
        Gaussian Noise      & 52.52 ± 0.21 & 70.55 ± 8.23  & 66.70 ± 0.19 & 66.32 ± 0.25 & 82.63 ± 3.59 & 88.10 ± 0.07 \\

        Shot Noise          & 44.70 ± 0.22 & 64.02 ± 10.87 & 57.85 ± 0.13 & 59.03 ± 0.17 & 78.94 ± 4.68 & 78.14 ± 1.53 \\

        \rowcolor[HTML]{EFEFEF}
        Impulse Noise       & 48.17 ± 0.19 & 73.01 ± 6.68  & 66.97 ± 1.82 & 67.78 ± 0.08 & 84.19 ± 3.05 & 86.53 ± 1.95 \\

        Defocus Blur        & 43.13 ± 0.00 & 75.38 ± 8.71  & 71.92 ± 0.00 & 73.08 ± 0.20 & 84.31 ± 3.89 & 88.45 ± 0.00 \\

        \rowcolor[HTML]{EFEFEF}
        Motion Blur         & 18.79 ± 0.07 & 71.69 ± 9.38  & 62.14 ± 0.15 & 63.62 ± 0.17 & 79.92 ± 4.78 & 78.59 ± 3.16 \\

        Zoom Blur           & 01.61 ± 0.00 & 12.73 ± 3.63  & 09.21 ± 0.00 & 09.93 ± 0.55 & 21.02 ± 5.04 & 16.92 ± 0.00 \\

        \rowcolor[HTML]{EFEFEF}
        Snow                & 35.54 ± 0.34 & 36.48 ± 11.52 & 56.68 ± 0.14 & 55.19 ± 0.18 & 71.09 ± 5.52 & 72.05 ± 4.35 \\

        Frost               & 36.52 ± 0.06 & 57.94 ± 6.50  & 53.83 ± 0.26 & 53.47 ± 0.32 & 75.35 ± 3.81 & 78.60 ± 1.74 \\

        \rowcolor[HTML]{EFEFEF}
        Fog                 & 58.48 ± 0.21 & 65.31 ± 6.58  & 83.18 ± 0.27 & 82.28 ± 0.06 & 85.07 ± 2.92 & 84.11 ± 0.12 \\

        Brightness          & 39.03 ± 0.00 & 54.83 ± 8.34  & 69.76 ± 1.69 & 70.79 ± 0.00 & 82.41 ± 1.89 & 80.70 ± 0.00 \\

        \rowcolor[HTML]{EFEFEF}
        Contrast            & 49.09 ± 0.00 & 00.99 ± 0.00  & 52.12 ± 0.00 & 54.61 ± 2.76 & 13.90 ± 1.79 & 13.64 ± 1.58 \\

        Elastic transform   & 37.37 ± 0.04 & 93.05 ± 1.27  & 76.89 ± 0.07 & 77.56 ± 1.20 & 94.52 ± 0.07 & 94.18 ± 0.09 \\

        \rowcolor[HTML]{EFEFEF}
        Pixelate            & 03.86 ± 0.00 & 88.82 ± 1.97  & 48.35 ± 0.00 & 49.68 ± 0.00 & 85.68 ± 2.38 & 89.83 ± 0.00 \\

        JPEG compression    & 56.24 ± 0.00 & 91.09 ± 1.73  & 69.68 ± 0.00 & 72.32 ± 2.93 & 92.67 ± 1.14 & 91.71 ± 0.00 \\

        \rowcolor[HTML]{EFEFEF}
        \midrule
        mPC  & 37.50 & 61.13 & 60.37 & 61.11 & 73.69 & \textbf{74.39} \\
        
        \bottomrule
    \end{tabular}
    }

\end{table}

\begin{table}[!htp]
\caption{\textbf{AP$_{50}$ performance over the perturbations for FLIR-C with severity level $2$ for FCOS.}}
\label{tab:fcos_flir_per_corruption}
    \centering
    \resizebox{\columnwidth}{!}{%
    \begin{tabular}{lcccccccc}

        \toprule

        {} & {} & \multicolumn{4}{c}{\multirow{2}{*}[-0.1em]{\textbf{FLIR-C}}} \\

        \multirow{2}{*}[-1em]{} & \multirow{2}{*}[-1em]{Zero-Shot} & 
        \multirow{2}{*}[-1em]{FT} &
        \multirow{2}{*}[-1em]{LP} & \multirow{2}{*}[-1em]{LP-FT} & \multirow{2}{*}[-1em]{WiSE-OD$_{ZS}$} & \multirow{2}{*}[-1em]{WiSE-OD$_{LP}$} \\

        \cmidrule(lr){2-7}
        \addlinespace[10pt]     
        \midrule
        
        \rowcolor[HTML]{EFEFEF}
        Gaussian Noise    & 23.65 ± 0.16 & 24.76 ± 2.44 & 32.89 ± 0.41 & 31.78 ± 1.07 & 38.53 ± 1.47 & 38.35 ± 1.20 \\
        
        Shot Noise        & 17.63 ± 0.26 & 15.39 ± 1.80 & 27.41 ± 0.17 & 25.10 ± 0.12 & 27.85 ± 1.06 & 30.10 ± 0.43 \\
        
        \rowcolor[HTML]{EFEFEF}
        Impulse Noise     & 14.12 ± 0.23 & 11.34 ± 0.88 & 20.93 ± 0.53 & 18.22 ± 0.05 & 21.07 ± 1.25 & 23.10 ± 0.19 \\
        
        Defocus Blur      & 19.23 ± 0.00 & 44.40 ± 1.40 & 33.08 ± 0.00 & 32.27 ± 0.32 & 48.14 ± 1.00 & 48.75 ± 0.85 \\
        
        \rowcolor[HTML]{EFEFEF}
        Motion Blur       & 20.77 ± 0.21 & 43.17 ± 1.44 & 33.95 ± 0.62 & 33.58 ± 0.31 & 45.65 ± 1.29 & 48.20 ± 0.50 \\
        
        Zoom Blur         & 6.73  ± 0.00 & 15.32 ± 0.14 & 11.81 ± 0.00 & 11.17 ± 0.00 & 15.44 ± 0.52 & 16.38 ± 0.54 \\
        
        \rowcolor[HTML]{EFEFEF}
        Snow              & 8.29  ± 0.32 & 07.63 ± 1.31 & 12.32 ± 0.18 & 13.04 ± 0.22 & 12.50 ± 1.08 & 11.42 ± 0.14 \\
        
        Frost             & 19.23 ± 0.26 & 30.80 ± 0.69 & 28.14 ± 0.27 & 27.60 ± 0.35 & 35.68 ± 0.55 & 36.37 ± 0.69 \\
        
        \rowcolor[HTML]{EFEFEF}
        Fog               & 51.56 ± 0.07 & 70.56 ± 0.41 & 63.60 ± 0.53 & 62.12 ± 0.14 & 74.39 ± 0.84 & 72.79 ± 0.26 \\
        
        Brightness        & 58.43 ± 0.00 & 69.69 ± 1.05 & 64.79 ± 0.00 & 64.78 ± 0.00 & 75.59 ± 0.94 & 73.02 ± 0.17 \\
        
        \rowcolor[HTML]{EFEFEF}
        Contrast          & 50.28 ± 0.00 & 70.41 ± 0.96 & 62.97 ± 0.10 & 61.36 ± 0.00 & 73.99 ± 0.69 & 73.02 ± 0.33 \\
        
        Elastic transform & 36.76 ± 0.48 & 64.18 ± 0.74 & 54.62 ± 0.82 & 52.40 ± 0.75 & 68.95 ± 0.52 & 66.47 ± 0.32 \\
        
        \rowcolor[HTML]{EFEFEF}
        Pixelate          & 32.65 ± 0.00 & 51.74 ± 0.86 & 45.58 ± 0.00 & 47.96 ± 0.00 & 59.03 ± 0.09 & 57.42 ± 0.00 \\
        
        JPEG compression  & 44.71 ± 0.00 & 55.63 ± 0.96 & 52.90 ± 0.00 & 52.61 ± 0.00 & 63.03 ± 0.74 & 59.33 ± 0.00 \\
        
        \bottomrule
        \rowcolor[HTML]{EFEFEF}
        mPC & 28.85 & 41.07 & 38.92 & 38.14 & \textbf{47.13} & 46.76  \\

        \bottomrule
    \end{tabular}
    }

\end{table}

\begin{table}[!htp]
\caption{\textbf{AP$_{50}$ performance over the perturbations for FLIR-C with severity level $2$ for RetinaNet.}}
\label{tab:retinanet_flir_per_corruption}
    \centering
    \resizebox{\columnwidth}{!}{%
    \begin{tabular}{lcccccccc}

        \toprule

        {} & {} & \multicolumn{4}{c}{\multirow{2}{*}[-0.1em]{\textbf{FLIR-C}}} \\

        \multirow{2}{*}[-1em]{} & \multirow{2}{*}[-1em]{Zero-Shot} & 
        \multirow{2}{*}[-1em]{FT} &
        \multirow{2}{*}[-1em]{LP} & \multirow{2}{*}[-1em]{LP-FT} & \multirow{2}{*}[-1em]{WiSE-OD$_{ZS}$} & \multirow{2}{*}[-1em]{WiSE-OD$_{LP}$} \\

        \cmidrule(lr){2-7}
        \addlinespace[10pt]     
        \midrule

        \rowcolor[HTML]{EFEFEF}
        Gaussian Noise    & 25.61 ± 0.20 & 26.47 ± 3.14 & 33.34 ± 0.38 & 31.44 ± 0.52 & 34.31 ± 2.83 & 33.09 ± 0.26 \\

        Shot Noise        & 19.36 ± 0.06 & 16.95 ± 2.60 & 25.49 ± 0.26 & 24.16 ± 0.26 & 24.15 ± 2.42 & 29.00 ± 0.85 \\

        \rowcolor[HTML]{EFEFEF}
        Impulse Noise     & 14.82 ± 0.17 & 15.90 ± 1.98 & 18.58 ± 0.28 & 19.12 ± 0.28 & 21.77 ± 1.50 & 24.58 ± 0.37 \\

        Defocus Blur      & 19.30 ± 0.00 & 46.15 ± 1.86 & 31.28 ± 0.00 & 30.62 ± 0.00 & 44.74 ± 1.61 & 48.43 ± 0.63 \\

        \rowcolor[HTML]{EFEFEF}
        Motion Blur       & 20.05 ± 0.16 & 45.96 ± 3.97 & 29.39 ± 0.11 & 30.28 ± 0.16 & 42.91 ± 2.90 & 50.99 ± 0.23 \\

        Zoom Blur         & 07.05 ± 0.00 & 15.99 ± 1.06 & 10.48 ± 0.00 & 10.55 ± 0.00 & 15.40 ± 0.32 & 16.03 ± 0.00 \\

        \rowcolor[HTML]{EFEFEF}
        Snow              & 07.91 ± 0.16 & 08.34 ± 1.84 & 09.29 ± 0.08 & 09.54 ± 0.34 & 11.09 ± 1.44 & 12.80 ± 1.38 \\

        Frost             & 17.63 ± 0.32 & 33.90 ± 3.17 & 24.81 ± 0.34 & 24.31 ± 0.44 & 33.02 ± 2.41 & 36.32 ± 0.31 \\

        \rowcolor[HTML]{EFEFEF}
        Fog               & 48.95 ± 0.13 & 70.48 ± 0.40 & 60.61 ± 0.11 & 60.32 ± 0.11 & 72.46 ± 0.50 & 74.75 ± 1.09 \\

        Brightness        & 56.74 ± 0.00 & 70.06 ± 0.30 & 61.90 ± 0.00 & 63.21 ± 0.00 & 74.80 ± 0.56 & 74.52 ± 1.18 \\

        \rowcolor[HTML]{EFEFEF}
        Contrast          & 47.50 ± 0.00 & 70.75 ± 0.38 & 59.36 ± 0.00 & 59.94 ± 0.00 & 71.89 ± 0.43 & 74.44 ± 0.48 \\

        Elastic transform & 34.27 ± 0.34 & 65.63 ± 0.54 & 50.84 ± 0.50 & 49.74 ± 0.42 & 68.19 ± 0.54 & 68.58 ± 1.09 \\

        \rowcolor[HTML]{EFEFEF}
        Pixelate          & 32.52 ± 0.00 & 55.52 ± 1.21 & 45.84 ± 0.00 & 45.96 ± 0.00 & 57.66 ± 0.30 & 59.29 ± 0.00 \\

        JPEG compression  & 44.10 ± 0.00 & 55.89 ± 0.86 & 52.80 ± 0.00 & 51.23 ± 0.00 & 62.57 ± 1.02 & 62.71 ± 0.00 \\

        \bottomrule
        \rowcolor[HTML]{EFEFEF}
        mPC & 28.27 & 42.71 & 36.71 & 36.45 & 45.35 & \textbf{47.53} \\

        \bottomrule
    \end{tabular}
    }

\end{table}

\section{Activation map analysis}
\label{sec:gradcam}

\begin{figure*}
\captionsetup[subfigure]{labelformat=empty}
\centering
\begin{subfigure}[t]{0.50\columnwidth}
    \caption{Zero-Shot}
    \makebox[-3pt][r]{\makebox[15pt]{\raisebox{50pt}{\rotatebox[origin=c]{90}{Gaussian Noise}}}}
    \includegraphics[width=\columnwidth]{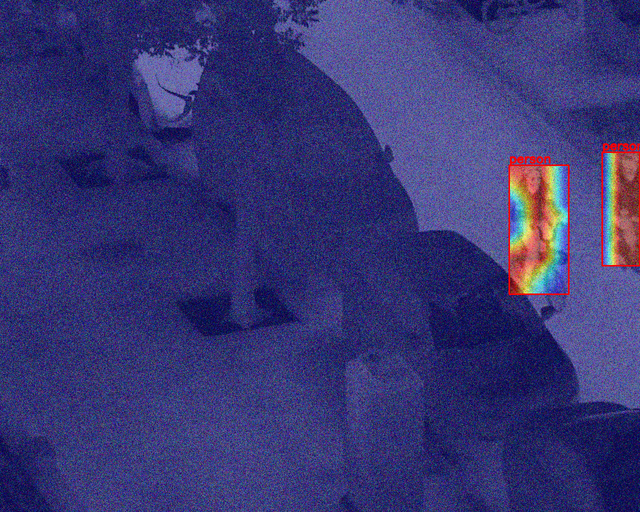}
\end{subfigure}
\begin{subfigure}[t]{0.50\columnwidth}
    \caption{WiSE-OD$_{ZS}$}
    \includegraphics[width=\columnwidth]{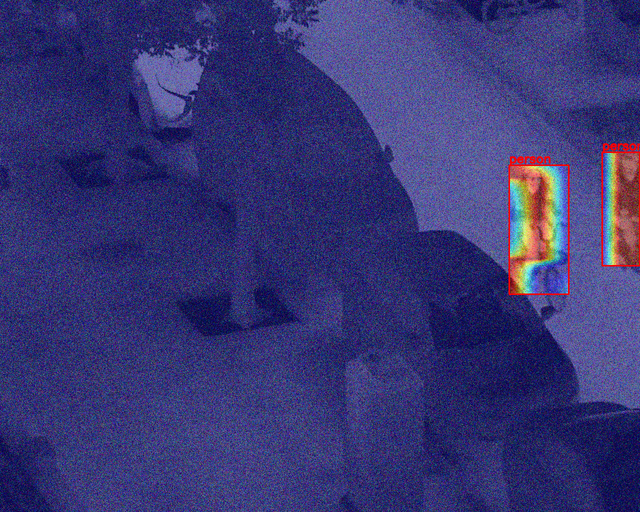}
\end{subfigure}
\begin{subfigure}[t]{0.50\columnwidth}
    \caption{FT}
    \includegraphics[width=\columnwidth]{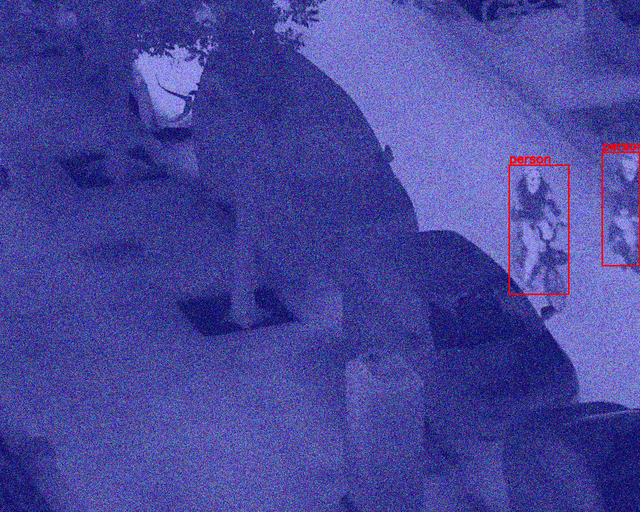}
\end{subfigure}

\begin{subfigure}[t]{0.50\columnwidth}
    \caption{Zero-Shot}
    \makebox[-3pt][r]{\makebox[15pt]{\raisebox{50pt}{\rotatebox[origin=c]{90}{Shot Noise}}}}
    \includegraphics[width=\columnwidth]{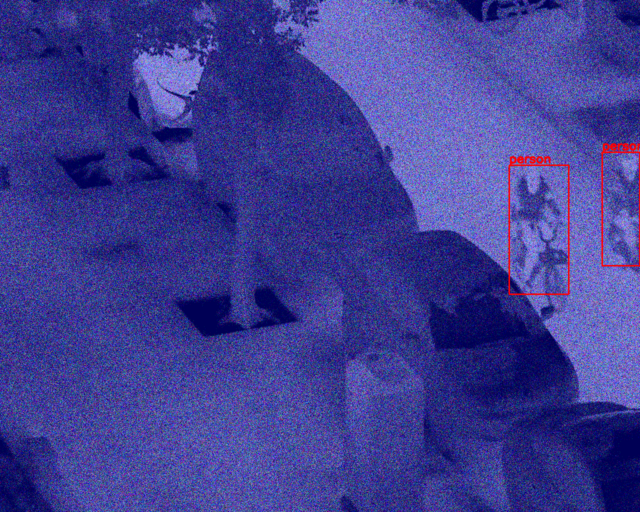}
\end{subfigure}
\begin{subfigure}[t]{0.50\columnwidth}
    \caption{WiSE-OD$_{ZS}$}
    \includegraphics[width=\columnwidth]{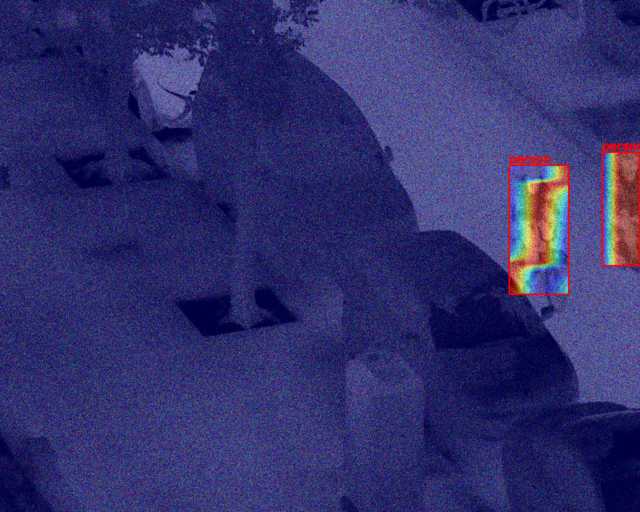}
\end{subfigure}
\begin{subfigure}[t]{0.50\columnwidth}
    \caption{FT}
    \includegraphics[width=\columnwidth]{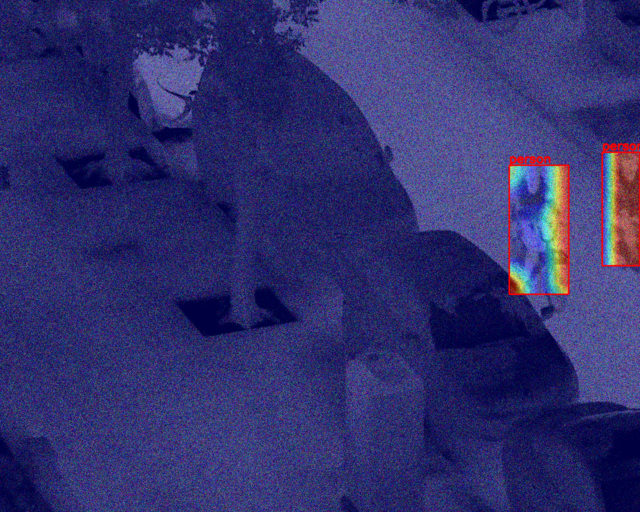}
\end{subfigure}

\begin{subfigure}[t]{0.50\columnwidth}
    \caption{Zero-Shot}
    \makebox[-3pt][r]{\makebox[15pt]{\raisebox{50pt}{\rotatebox[origin=c]{90}{Impulse Noise}}}}
    \includegraphics[width=\columnwidth]{imgs/gradcam/llvip/llvip_fasterrcnn_0_6_alpha0.0impulse_noise5_cam.png}
\end{subfigure}
\begin{subfigure}[t]{0.50\columnwidth}
    \caption{WiSE-OD$_{ZS}$}
    \includegraphics[width=\columnwidth]{imgs/gradcam/llvip/llvip_fasterrcnn_0_6_alpha0.5impulse_noise5_cam.png}
\end{subfigure}
\begin{subfigure}[t]{0.50\columnwidth}
    \caption{FT}
    \includegraphics[width=\columnwidth]{imgs/gradcam/llvip/llvip_fasterrcnn_0_6_alpha1.0impulse_noise5_cam.png}
\end{subfigure}

\begin{subfigure}[t]{0.50\columnwidth}
    \caption{Zero-Shot}
    \makebox[-3pt][r]{\makebox[15pt]{\raisebox{50pt}{\rotatebox[origin=c]{90}{Defocus Blur}}}}
    \includegraphics[width=\columnwidth]{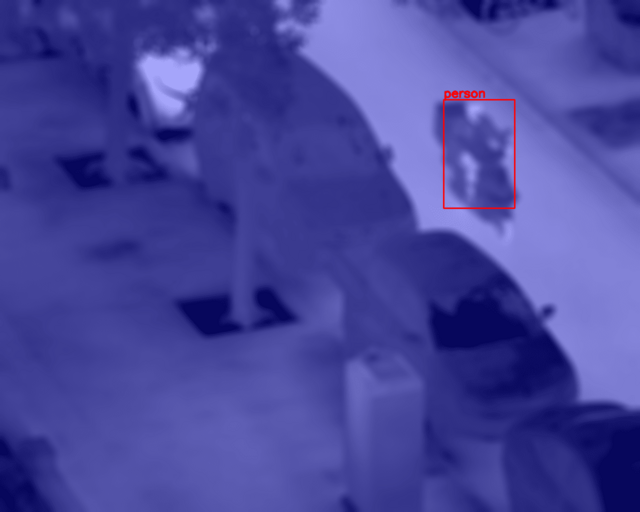}
\end{subfigure}
\begin{subfigure}[t]{0.50\columnwidth}
    \caption{WiSE-OD$_{ZS}$}
    \includegraphics[width=\columnwidth]{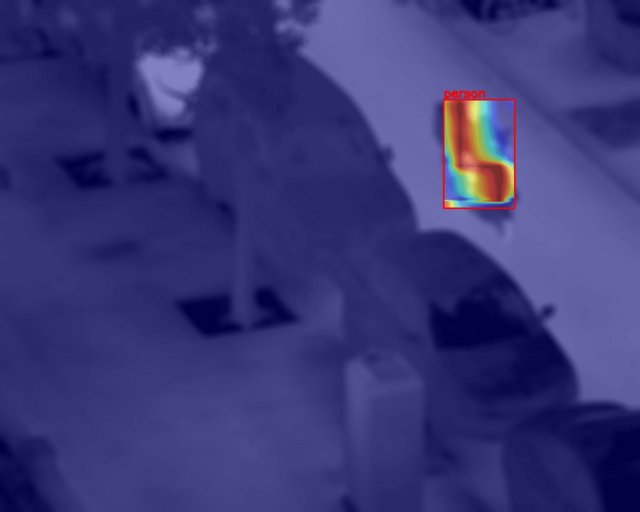}
\end{subfigure}
\begin{subfigure}[t]{0.50\columnwidth}
    \caption{FT}
    \includegraphics[width=\columnwidth]{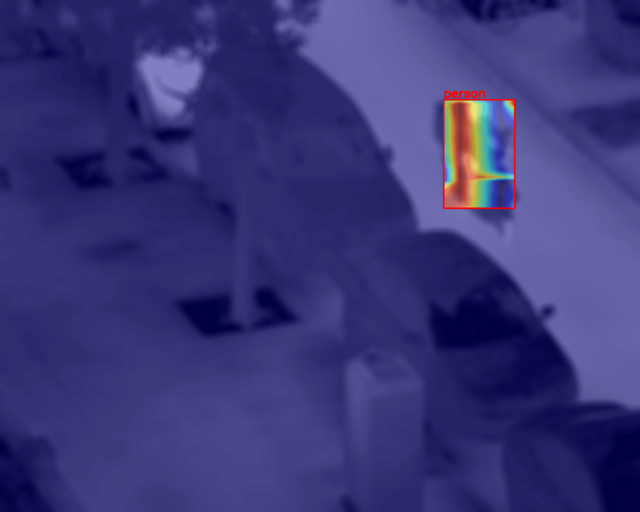}
\end{subfigure}

\begin{subfigure}[t]{0.50\columnwidth}
    \caption{Zero-Shot}
    \makebox[-3pt][r]{\makebox[15pt]{\raisebox{50pt}{\rotatebox[origin=c]{90}{Motion Blur}}}}
    \includegraphics[width=\columnwidth]{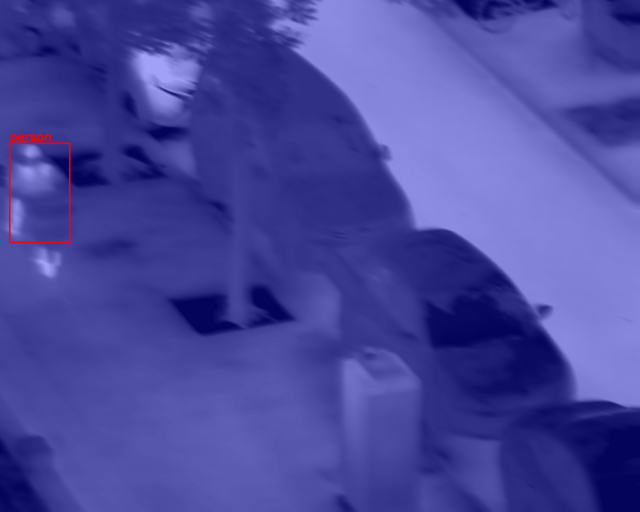}
\end{subfigure}
\begin{subfigure}[t]{0.50\columnwidth}
    \caption{WiSE-OD$_{ZS}$}
    \includegraphics[width=\columnwidth]{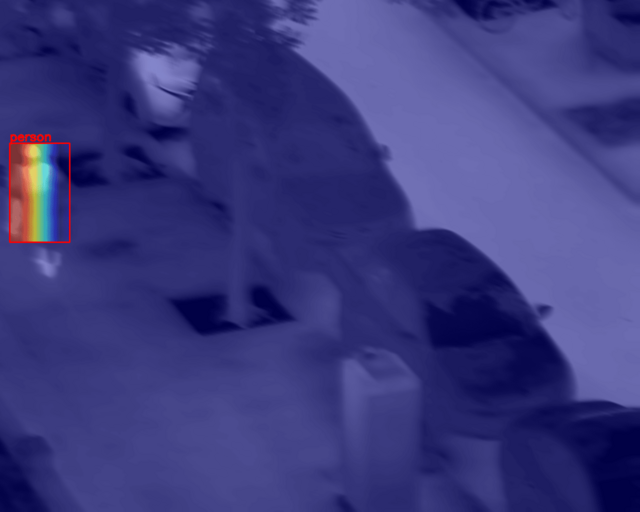}
\end{subfigure}
\begin{subfigure}[t]{0.50\columnwidth}
    \caption{FT}
    \includegraphics[width=\columnwidth]{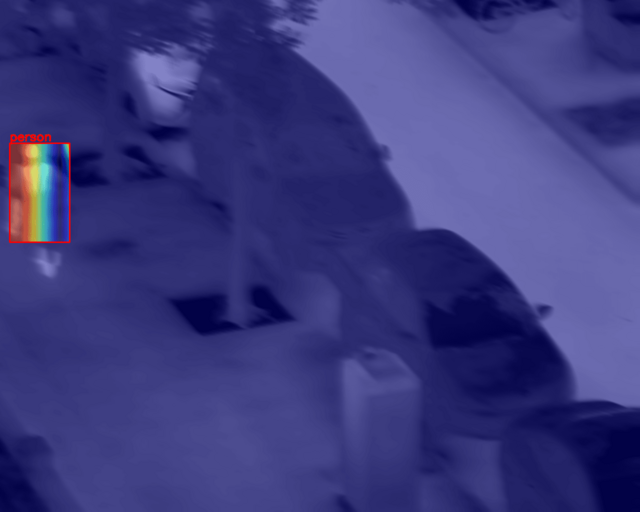}
\end{subfigure}

\caption{\textbf{Activation map analysis for zero-shot COCO pre-train Faster R-CNN detector, the WiSE-OD$_{ZS}$ and FT detector on IR for LLVIP-C dataset.} In red are the GTs, and for the WiSE-OD$_{ZS}$, the models are able to activate the features that represent a person for such corruptions (Part 1 with $5$ of the $14$ corruptions).}
\label{fig:activation_map_part1}
\end{figure*}

\begin{figure*}
\captionsetup[subfigure]{labelformat=empty}
\centering
\begin{subfigure}[t]{0.50\columnwidth}
    \caption{Zero-Shot}
    \makebox[-3pt][r]{\makebox[15pt]{\raisebox{50pt}{\rotatebox[origin=c]{90}{Zoom Blur}}}}
    \includegraphics[width=\columnwidth]{imgs/gradcam/llvip/llvip_fasterrcnn_0_5_alpha0.0zoom_blur5_cam.png}
\end{subfigure}
\begin{subfigure}[t]{0.50\columnwidth}
    \caption{WiSE-OD$_{ZS}$}
    \includegraphics[width=\columnwidth]{imgs/gradcam/llvip/llvip_fasterrcnn_0_5_alpha0.5zoom_blur5_cam.png}
\end{subfigure}
\begin{subfigure}[t]{0.50\columnwidth}
    \caption{FT}
    \includegraphics[width=\columnwidth]{imgs/gradcam/llvip/llvip_fasterrcnn_0_5_alpha1.0zoom_blur5_cam.png}
\end{subfigure}

\begin{subfigure}[t]{0.50\columnwidth}
    \caption{Zero-Shot}
    \makebox[-3pt][r]{\makebox[15pt]{\raisebox{50pt}{\rotatebox[origin=c]{90}{Snow}}}}
    \includegraphics[width=\columnwidth]{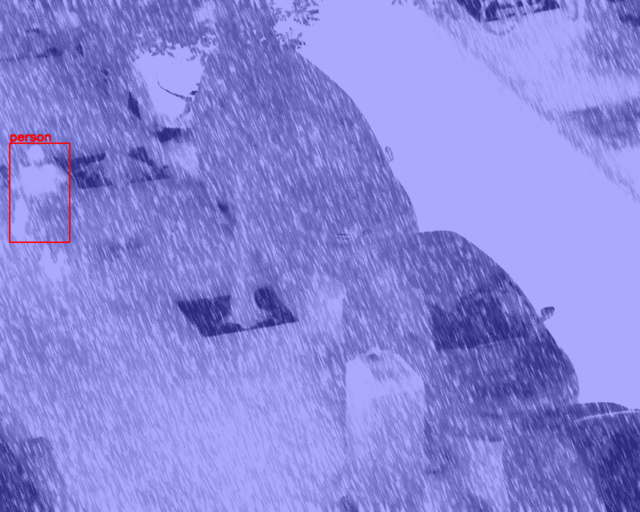}
\end{subfigure}
\begin{subfigure}[t]{0.50\columnwidth}
    \caption{WiSE-OD$_{ZS}$}
    \includegraphics[width=\columnwidth]{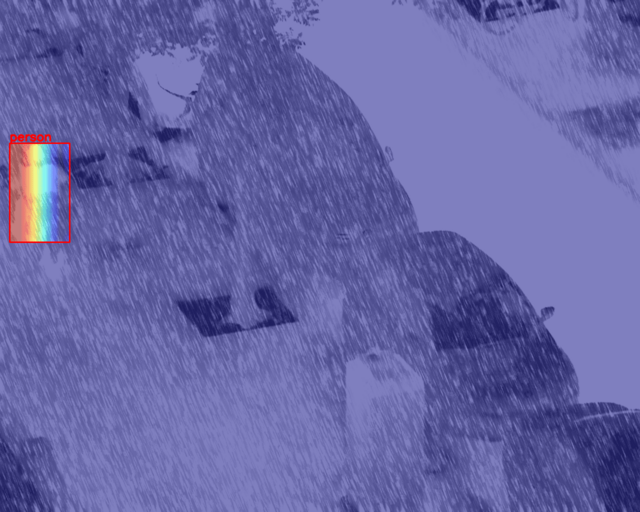}
\end{subfigure}
\begin{subfigure}[t]{0.50\columnwidth}
    \caption{FT}
    \includegraphics[width=\columnwidth]{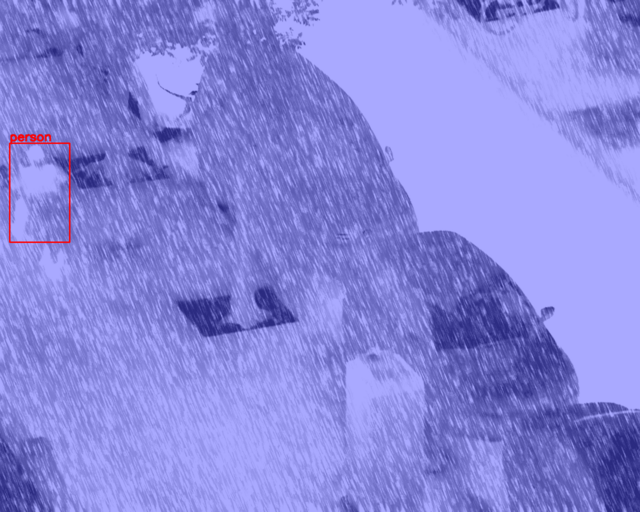}
\end{subfigure}

\begin{subfigure}[t]{0.50\columnwidth}
    \caption{Zero-Shot}
    \makebox[-3pt][r]{\makebox[15pt]{\raisebox{50pt}{\rotatebox[origin=c]{90}{Frost}}}}
    \includegraphics[width=\columnwidth]{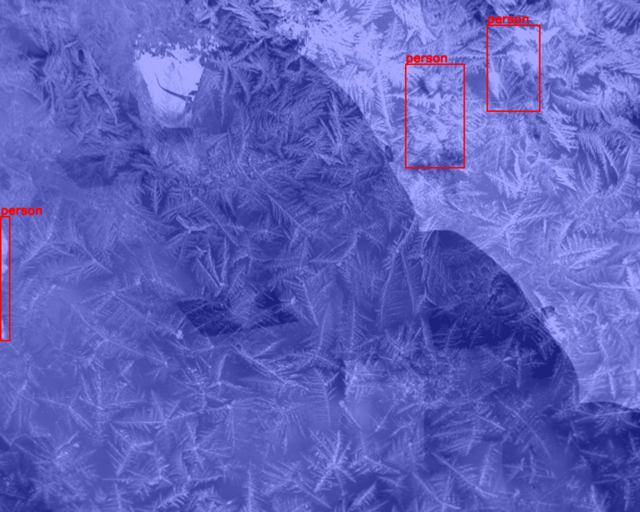}
\end{subfigure}
\begin{subfigure}[t]{0.50\columnwidth}
    \caption{WiSE-OD$_{ZS}$}
    \includegraphics[width=\columnwidth]{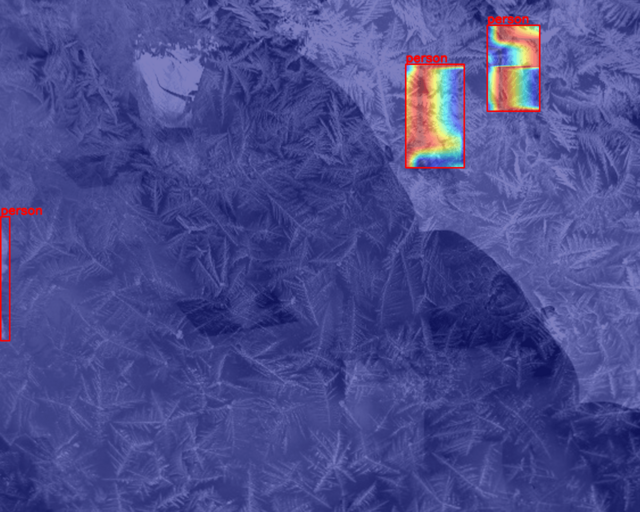}
\end{subfigure}
\begin{subfigure}[t]{0.50\columnwidth}
    \caption{FT}
    \includegraphics[width=\columnwidth]{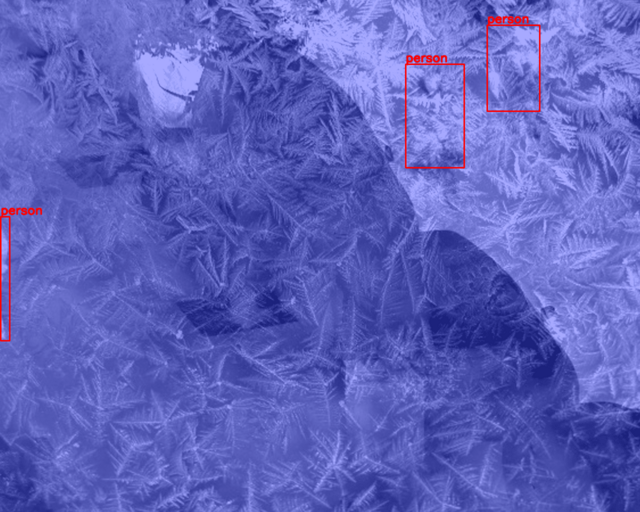}
\end{subfigure}

\begin{subfigure}[t]{0.50\columnwidth}
    \caption{Zero-Shot}
    \makebox[-3pt][r]{\makebox[15pt]{\raisebox{50pt}{\rotatebox[origin=c]{90}{Fog}}}}
    \includegraphics[width=\columnwidth]{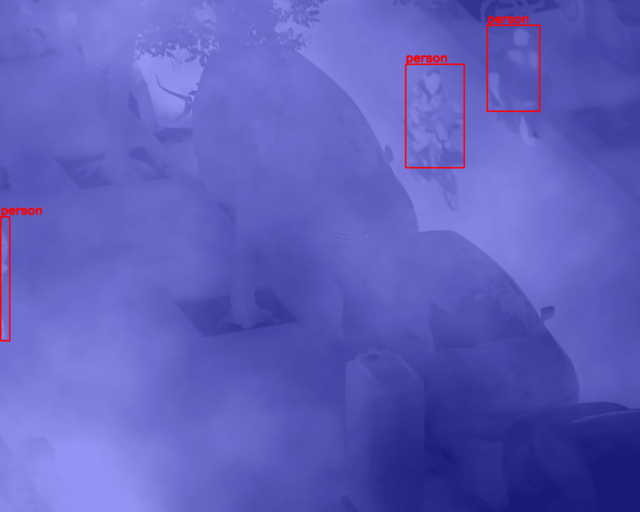}
\end{subfigure}
\begin{subfigure}[t]{0.50\columnwidth}
    \caption{WiSE-OD$_{ZS}$}
    \includegraphics[width=\columnwidth]{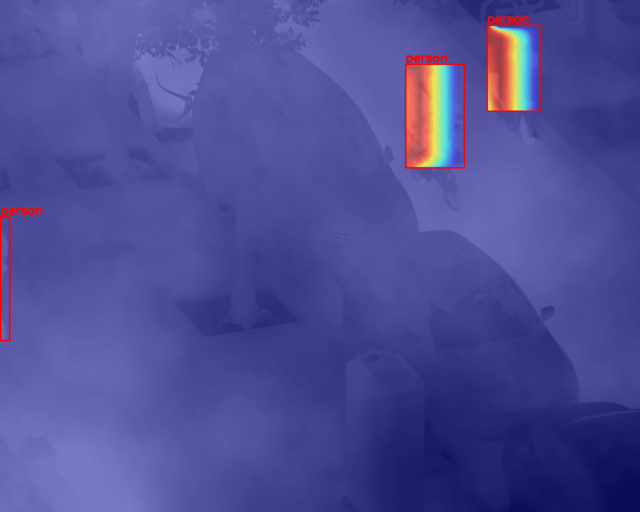}
\end{subfigure}
\begin{subfigure}[t]{0.50\columnwidth}
    \caption{FT}
    \includegraphics[width=\columnwidth]{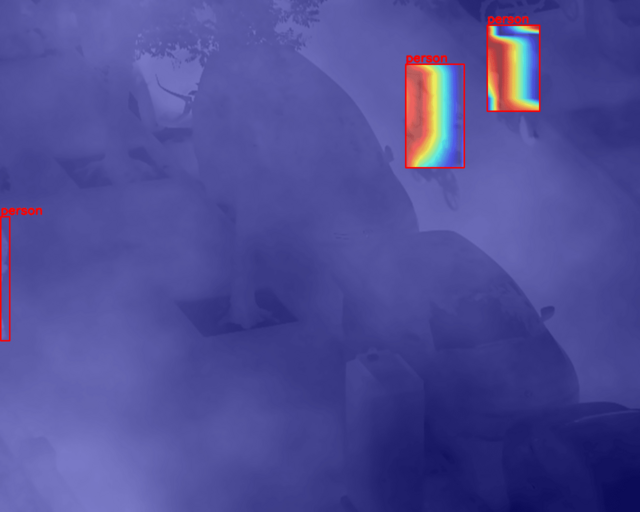}
\end{subfigure}

\begin{subfigure}[t]{0.50\columnwidth}
    \caption{Zero-Shot}
    \makebox[-3pt][r]{\makebox[15pt]{\raisebox{50pt}{\rotatebox[origin=c]{90}{Brightness}}}}
    \includegraphics[width=\columnwidth]{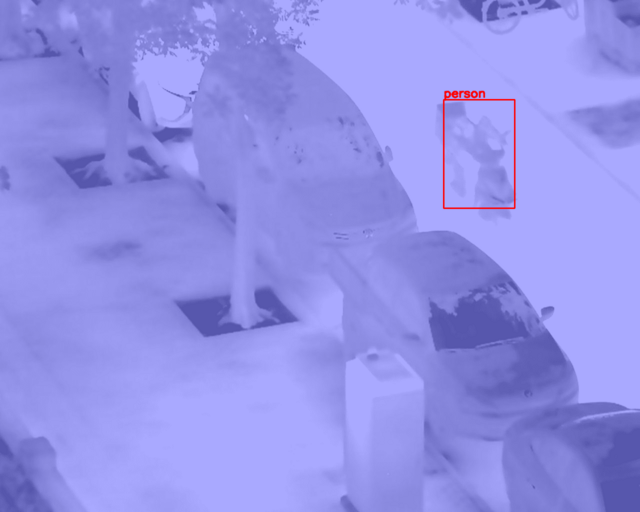}
\end{subfigure}
\begin{subfigure}[t]{0.50\columnwidth}
    \caption{WiSE-OD$_{ZS}$}
    \includegraphics[width=\columnwidth]{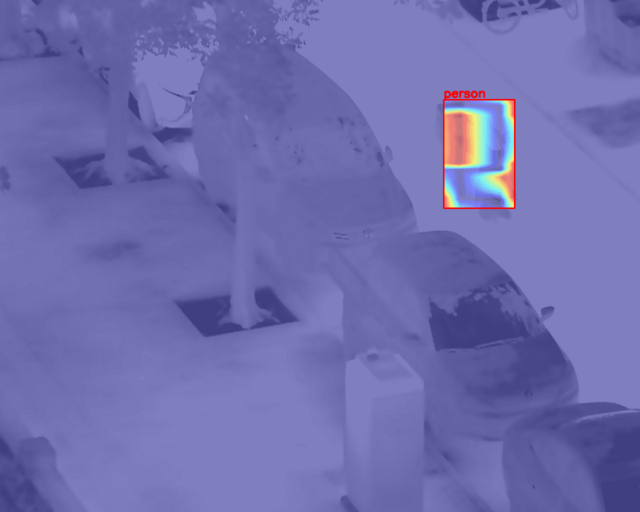}
\end{subfigure}
\begin{subfigure}[t]{0.50\columnwidth}
    \caption{FT}
    \includegraphics[width=\columnwidth]{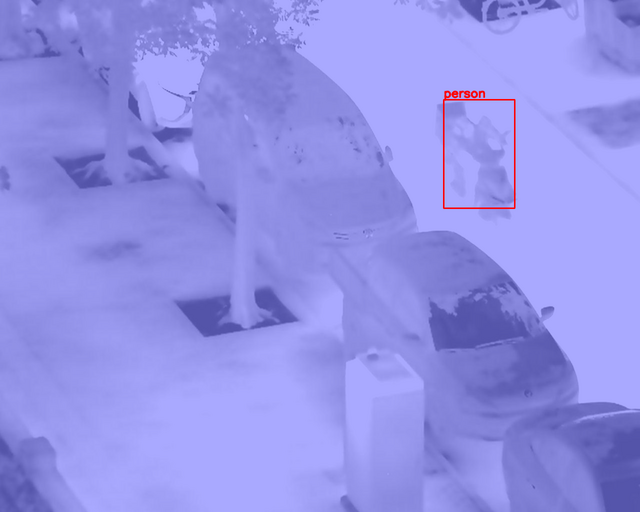}
\end{subfigure}

\caption{\textbf{Activation map analysis for zero-shot COCO pre-train Faster R-CNN detector, the WiSE-OD$_{ZS}$ and FT detector on IR for LLVIP-C dataset.} In red are the GTs, and for the WiSE-OD$_{ZS}$, the models are able to activate the features that represent a person for such corruptions (Part 2 with $5$ of the $14$ corruptions).}
\label{fig:activation_map_part2}
\end{figure*}

\begin{figure*}
\captionsetup[subfigure]{labelformat=empty}
\centering

\begin{subfigure}[t]{0.50\columnwidth}
    \caption{Zero-Shot}
    \makebox[-3pt][r]{\makebox[15pt]{\raisebox{50pt}{\rotatebox[origin=c]{90}{Contrast}}}}
    \includegraphics[width=\columnwidth]{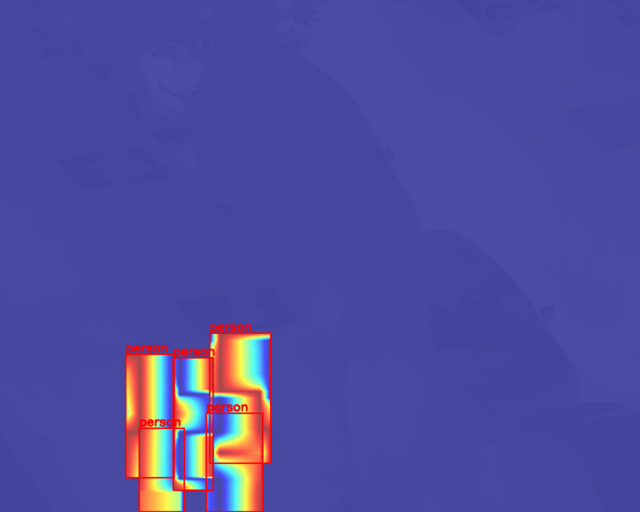}
\end{subfigure}
\begin{subfigure}[t]{0.50\columnwidth}
    \caption{WiSE-OD$_{ZS}$}
    \includegraphics[width=\columnwidth]{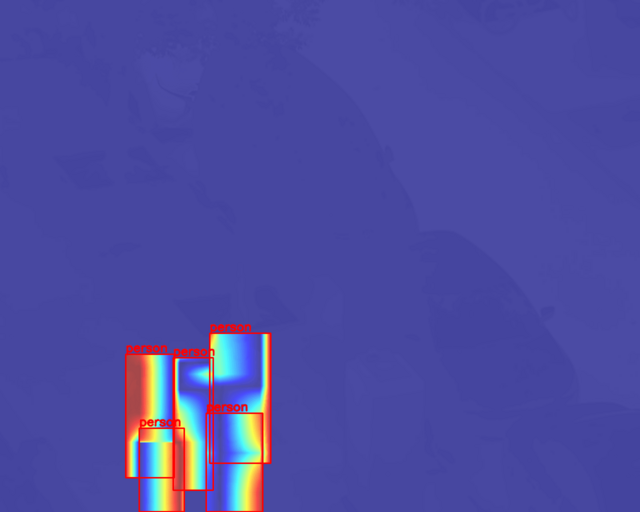}
\end{subfigure}
\begin{subfigure}[t]{0.50\columnwidth}
    \caption{FT}
    \includegraphics[width=\columnwidth]{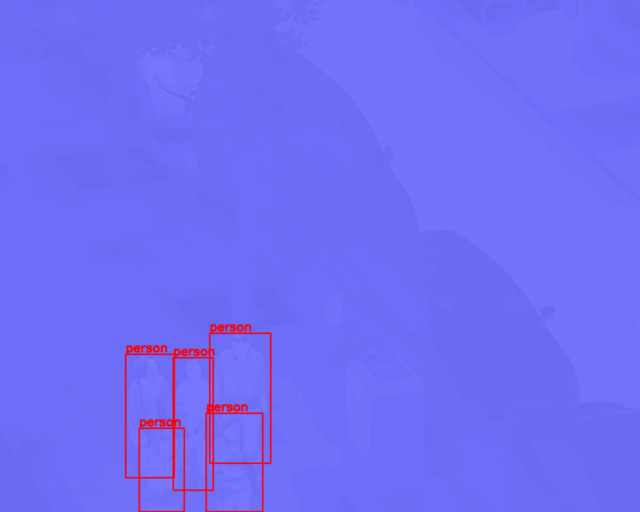}
\end{subfigure}

\begin{subfigure}[t]{0.50\columnwidth}
    \caption{Zero-Shot}
    \makebox[-3pt][r]{\makebox[15pt]{\raisebox{50pt}{\rotatebox[origin=c]{90}{Elastic Transform}}}}
    \includegraphics[width=\columnwidth]{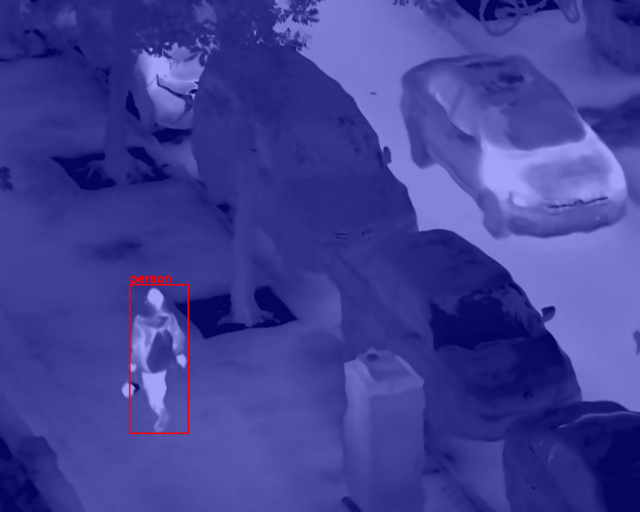}
\end{subfigure}
\begin{subfigure}[t]{0.50\columnwidth}
    \caption{WiSE-OD$_{ZS}$}
    \includegraphics[width=\columnwidth]{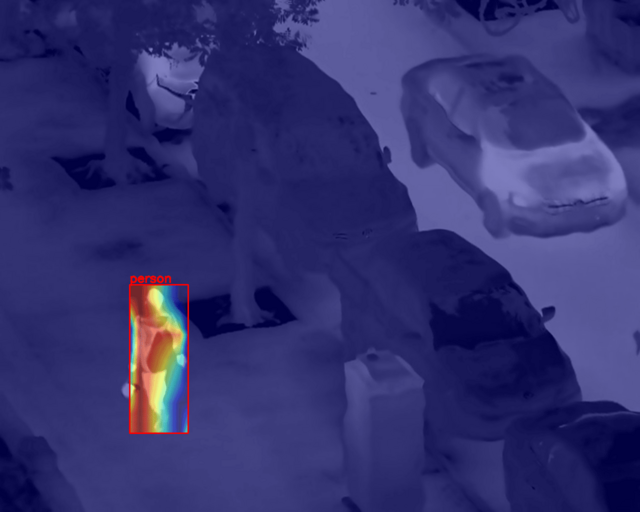}
\end{subfigure}
\begin{subfigure}[t]{0.50\columnwidth}
    \caption{FT}
    \includegraphics[width=\columnwidth]{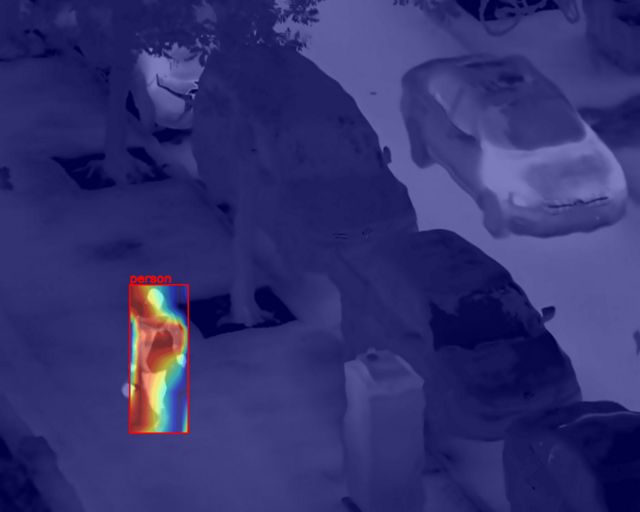}
\end{subfigure}

\begin{subfigure}[t]{0.50\columnwidth}
    \caption{Zero-Shot}
    \makebox[-3pt][r]{\makebox[15pt]{\raisebox{50pt}{\rotatebox[origin=c]{90}{Pixelate}}}}
    \includegraphics[width=\columnwidth]{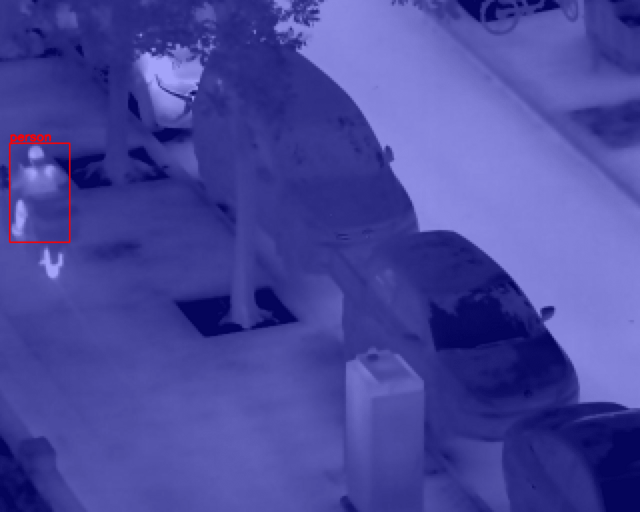}
\end{subfigure}
\begin{subfigure}[t]{0.50\columnwidth}
    \caption{WiSE-OD$_{ZS}$}
    \includegraphics[width=\columnwidth]{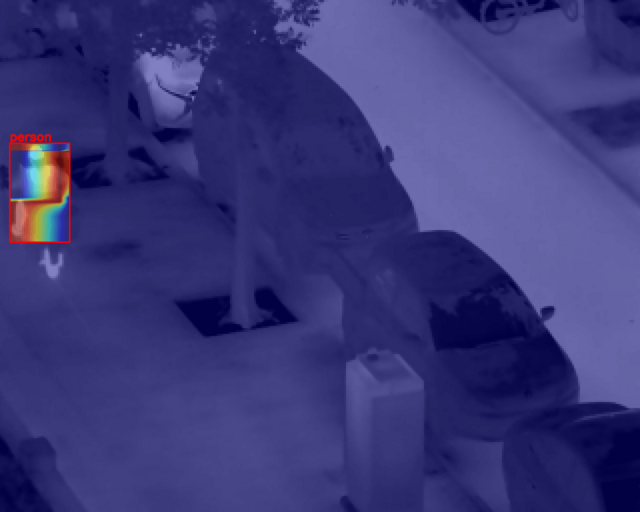}
\end{subfigure}
\begin{subfigure}[t]{0.50\columnwidth}
    \caption{FT}
    \includegraphics[width=\columnwidth]{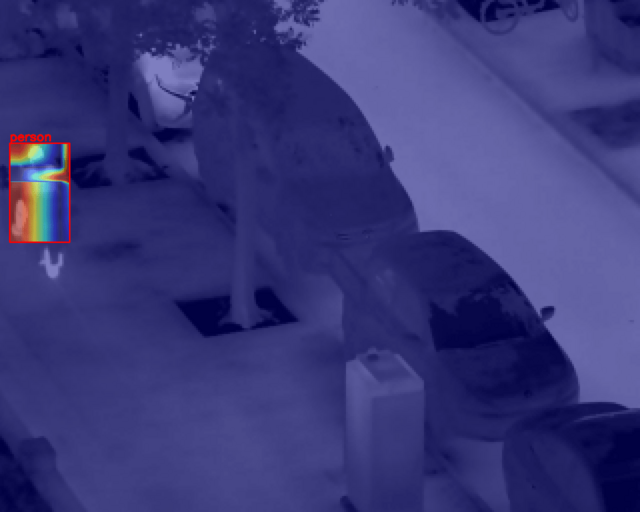}
\end{subfigure}

\begin{subfigure}[t]{0.50\columnwidth}
    \caption{Zero-Shot}
    \makebox[-3pt][r]{\makebox[15pt]{\raisebox{50pt}{\rotatebox[origin=c]{90}{JPEG Compression}}}}
    \includegraphics[width=\columnwidth]{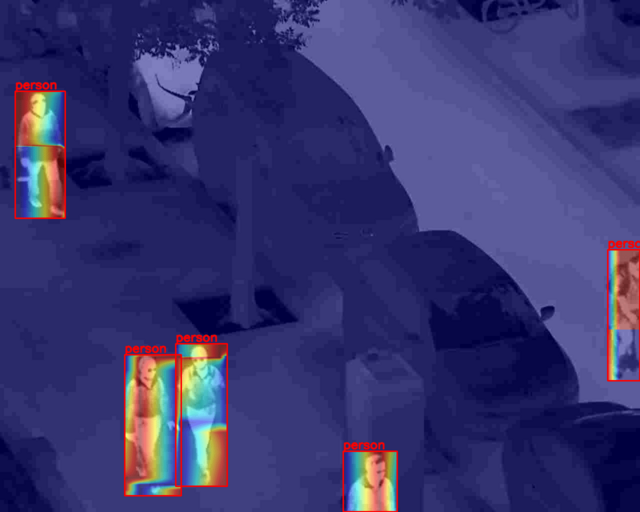}
\end{subfigure}
\begin{subfigure}[t]{0.50\columnwidth}
    \caption{WiSE-OD$_{ZS}$}
    \includegraphics[width=\columnwidth]{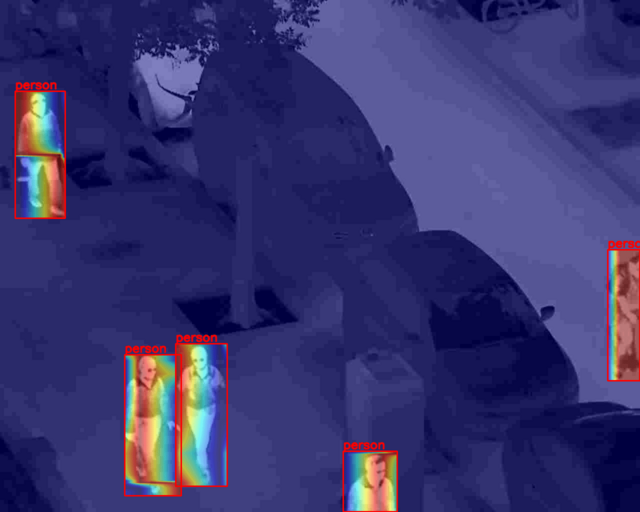}
\end{subfigure}
\begin{subfigure}[t]{0.50\columnwidth}
    \caption{FT}
    \includegraphics[width=\columnwidth]{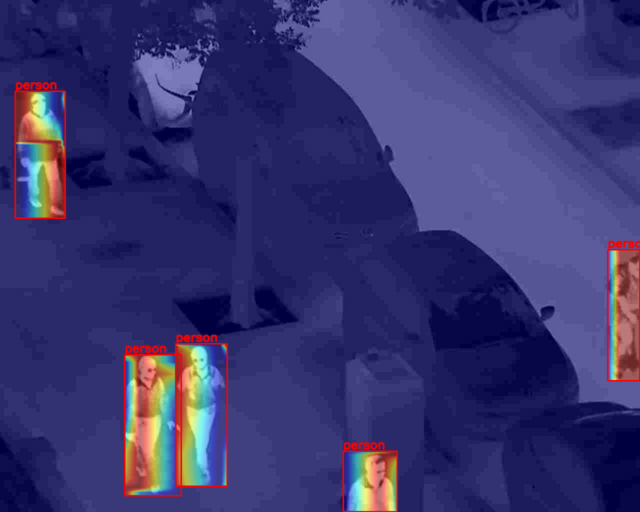}
\end{subfigure}

\caption{\textbf{Activation map analysis for zero-shot COCO pre-train Faster R-CNN detector, the WiSE-OD$_{ZS}$ and FT detector on IR for LLVIP-C dataset}. In red are the GTs, and for the WiSE-OD$_{ZS}$, the models are able to activate the features that represent a person for such corruptions (Part 3 with $4$ of the $14$ corruptions).}
\label{fig:activation_map_part3}
\end{figure*}

In this section, we provide more plots with the Grad-CAM activations. Here, we divided into three figures due to space constraints: ~\fref{fig:activation_map_part1}, 
~\fref{fig:activation_map_part2}, and ~\fref{fig:activation_map_part3}. In many cases, the WiSE-OD$_{ZS}$ had a person detected on average, which is shown by the red highlighted part of the images.


\section{WiSE-OD\texorpdfstring{$_{ZS}$}{ZS}: Ablation study on \texorpdfstring{$\lambda$}{lambda}}
\label{sec:ablation_lambdas}

In this section, we expanded our ablation study. In the~\tref{tab:fasterrcnn_llvip_ablation_lambda_wiseft}, we have the full study for Faster R-CNN with LLVIP-C and in~\tref{tab:fasterrcnn_flir_ablation_lambda_wiseft} for FLIR-C, as well as the FCOS in~\tref{tab:fcos_flir_ablation_lambda_wiseft} and RetinaNet in~\tref{tab:retinanet_flir_ablation_lambda_wiseft}. As mentioned in the main manuscript, the $\lambda$ = 0.5 has a 
 good performance under corruption without having to tune the $\lambda$ over a validation set, so it is a good choice for most of the cases; to be better in a specific corruption, we could tune the $\lambda$.

\begin{table*}[ht]
\caption{\textbf{Ablation of $\lambda$ over LLVIP-C dataset for Faster R-CNN. Where $\lambda = 0.0$ represents the zero-shot model, $\lambda = 0.5$ represents default WiSE-OD$_{ZS}$ and $\lambda = 1.0$ represents the fine-tuning model. For LLVIP-C, the severity level is $5$.}}
\label{tab:fasterrcnn_llvip_ablation_lambda_wiseft}

    \centering
    \resizebox{\textwidth}{!}{%
    \begin{tabular}{lcccccccccccc}

        \toprule \\
        
        {} &  \multicolumn{12}{c}{\textbf{Test Set IR (Dataset: LLVIP-C)}} \\

        \multirow{2}{*}[-1em]{} & \multirow{2}{*}[-1em]{$\theta(\lambda=0.0)$} & 
        \multirow{2}{*}[-1em]{$\theta(\lambda=0.1)$} & \multirow{2}{*}[-1em]{$\theta(\lambda=0.2)$} & \multirow{2}{*}[-1em]{$\theta(\lambda=0.3)$} & \multirow{2}{*}[-1em]{$\theta(\lambda=0.4)$} & 
        \multirow{2}{*}[-1em]{$\theta(\lambda=0.5)$} & \multirow{2}{*}[-1em]{$\theta(\lambda=0.6)$} & \multirow{2}{*}[-1em]{$\theta(\lambda=0.7)$} & \multirow{2}{*}[-1em]{$\theta(\lambda=0.8)$} & \multirow{2}{*}[-1em]{$\theta(\lambda=0.9)$} & \multirow{2}{*}[-1em]{$\theta(\lambda=1.0)$} \\

        \cmidrule(lr){2-12}
        \addlinespace[10pt]

        \midrule

        Original & 71.21 ± 0.02 & 90.88 ± 0.12 & 93.88 ± 0.28 & 95.16 ± 0.10 & 95.73 ± 0.15 & 96.06 ± 0.22 & 96.03 ± 0.29 & 95.88 ± 0.33 & 95.41 ± 0.60 & 94.72 ± 0.53 & 93.68 ± 0.86 &  \\

        \midrule
        \rowcolor[HTML]{EFEFEF}
         Gaussian Noise & 59.24 ± 0.07 & 83.30 ± 0.26 & 86.52 ± 0.40 & 87.13 ± 0.50 & 87.34 ± 0.15 & 86.68 ± 0.44 & 85.04 ± 1.02 & 82.37 ± 1.94 & 78.47 ± 3.43 & 73.39 ± 5.44 & 67.46 ± 7.45 &  \\

        Shot Noise & 51.48 ± 0.14 & 80.36 ± 0.15 & 83.86 ± 0.70 & 85.49 ± 0.60 & 86.00 ± 0.16 & 85.26 ± 0.50 & 83.47 ± 1.19 & 80.54 ± 2.08 & 76.42 ± 3.74 & 71.03 ± 5.67 & 64.83 ± 7.79 &  \\

        \rowcolor[HTML]{EFEFEF}
        Impulse Noise & 56.62 ± 0.07 & 82.52 ± 0.23 & 86.93 ± 0.65 & 88.47 ± 0.83 & 88.90 ± 0.32 & 88.54 ± 0.33 & 87.09 ± 0.89 & 84.64 ± 1.73 & 80.94 ± 2.70 & 76.78 ± 4.21 & 71.32 ± 6.33 &  \\

        Defocus Blur & 47.90 ± 0.08 & 82.35 ± 0.29 & 88.41 ± 0.31 & 89.78 ± 0.61 & 90.22 ± 0.78 & 89.74 ± 0.98 & 88.89 ± 1.54 & 87.42 ± 2.19 & 85.85 ± 2.55 & 83.41 ± 3.10 & 80.48 ± 3.60 &  \\

        \rowcolor[HTML]{EFEFEF}
        Motion Blur & 26.39 ± 0.23 & 71.73 ± 0.89 & 81.10 ± 0.44 & 85.01 ± 0.23 & 86.54 ± 0.49 & 86.81 ± 0.71 & 86.30 ± 1.22 & 84.99 ± 1.56 & 83.24 ± 1.70 & 80.90 ± 2.21 & 78.32 ± 3.18 &  \\
        
        Zoom Blur & 02.47 ± 0.02 & 20.59 ± 0.62 & 27.97 ± 0.62 & 28.10 ± 1.23 & 26.04 ± 2.00 & 22.83 ± 2.44 & 19.69 ± 2.18 & 17.03 ± 2.09 & 14.46 ± 1.82 & 12.63 ± 1.72 & 11.18 ± 1.56 &  \\

        \rowcolor[HTML]{EFEFEF}
        Snow & 33.65 ± 0.01 & 59.62 ± 0.70 & 67.67 ± 0.77 & 70.10 ± 0.36 & 69.55 ± 1.04 & 65.97 ± 1.90 & 59.54 ± 2.56 & 50.03 ± 3.09 & 38.50 ± 2.61 & 25.35 ± 3.58 & 13.46 ± 4.45 &  \\
        
        Frost  & 33.25 ± 0.38 & 63.68 ± 0.18 & 72.31 ± 0.23 & 75.45 ± 0.31 & 76.36 ± 0.33 & 75.87 ± 0.39 & 73.83 ± 0.27 & 70.41 ± 0.37 & 65.10 ± 0.57 & 57.51 ± 1.79 & 47.32 ± 3.45 &  \\

        \rowcolor[HTML]{EFEFEF}
        Fog  & 59.60 ± 0.10 & 85.74 ± 0.04 & 89.79 ± 0.43 & 90.17 ± 0.96 & 88.01 ± 1.99 & 84.51 ± 3.80 & 78.78 ± 6.13 & 72.14 ± 8.00 & 64.47 ± 9.62 & 57.25 ± 10.1 & 50.90 ± 10.0 &  \\

        Brightness & 41.77 ± 0.03 & 75.37 ± 0.17 & 82.38 ± 0.17 & 84.12 ± 0.20 & 84.07 ± 0.58 & 82.10 ± 1.20 & 78.20 ± 1.97 & 71.94 ± 2.81 & 62.81 ± 3.16 & 49.92 ± 4.49 & 35.36 ± 6.97 &  \\

        \rowcolor[HTML]{EFEFEF}
        Contrast & 47.48 ± 0.04 & 58.36 ± 1.99 & 50.59 ± 4.48 & 34.73 ± 5.82 & 20.25 ± 4.48 & 10.57 ± 3.82 & 04.99 ± 2.13 & 02.25 ± 1.17 & 00.77 ± 0.31 & 00.33 ± 0.47 & 00.00 ± 0.00 &  \\

        Elastic transform  & 52.42 ± 0.18 & 84.78 ± 0.42 & 89.92 ± 0.51 & 92.67 ± 0.27 & 94.10 ± 0.05 & 94.72 ± 0.07 & 94.94 ± 0.11 & 94.92 ± 0.17 & 94.32 ± 0.34 & 93.73 ± 0.36 & 92.41 ± 0.93 &  \\

        \rowcolor[HTML]{EFEFEF}
        Pixelate & 03.95 ± 0.01 & 42.41 ± 1.64 & 66.27 ± 3.14 & 76.01 ± 3.53 & 81.29 ± 3.70 & 85.06 ± 3.32 & 87.01 ± 2.91 & 88.18 ± 2.82 & 88.97 ± 2.35 & 88.75 ± 2.39 & 87.69 ± 2.67 &  \\
        
        JPEG compression  & 57.22 ± 0.02 & 82.36 ± 0.73 & 87.87 ± 0.89 & 90.49 ± 1.05 & 91.95 ± 1.11 & 92.59 ± 1.22 & 92.62 ± 1.25 & 92.42 ± 1.20 & 91.86 ± 1.38 & 90.53 ± 1.47 & 88.93 ± 1.69 &  \\

        \midrule
        \rowcolor[HTML]{EFEFEF}
        mPC & 40.96 &  69.51 & 75.82 & 76.98  & 76.47 & 75.08 &  72.88 & 69.94 & 66.15 & 61.53 & 56.40 \\

        \bottomrule
    \end{tabular}
    }

\end{table*}

\begin{table*}[ht]
\caption{\textbf{Ablation of $\lambda$ over FLIR-C dataset for Faster R-CNN. Where $\lambda = 0.0$ represents the zero-shot model, $\lambda = 0.5$ represents default WiSE-OD$_{ZS}$ and $\lambda = 1.0$ represents the fine-tuning model. For FLIR-C, the severity level is $2$.}}
\label{tab:fasterrcnn_flir_ablation_lambda_wiseft}

\label{tab:img2img_main_table}
    \centering
    \resizebox{\textwidth}{!}{%
    \begin{tabular}{lcccccccccccc}

        \toprule \\
        
        {} &  \multicolumn{12}{c}{\textbf{Test Set IR (Dataset: FLIR-C)}} \\
        
        \multirow{2}{*}[-1em]{} & \multirow{2}{*}[-1em]{$\theta(\lambda=0.0)$} & 
        \multirow{2}{*}[-1em]{$\theta(\lambda=0.1)$} & \multirow{2}{*}[-1em]{$\theta(\lambda=0.2)$} & \multirow{2}{*}[-1em]{$\theta(\lambda=0.3)$} & \multirow{2}{*}[-1em]{$\theta(\lambda=0.4)$} & 
        \multirow{2}{*}[-1em]{$\theta(\lambda=0.5)$} & \multirow{2}{*}[-1em]{$\theta(\lambda=0.6)$} & \multirow{2}{*}[-1em]{$\theta(\lambda=0.7)$} & \multirow{2}{*}[-1em]{$\theta(\lambda=0.8)$} & \multirow{2}{*}[-1em]{$\theta(\lambda=0.9)$} & \multirow{2}{*}[-1em]{$\theta(\lambda=1.0)$} \\
        
        \cmidrule(lr){2-12}
        \addlinespace[10pt]        

        \midrule

        Original & 65.52 ± 0.07 & 73.51 ± 0.14 & 77.49 ± 0.10 & 79.74 ± 0.10 & 80.87 ± 0.09 & 82.20 ± 0.07 & 81.31 ± 0.17 & 80.95 ± 0.26 & 80.18 ± 0.11 & 79.03 ± 0.09 & 77.57 ± 0.24 &  \\

        \midrule
        \rowcolor[HTML]{EFEFEF}
         Gaussian Noise & 31.21 ± 0.29 & 38.94 ± 1.77 & 42.62 ± 2.92 & 44.05 ± 3.60 & 44.11 ± 4.31 & 42.49 ± 4.48 & 40.60 ± 4.35 & 37.88 ± 4.13 & 34.85 ± 3.85 & 31.43 ± 3.37 & 28.07 ± 2.91 &  \\

        Shot Noise & 25.26 ± 0.12 & 31.53 ± 1.55 & 33.91 ± 2.98 & 33.86 ± 4.03 & 32.81 ± 4.12 & 30.45 ± 3.96 & 27.89 ± 3.63 & 24.81 ± 3.25 & 21.88 ± 2.93 & 18.79 ± 2.52 & 15.73 ± 2.05 &  \\

        \rowcolor[HTML]{EFEFEF}
        Impulse Noise & 17.69 ± 0.03 & 22.94 ± 1.52 & 24.85 ± 2.26 & 25.00 ± 2.70 & 24.15 ± 2.64 & 22.51 ± 2.78 & 21.18 ± 2.89 & 18.95 ± 2.49 & 16.96 ± 2.17 & 15.15 ± 2.33 & 13.22 ± 2.27 &  \\

        Defocus Blur & 25.32 ± 0.22 & 37.37 ± 1.81 & 44.57 ± 2.40 & 49.45 ± 2.37 & 52.30 ± 2.01 & 54.08 ± 1.74 & 54.99 ± 1.38 & 55.33 ± 1.10 & 55.00 ± 0.98 & 54.01 ± 0.64 & 52.47 ± 0.99 &  \\

        \rowcolor[HTML]{EFEFEF}
        Motion Blur & 25.01 ± 0.25 & 34.24 ± 1.35 & 40.63 ± 2.17 & 45.17 ± 2.38 & 48.75 ± 2.17 & 51.03 ± 2.16 & 53.19 ± 2.13 & 53.96 ± 2.00 & 53.85 ± 2.19 & 53.29 ± 2.18 & 51.71 ± 2.12 &  \\
        
        Zoom Blur & 08.98 ± 0.05 & 11.77 ± 0.49 & 13.72 ± 0.97 & 15.20 ± 1.01 & 16.10 ± 1.01 & 16.93 ± 1.00 & 17.69 ± 1.10 & 18.02 ± 1.11 & 18.32 ± 1.09 & 18.24 ± 0.94 & 17.97 ± 0.90 &  \\

        \rowcolor[HTML]{EFEFEF}
        Snow & 09.84 ± 0.14 & 13.39 ± 1.37 & 14.55 ± 2.07 & 15.19 ± 2.60 & 14.85 ± 2.69 & 13.94 ± 2.66 & 12.73 ± 2.68 & 11.47 ± 2.59 & 10.36 ± 2.48 & 08.99 ± 2.25 & 07.86 ± 2.01 &  \\
        
        Frost  & 21.96 ± 0.50 & 29.17 ± 1.92 & 33.37 ± 2.39 & 35.98 ± 2.87 & 37.34 ± 3.17 & 37.97 ± 3.63 & 37.71 ± 3.59 & 37.22 ± 3.82 & 36.47 ± 4.00 & 35.17 ± 4.45 & 33.87 ± 4.69 &  \\

        \rowcolor[HTML]{EFEFEF}
        Fog  & 56.36 ± 0.28 & 67.20 ± 1.18 & 72.17 ± 0.86 & 75.52 ± 0.67 & 77.56 ± 0.97 & 78.68 ± 1.24 & 78.85 ± 1.00 & 78.71 ± 1.23 & 78.11 ± 1.49 & 76.92 ± 1.12 & 73.61 ± 0.06 &  \\

        Brightness & 64.41 ± 0.26 & 71.92 ± 0.51 & 75.68 ± 0.19 & 75.60 ± 0.13 & 79.09 ± 0.49 & 79.72 ± 0.35 & 79.53 ± 0.96 & 78.56 ± 1.28 & 77.79 ± 1.24 & 76.54 ± 1.33 & 75.18 ± 0.99 &  \\

        \rowcolor[HTML]{EFEFEF}
        Contrast & 54.59 ± 0.04 & 66.11 ± 0.79 & 71.38 ± 0.95 & 74.78 ± 1.19 & 77.03 ± 0.90 & 78.36 ± 1.06 & 78.62 ± 1.12 & 78.36 ± 0.99 & 78.02 ± 1.09 & 76.87 ± 1.15 & 75.47 ± 1.29 &  \\

        Elastic transform  & 41.88 ± 0.24 & 55.93 ± 0.47 & 63.89 ± 0.37 & 68.62 ± 0.39 & 71.71 ± 0.79 & 73.39 ± 0.40 & 73.66 ± 0.41 & 73.37 ± 0.25 & 72.51 ± 0.58 & 71.51 ± 0.42 & 69.68 ± 1.15 &  \\

        \rowcolor[HTML]{EFEFEF}
        Pixelate & 38.67 ± 0.11 & 49.47 ± 1.66 & 55.23 ± 2.37 & 58.37 ± 2.52 & 60.02 ± 2.70 & 61.12 ± 3.13 & 60.93 ± 3.43 & 60.12 ± 4.13 & 58.87 ± 4.58 & 57.14 ± 5.60 & 54.91 ± 6.21 &  \\
        
        JPEG compression  & 50.24 ± 0.14 & 59.12 ± 0.61 & 63.21 ± 0.56 & 65.59 ± 0.63 & 66.55 ± 0.80 & 66.65 ± 0.89 & 66.12 ± 1.48 & 64.94 ± 1.84 & 63.27 ± 2.36 & 60.63 ± 2.39 & 57.55 ± 3.04 &  \\

        \midrule
        \rowcolor[HTML]{EFEFEF}
        mPC & 33.67 & 42.07 & 46.41 & 48.74 & 50.16 & 50.52 & 50.26 & 49.40 & 48.30 & 46.76 & 44.80 \\
        
        \bottomrule
    \end{tabular}
    }

\end{table*}

\begin{table*}[ht]
\caption{\textbf{Ablation of $\lambda$ over FLIR-C dataset for FCOS. Where $\lambda = 0.0$ represents the zero-shot model, $\lambda = 0.5$ represents default WiSE-OD$_{ZS}$ and $\lambda = 1.0$ represents the fine-tuning model. For FLIR-C, the severity level is $2$.}}
\label{tab:fcos_flir_ablation_lambda_wiseft}
    \centering
    \resizebox{\textwidth}{!}{%
    \begin{tabular}{lcccccccccccc}

        \toprule \\
        
        {} &  \multicolumn{12}{c}{\textbf{Test Set IR (Dataset: FLIR-C)}} \\
        
        \multirow{2}{*}[-1em]{} & \multirow{2}{*}[-1em]{$\theta(\lambda=0.0)$} & 
        \multirow{2}{*}[-1em]{$\theta(\lambda=0.1)$} & \multirow{2}{*}[-1em]{$\theta(\lambda=0.2)$} & \multirow{2}{*}[-1em]{$\theta(\lambda=0.3)$} & \multirow{2}{*}[-1em]{$\theta(\lambda=0.4)$} & 
        \multirow{2}{*}[-1em]{$\theta(\lambda=0.5)$} & \multirow{2}{*}[-1em]{$\theta(\lambda=0.6)$} & \multirow{2}{*}[-1em]{$\theta(\lambda=0.7)$} & \multirow{2}{*}[-1em]{$\theta(\lambda=0.8)$} & \multirow{2}{*}[-1em]{$\theta(\lambda=0.9)$} & \multirow{2}{*}[-1em]{$\theta(\lambda=1.0)$} \\
        
        \cmidrule(lr){2-12}
        \addlinespace[10pt]        

        \midrule

        Original & 59.90 ± 0.00 & 69.17 ± 0.00 & 73.37 ± 0.70 & 75.92 ± 0.61 & 76.42 ± 0.00 & 77.82 ± 0.00 & 76.13 ± 0.70 & 75.64 ± 1.17 & 73.84 ± 0.00 & 73.18 ± 1.00 & 71.78 ± 1.27 &  \\

      \midrule
        \rowcolor[HTML]{EFEFEF}
         Gaussian Noise & 23.65 ± 0.16 & 35.87 ± 0.60 & 39.92 ± 0.34 & 40.94 ± 0.69 & 40.21 ± 1.22 & 38.53 ± 1.47 & 36.49 ± 1.58 & 33.92 ± 1.84 & 31.18 ± 1.93 & 27.93 ± 2.20 & 24.76 ± 2.44 &  \\

        Shot Noise & 17.63 ± 0.26 & 28.36 ± 0.75 & 31.61 ± 0.30 & 31.49 ± 0.15 & 30.10 ± 0.62 & 27.85 ± 1.06 & 25.41 ± 1.18 & 22.97 ± 1.38 & 20.49 ± 1.46 & 17.95 ± 1.66 & 15.39 ± 1.80 &  \\

        \rowcolor[HTML]{EFEFEF}
        Impulse Noise & 14.12 ± 0.23 & 22.48 ± 0.30 & 24.23 ± 0.54 & 23.70 ± 0.82 & 22.94 ± 1.06 & 21.07 ± 1.25 & 19.60 ± 0.93 & 17.57 ± 0.84 & 15.50 ± 0.70 & 13.98 ± 0.97 & 11.34 ± 0.88 &  \\

        Defocus Blur & 19.23 ± 0.00 & 32.58 ± 0.60 & 39.97 ± 0.61 & 44.29 ± 1.13 & 46.66 ± 1.10 & 48.14 ± 1.00 & 48.96 ± 1.24 & 48.92 ± 1.33 & 47.87 ± 1.58 & 46.38 ± 1.54 & 44.40 ± 1.40 &  \\

        \rowcolor[HTML]{EFEFEF}
        Motion Blur & 20.77 ± 0.21 & 31.73 ± 0.59 & 37.81 ± 0.69 & 41.68 ± 1.03 & 44.17 ± 1.43 & 45.65 ± 1.29 & 46.54 ± 0.90 & 46.60 ± 0.96 & 46.01 ± 1.07 & 44.90 ± 1.19 & 43.17 ± 1.44 &  \\
        
        Zoom Blur & 06.73 ± 0.00 & 11.12 ± 0.35 & 13.23 ± 0.43 & 14.27 ± 0.52 & 14.94 ± 0.54 & 15.44 ± 0.52 & 15.70 ± 0.51 & 15.89 ± 0.18 & 16.06 ± 0.19 & 15.91 ± 0.15 & 15.32 ± 0.14 &  \\

        \rowcolor[HTML]{EFEFEF}
        Snow & 08.29 ± 0.32 & 11.98 ± 0.26 & 13.57 ± 0.41 & 13.62 ± 0.71 & 13.19 ± 1.05 & 12.50 ± 1.08 & 11.63 ± 1.20 & 10.85 ± 1.26 & 09.83 ± 1.24 & 08.73 ± 1.22 & 07.63 ± 1.31 &  \\
        
        Frost  & 19.23 ± 0.26 & 28.62 ± 0.65 & 32.77 ± 0.39 & 34.68 ± 0.25 & 35.33 ± 0.38 & 35.68 ± 0.55 & 35.44 ± 0.77 & 34.87 ± 0.89 & 33.96 ± 0.73 & 32.55 ± 0.68 & 30.80 ± 0.69 &  \\

        \rowcolor[HTML]{EFEFEF}
        Fog  & 51.56 ± 0.07 & 63.39 ± 0.15 & 68.76 ± 0.18 & 72.17 ± 0.55 & 73.84 ± 0.71 & 74.39 ± 0.84 & 74.45 ± 0.85 & 73.36 ± 1.40 & 72.63 ± 1.28 & 71.42 ± 1.25 & 70.56 ± 0.41 &  \\

        Brightness & 58.43 ± 0.00 & 67.75 ± 0.15 & 72.80 ± 0.42 & 75.20 ± 0.50 & 75.88 ± 0.88 & 75.59 ± 0.94 & 74.73 ± 0.90 & 73.78 ± 1.00 & 72.72 ± 1.04 & 71.21 ± 1.10 & 69.69 ± 1.05 &  \\

        \rowcolor[HTML]{EFEFEF}
        Contrast & 50.28 ± 0.00 & 62.55 ± 0.12 & 68.03 ± 0.30 & 71.68 ± 0.53 & 73.50 ± 0.67 & 73.99 ± 0.69 & 73.83 ± 1.02 & 73.28 ± 1.39 & 72.75 ± 1.12 & 71.68 ± 1.15 & 70.41 ± 0.96 &  \\

        Elastic transform  & 36.76 ± 0.48 & 53.13 ± 0.82 & 61.25 ± 0.88 & 65.65 ± 0.70 & 67.93 ± 0.34 & 68.95 ± 0.52 & 68.84 ± 0.48 & 68.30 ± 0.81 & 67.21 ± 1.01 & 65.96 ± 1.01 & 64.18 ± 0.74 &  \\

        \rowcolor[HTML]{EFEFEF}
        Pixelate & 32.65 ± 0.00 & 45.53 ± 0.92 & 52.09 ± 0.78 & 55.86 ± 0.74 & 57.68 ± 0.95 & 59.03 ± 0.09 & 57.56 ± 0.54 & 57.57 ± 0.67 & 56.31 ± 0.63 & 54.36 ± 0.82 & 51.74 ± 0.86 &  \\
        
        JPEG compression  & 44.71 ± 0.00 & 55.25 ± 0.38 & 59.69 ± 0.05 & 62.18 ± 0.29 & 63.13 ± 0.58 & 63.03 ± 0.74 & 62.49 ± 0.45 & 61.14 ± 0.36 & 59.44 ± 0.34 & 57.37 ± 0.64 & 55.63 ± 0.96 &  \\

        \midrule
        \rowcolor[HTML]{EFEFEF}
        mPC & 28.85 & 39.31 & 43.98 & 46.24 & 47.10 & 47.13 & 46.54 & 45.64 & 44.42 & 42.88 & 41.07 & \\

        \bottomrule
    \end{tabular}
    }

\end{table*}

\begin{table*}[ht]
\caption{\textbf{Ablation of $\lambda$ over FLIR-C dataset for RetinaNet. Where $\lambda = 0.0$ represents the zero-shot model, $\lambda = 0.5$ represents default WiSE-OD$_{ZS}$ and $\lambda = 1.0$ represents the fine-tuning model. For FLIR-C, the severity level is $2$.}}
\label{tab:retinanet_flir_ablation_lambda_wiseft}
    \centering
    \resizebox{\textwidth}{!}{%
    \begin{tabular}{lcccccccccccc}

        \toprule \\
        
        {} &  \multicolumn{12}{c}{\textbf{Test Set IR (Dataset: FLIR-C)}} \\
        
        \multirow{2}{*}[-1em]{} & \multirow{2}{*}[-1em]{$\theta(\lambda=0.0)$} & 
        \multirow{2}{*}[-1em]{$\theta(\lambda=0.1)$} & \multirow{2}{*}[-1em]{$\theta(\lambda=0.2)$} & \multirow{2}{*}[-1em]{$\theta(\lambda=0.3)$} & \multirow{2}{*}[-1em]{$\theta(\lambda=0.4)$} & 
        \multirow{2}{*}[-1em]{$\theta(\lambda=0.5)$} & \multirow{2}{*}[-1em]{$\theta(\lambda=0.6)$} & \multirow{2}{*}[-1em]{$\theta(\lambda=0.7)$} & \multirow{2}{*}[-1em]{$\theta(\lambda=0.8)$} & \multirow{2}{*}[-1em]{$\theta(\lambda=0.9)$} & \multirow{2}{*}[-1em]{$\theta(\lambda=1.0)$} \\
        
        \cmidrule(lr){2-12}
        \addlinespace[10pt]        

        \midrule

        Original & 58.46 ± 0.00 & 67.08 ± 0.22 & 71.71 ± 0.32 & 74.93 ± 0.15 & 76.38 ± 0.62 & 77.34 ± 0.05 & 77.00 ± 0.19 & 76.88 ± 0.01 & 76.08 ± 0.15 & 74.35 ± 0.30 & 73.38 ± 0.89 &  \\
        
        \midrule
        \rowcolor[HTML]{EFEFEF}
         Gaussian Noise & 25.61 ± 0.20 & 32.69 ± 0.58 & 35.32 ± 0.84 & 35.81 ± 1.74 & 35.33 ± 2.47 & 34.31 ± 2.83 & 33.28 ± 2.92 & 31.93 ± 3.02 & 30.34 ± 3.13 & 28.58 ± 3.11 & 26.47 ± 3.14 &  \\

        Shot Noise & 19.36 ± 0.06 & 25.03 ± 0.61 & 26.67 ± 0.46 & 26.40 ± 1.27 & 25.48 ± 2.02 & 24.15 ± 2.42 & 22.78 ± 2.53 & 21.43 ± 2.41 & 19.98 ± 2.53 & 18.43 ± 2.63 & 16.95 ± 2.60 &  \\

        \rowcolor[HTML]{EFEFEF}
        Impulse Noise & 14.82 ± 0.17 & 20.28 ± 0.66 & 22.12 ± 0.18 & 22.72 ± 0.54 & 21.78 ± 1.37 & 21.77 ± 1.50 & 20.62 ± 1.79 & 19.08 ± 1.54 & 17.98 ± 1.93 & 16.96 ± 2.15 & 15.90 ± 1.98 &  \\

        Defocus Blur & 19.30 ± 0.00 & 29.18 ± 0.55 & 35.62 ± 0.32 & 39.61 ± 0.79 & 42.52 ± 1.27 & 44.74 ± 1.61 & 46.50 ± 1.67 & 47.22 ± 1.77 & 47.46 ± 1.81 & 47.04 ± 1.84 & 46.15 ± 1.86 &  \\

        \rowcolor[HTML]{EFEFEF}
        Motion Blur & 20.05 ± 0.16 & 26.99 ± 0.15 & 32.06 ± 0.46 & 36.34 ± 1.21 & 39.83 ± 2.32 & 42.91 ± 2.90 & 45.13 ± 3.39 & 46.75 ± 3.63 & 47.25 ± 3.87 & 46.89 ± 3.94 & 45.96 ± 3.97 &  \\
        
        Zoom Blur & 07.05 ± 0.00 & 10.04 ± 0.26 & 11.99 ± 0.40 & 13.25 ± 0.26 & 14.43 ± 0.28 & 15.40 ± 0.32 & 16.07 ± 0.52 & 16.43 ± 0.65 & 16.52 ± 0.81 & 16.37 ± 0.93 & 15.99 ± 1.06 &  \\

        \rowcolor[HTML]{EFEFEF}
        Snow & 07.91 ± 0.16 & 09.65 ± 0.25 & 10.62 ± 0.40 & 11.12 ± 0.85 & 11.29 ± 1.21 & 11.09 ± 1.44 & 10.83 ± 1.73 & 10.40 ± 1.91 & 09.79 ± 1.88 & 09.06 ± 1.91 & 08.34 ± 1.84 &  \\
        
        Frost  & 17.63 ± 0.32 & 23.01 ± 0.57 & 26.44 ± 0.96 & 29.27 ± 1.55 & 31.40 ± 1.95 & 33.02 ± 2.41 & 34.19 ± 2.59 & 34.95 ± 2.84 & 34.91 ± 3.13 & 34.53 ± 3.04 & 33.90 ± 3.17 &  \\

        \rowcolor[HTML]{EFEFEF}
        Fog  & 48.95 ± 0.13 & 58.32 ± 0.38 & 64.25 ± 0.24 & 68.23 ± 0.25 & 70.96 ± 0.45 & 72.46 ± 0.50 & 73.24 ± 0.26 & 73.41 ± 0.45 & 72.88 ± 0.24 & 71.92 ± 0.28 & 70.48 ± 0.40 &  \\

        Brightness & 56.74 ± 0.00 & 65.14 ± 0.27 & 69.99 ± 0.39 & 72.90 ± 0.28 & 74.43 ± 0.64 & 74.80 ± 0.56 & 74.84 ± 0.55 & 74.02 ± 0.40 & 73.09 ± 0.11 & 71.80 ± 0.16 & 70.06 ± 0.30 &  \\

        \rowcolor[HTML]{EFEFEF}
        Contrast & 47.50 ± 0.00 & 56.81 ± 0.50 & 62.87 ± 0.31 & 67.26 ± 0.07 & 70.04 ± 0.23 & 71.89 ± 0.43 & 72.81 ± 0.27 & 73.09 ± 0.20 & 72.80 ± 0.10 & 71.94 ± 0.12 & 70.75 ± 0.38 &  \\

        Elastic transform  & 34.27 ± 0.34 & 48.18 ± 0.94 & 57.07 ± 0.64 & 62.39 ± 0.47 & 65.90 ± 0.29 & 68.19 ± 0.54 & 69.25 ± 0.68 & 69.51 ± 0.44 & 68.86 ± 0.53 & 67.56 ± 0.53 & 65.63 ± 0.54 &  \\

        \rowcolor[HTML]{EFEFEF}
        Pixelate & 32.52 ± 0.00 & 41.93 ± 0.67 & 48.02 ± 0.74 & 52.27 ± 0.66 & 55.46 ± 0.47 & 57.66 ± 0.30 & 58.83 ± 0.15 & 59.10 ± 0.40 & 58.58 ± 0.58 & 57.48 ± 0.82 & 55.52 ± 1.21 &  \\
        
        JPEG compression  & 44.10 ± 0.00 & 52.89 ± 0.51 & 57.47 ± 0.47 & 60.21 ± 0.60 & 61.79 ± 0.86 & 62.57 ± 1.02 & 62.33 ± 0.77 & 61.31 ± 0.63 & 59.89 ± 0.65 & 58.10 ± 0.54 & 55.89 ± 0.86 &  \\

        \midrule
        \rowcolor[HTML]{EFEFEF}
        mPC & 28.27 & 35.72 & 40.03 & 42.69 & 44.33 & 45.35 & 45.76 & 45.61 & 45.02 & 44.04 & 42.71 & \\
        
        \bottomrule
    \end{tabular}
    }

\end{table*}

\section{Performance over different corruption levels}
\label{sec:diff_corruption_levels}

\begin{figure*}
\captionsetup[subfigure]{labelformat=empty}
\centering

\begin{subfigure}[t]{0.60\columnwidth}
    
    \includegraphics[width=\columnwidth]{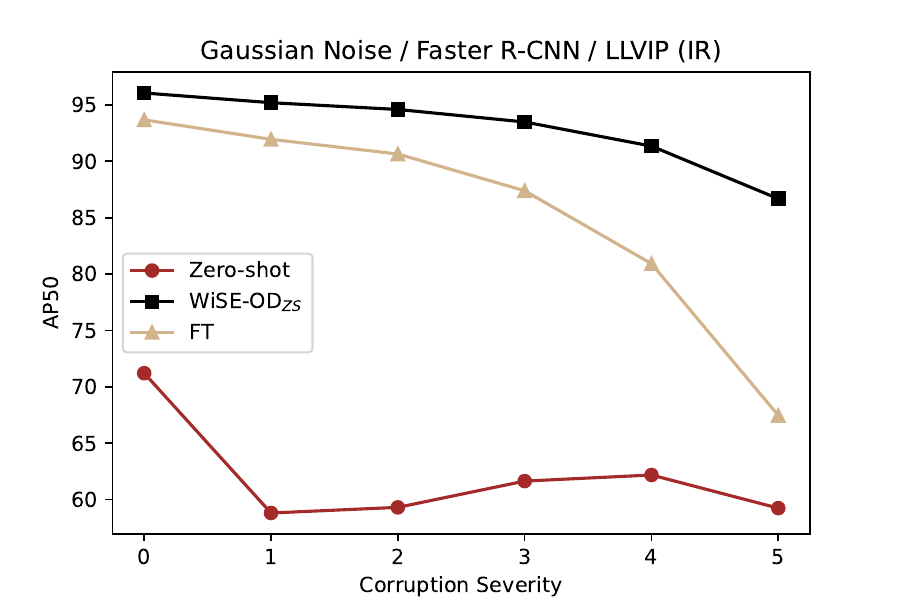}
\end{subfigure}
\begin{subfigure}[t]{0.60\columnwidth}
    
    \includegraphics[width=\columnwidth]{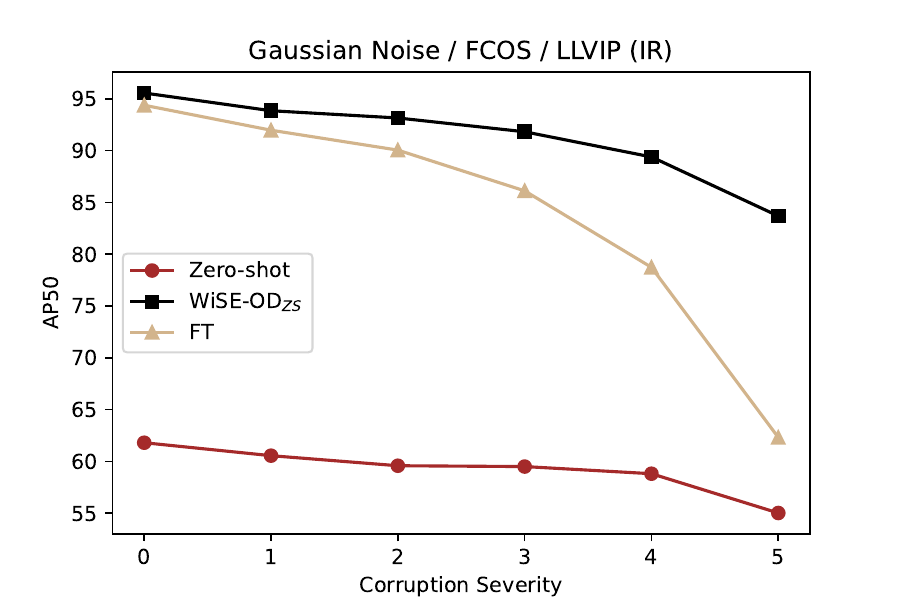}
\end{subfigure}
\begin{subfigure}[t]{0.60\columnwidth}
    
    \includegraphics[width=\columnwidth]{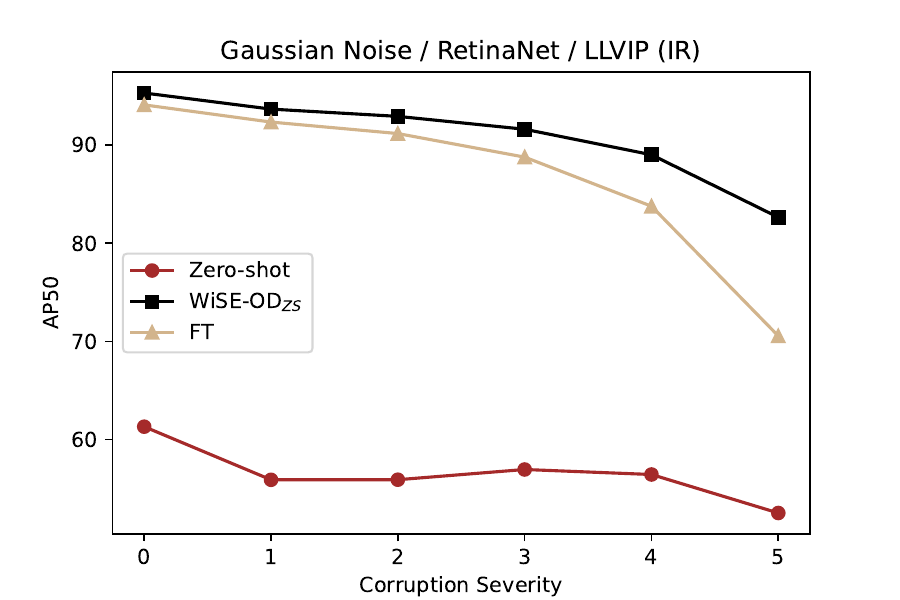}
\end{subfigure}

\begin{subfigure}[t]{0.60\columnwidth}
    
    \includegraphics[width=\columnwidth]{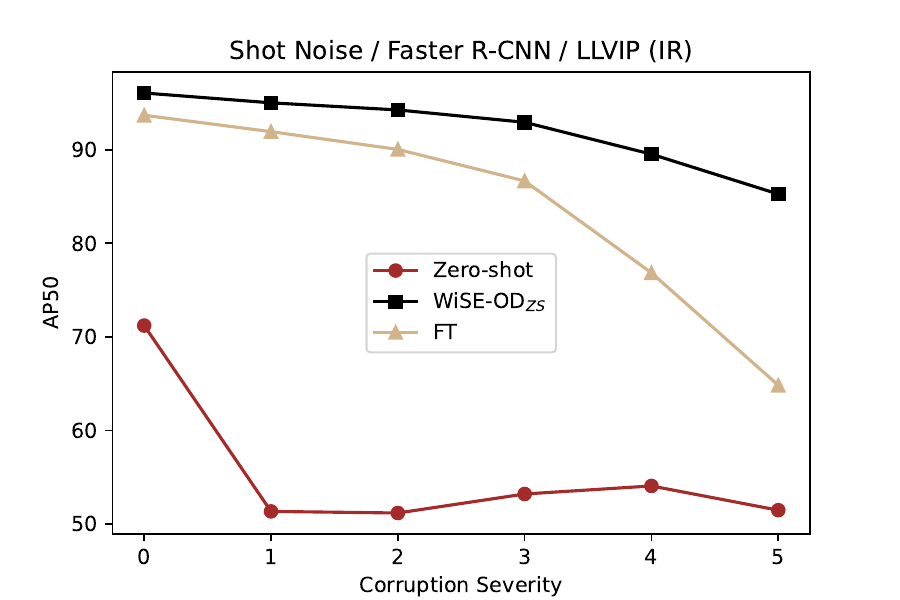}
\end{subfigure}
\begin{subfigure}[t]{0.60\columnwidth}
    
    \includegraphics[width=\columnwidth]{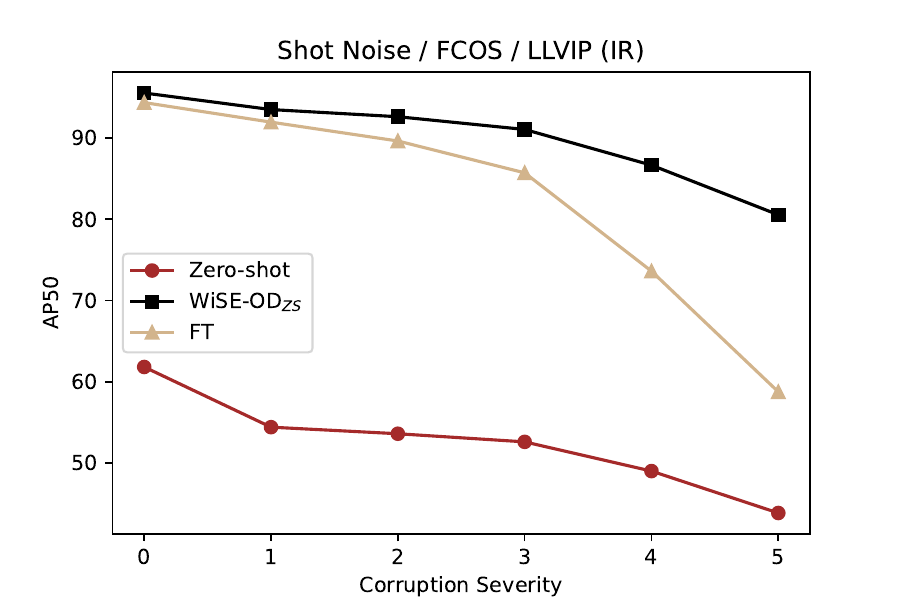}
\end{subfigure}
\begin{subfigure}[t]{0.60\columnwidth}
    
    \includegraphics[width=\columnwidth]{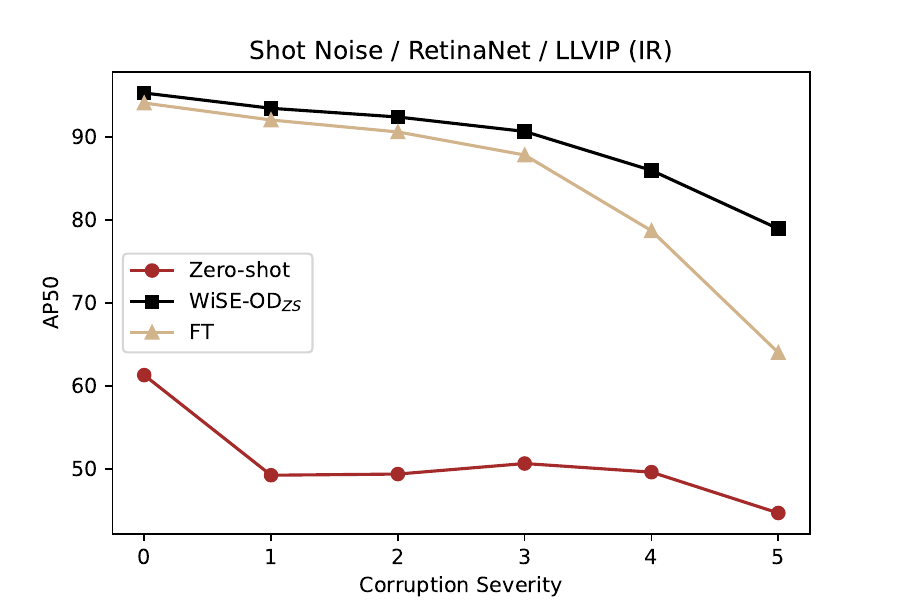}
\end{subfigure}

\begin{subfigure}[t]{0.60\columnwidth}
    
    \includegraphics[width=\columnwidth]{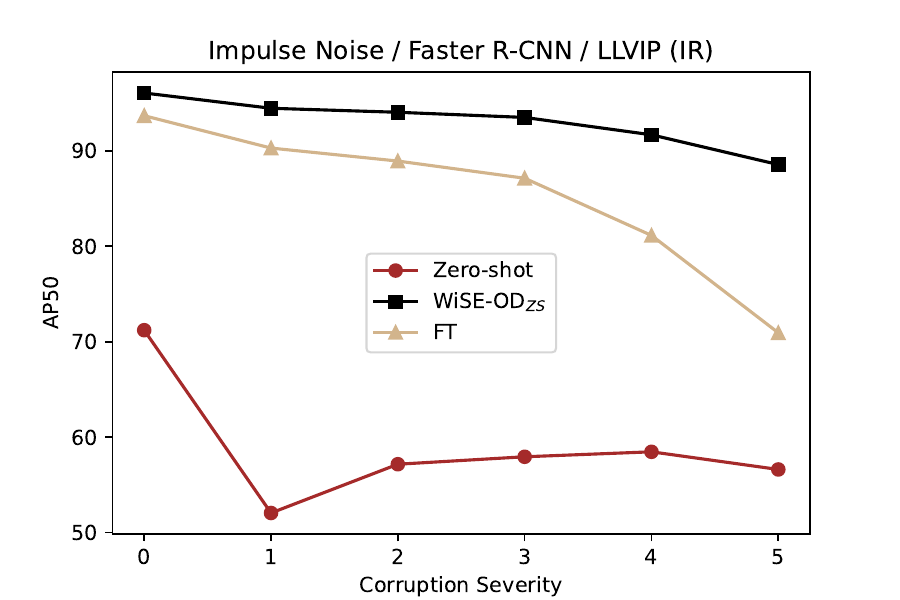}
\end{subfigure}
\begin{subfigure}[t]{0.60\columnwidth}
    
    \includegraphics[width=\columnwidth]{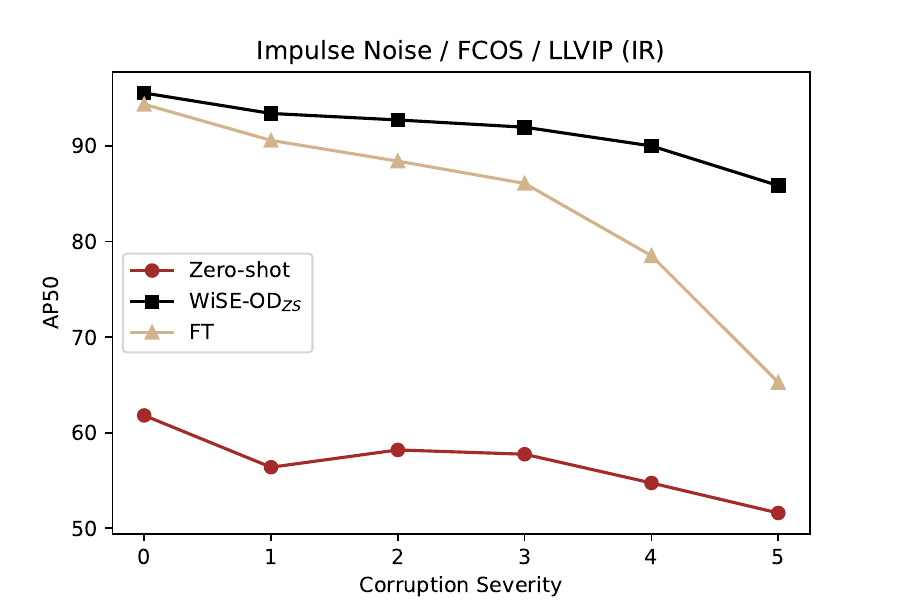}
\end{subfigure}
\begin{subfigure}[t]{0.60\columnwidth}
    
    \includegraphics[width=\columnwidth]{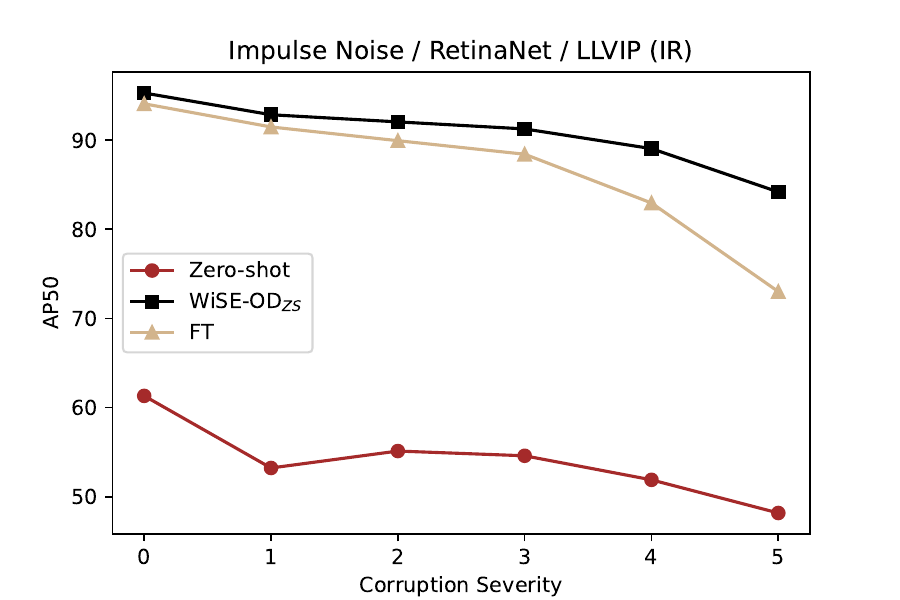}
\end{subfigure}

\begin{subfigure}[t]{0.60\columnwidth}
    
    \includegraphics[width=\columnwidth]{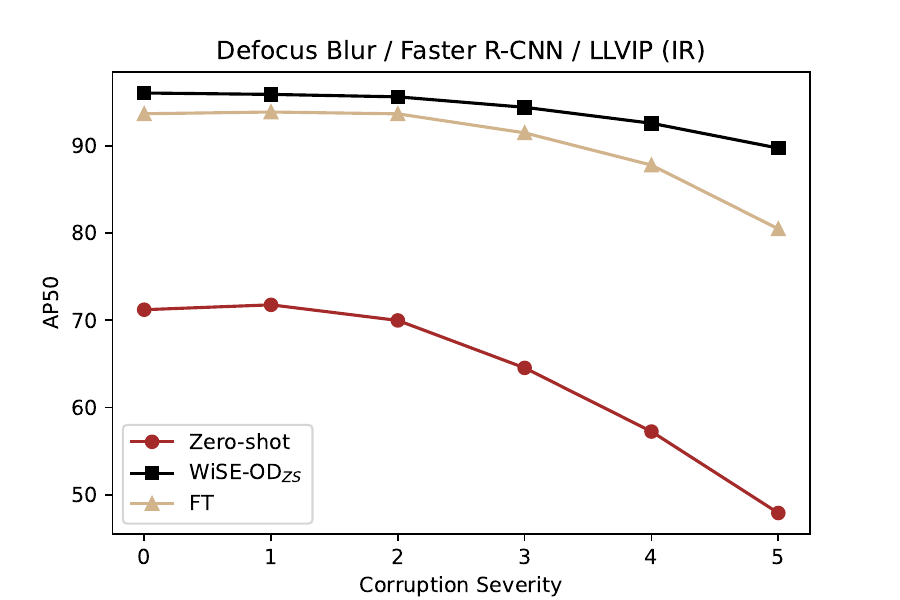}
\end{subfigure}
\begin{subfigure}[t]{0.60\columnwidth}
    
    \includegraphics[width=\columnwidth]{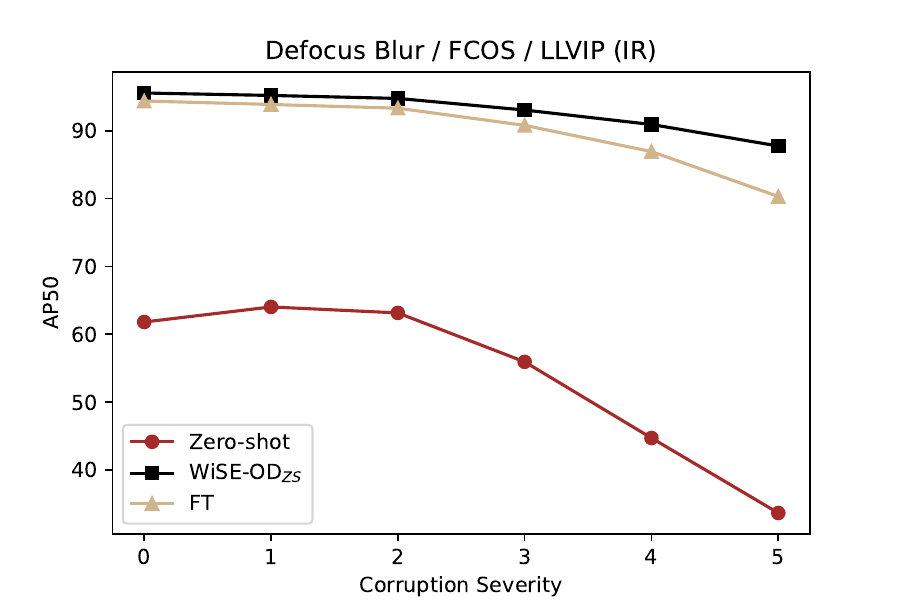}
\end{subfigure}
\begin{subfigure}[t]{0.60\columnwidth}
    
    \includegraphics[width=\columnwidth]{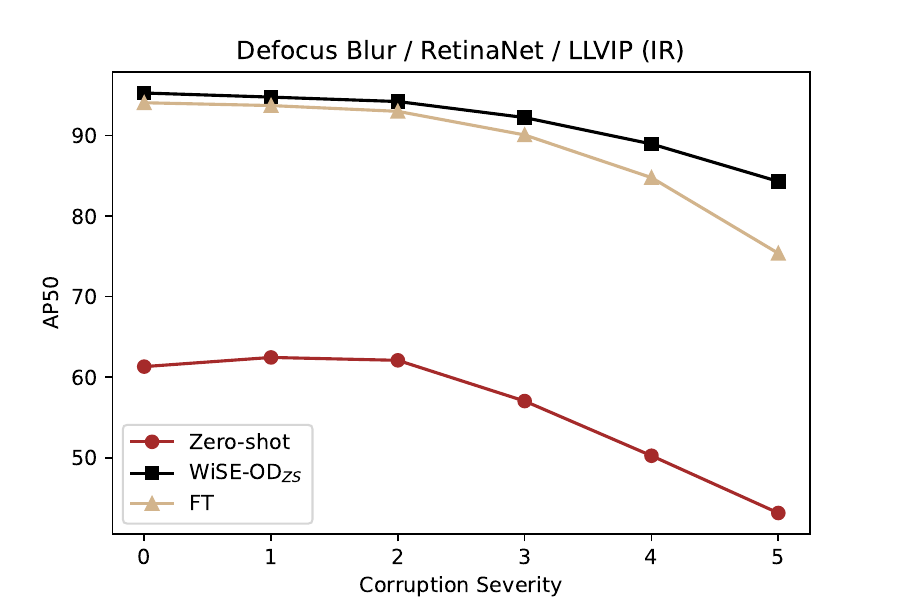}
\end{subfigure}

\begin{subfigure}[t]{0.60\columnwidth}
    
    \includegraphics[width=\columnwidth]{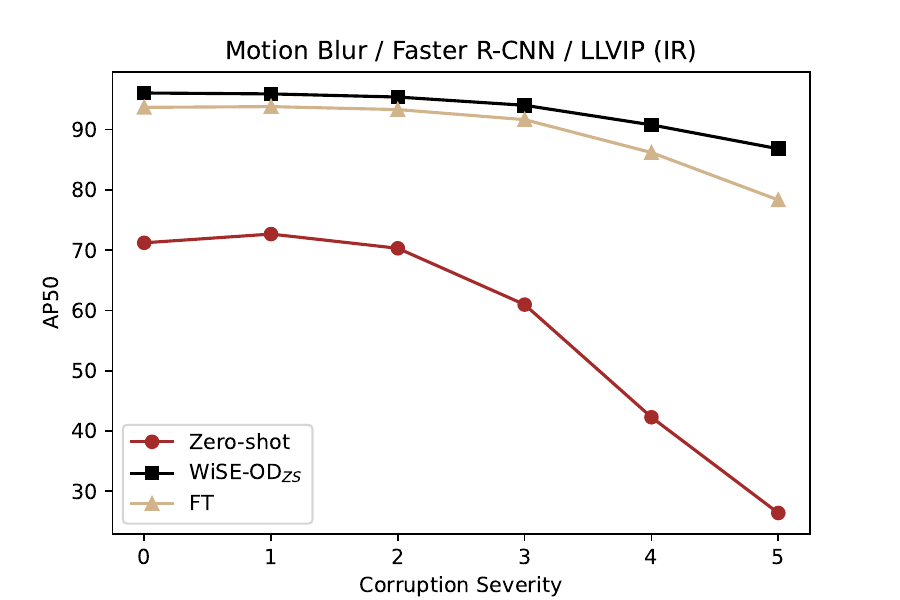}
\end{subfigure}
\begin{subfigure}[t]{0.60\columnwidth}
    
    \includegraphics[width=\columnwidth]{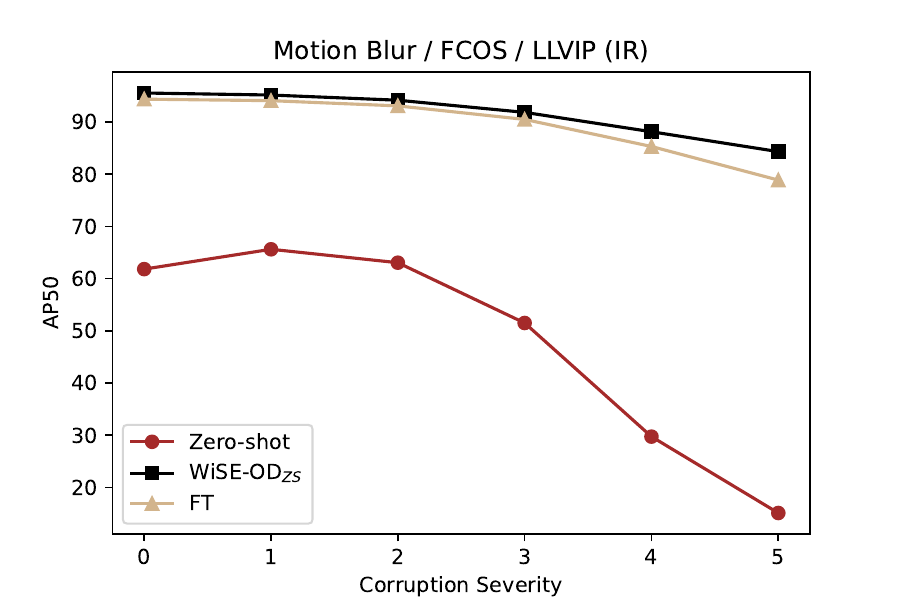}
\end{subfigure}
\begin{subfigure}[t]{0.60\columnwidth}
    
    \includegraphics[width=\columnwidth]{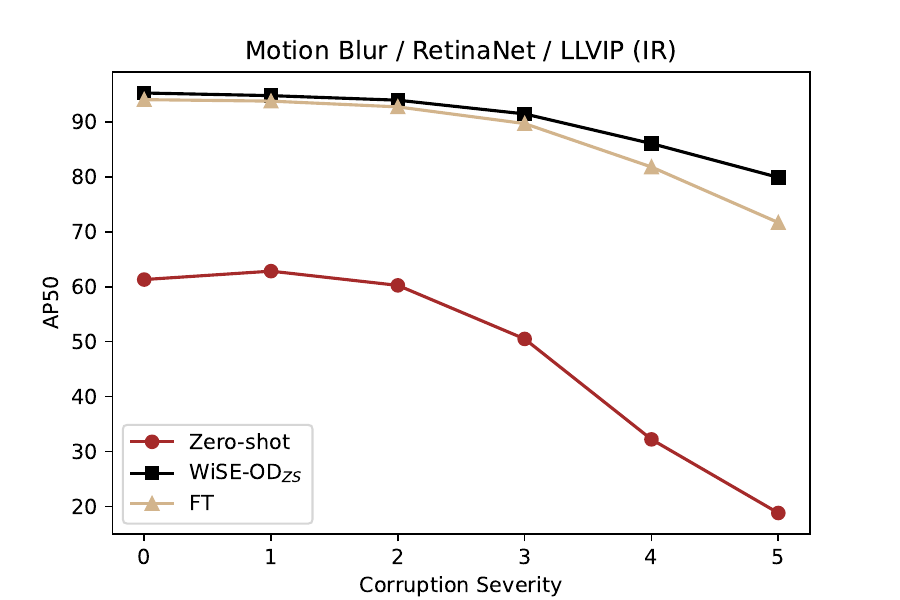}
\end{subfigure}

\begin{subfigure}[t]{0.60\columnwidth}
    
    \includegraphics[width=\columnwidth]{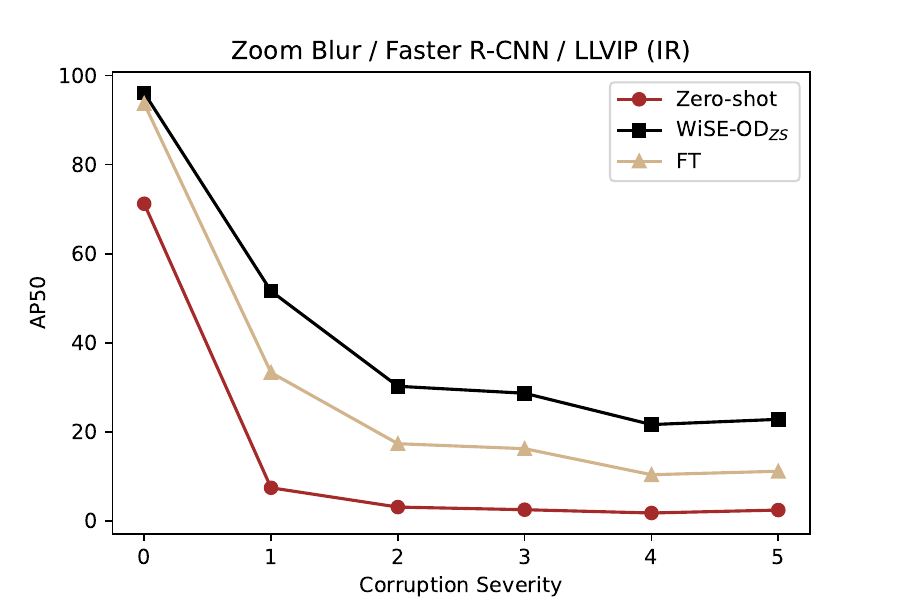}
\end{subfigure}
\begin{subfigure}[t]{0.60\columnwidth}
    
    \includegraphics[width=\columnwidth]{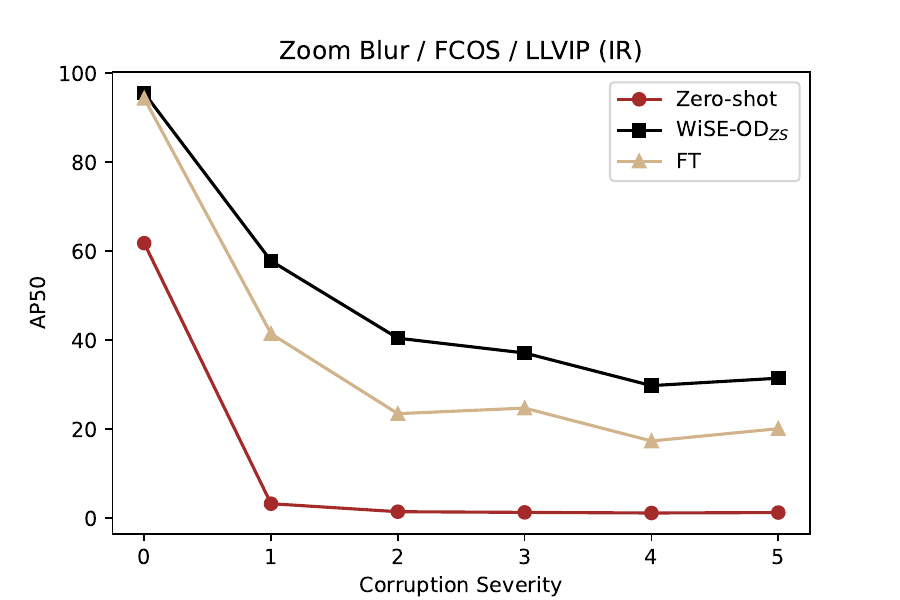}
\end{subfigure}
\begin{subfigure}[t]{0.60\columnwidth}
    
    \includegraphics[width=\columnwidth]{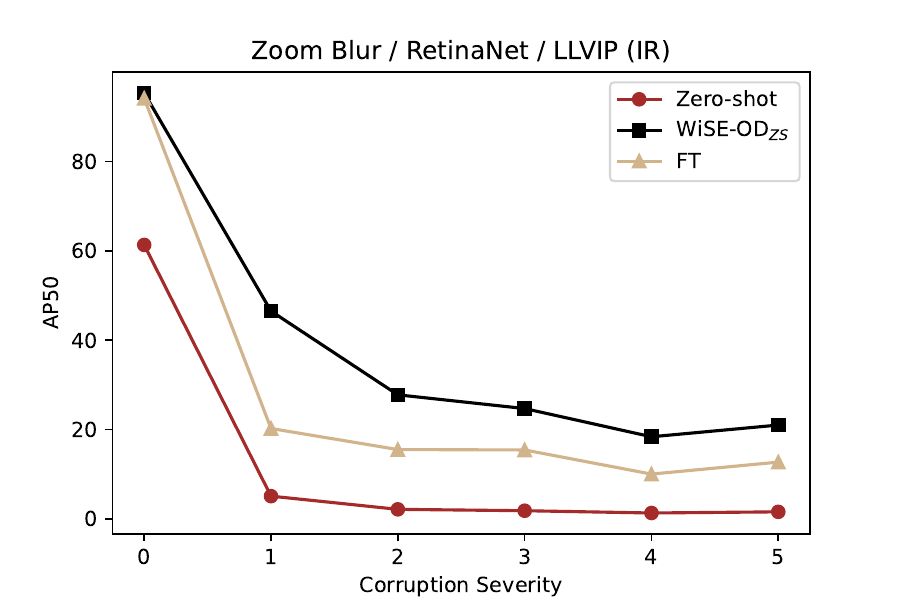}
\end{subfigure}

\caption{\textbf{AP$_{50}$ performance for all detectors over different corruption severity levels for Gaussian Blur, Shot Noise, Impulse Noise, Defocus Blur, Motion Blur and Zoom Blur.} For each perturbation, we evaluated different levels of corruption for the Zero-Shot, WiSE-OD$_{ZS}$, and FT models for LLVIP-C.}

\label{fig:plots_llvip_part1}

\end{figure*}

\begin{figure*}
\captionsetup[subfigure]{labelformat=empty}
\centering

\begin{subfigure}[t]{0.60\columnwidth}
    
    \includegraphics[width=\columnwidth]{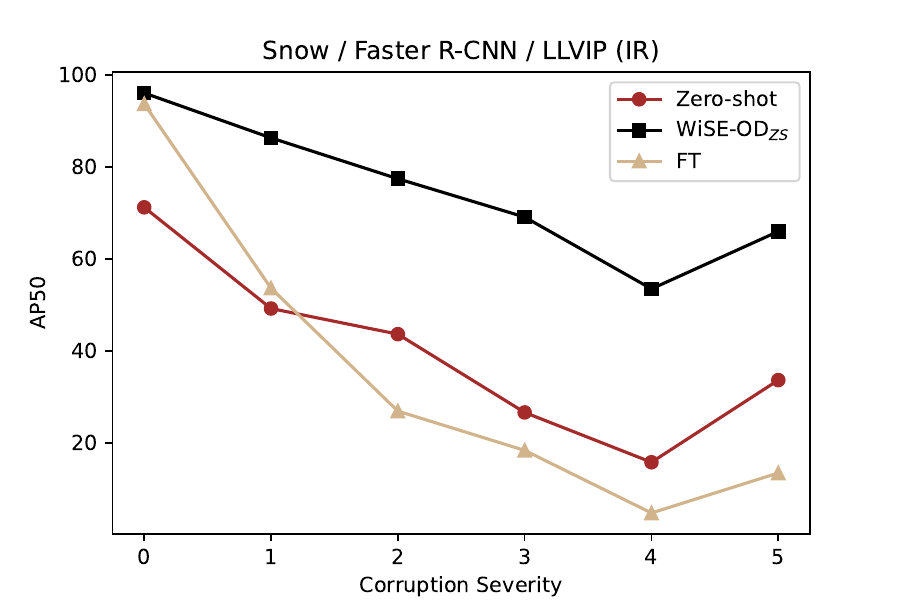}
\end{subfigure}
\begin{subfigure}[t]{0.60\columnwidth}
    
    \includegraphics[width=\columnwidth]{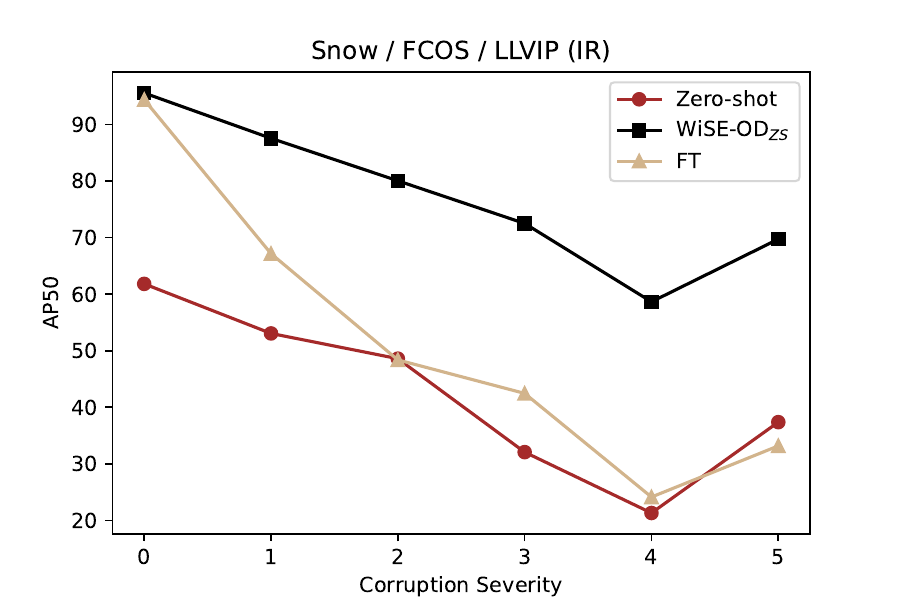}
\end{subfigure}
\begin{subfigure}[t]{0.60\columnwidth}
    
    \includegraphics[width=\columnwidth]{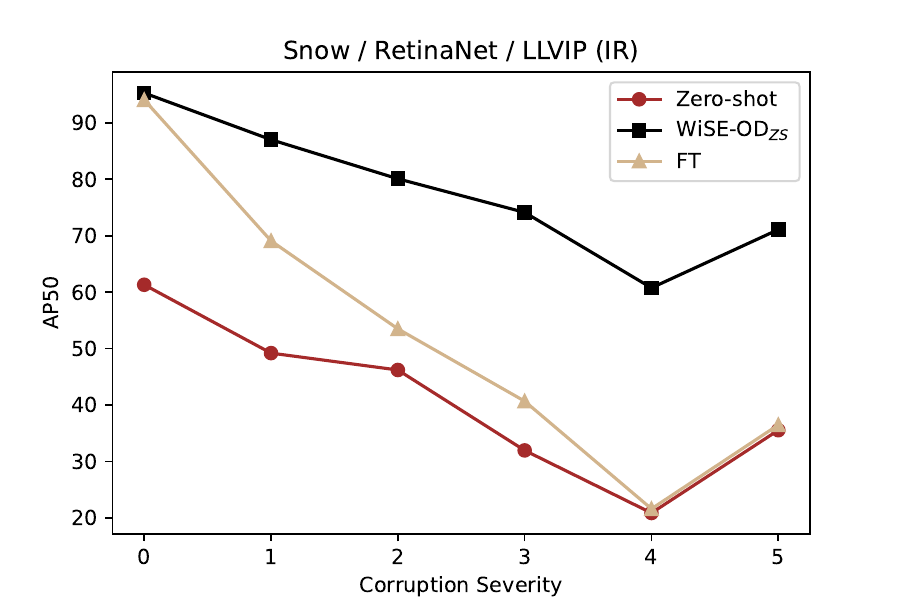}
\end{subfigure}

\begin{subfigure}[t]{0.60\columnwidth}
    
    \includegraphics[width=\columnwidth]{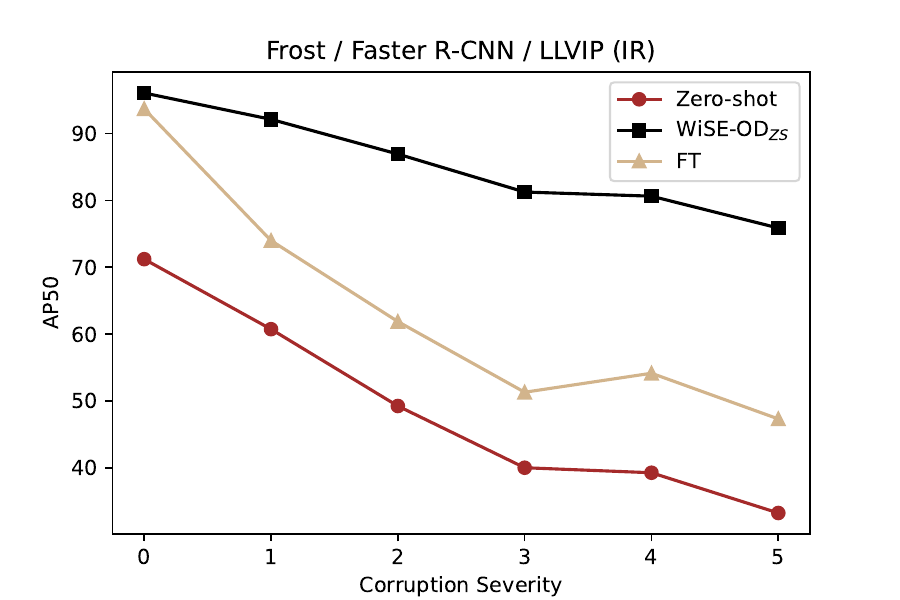}
\end{subfigure}
\begin{subfigure}[t]{0.60\columnwidth}
    
    \includegraphics[width=\columnwidth]{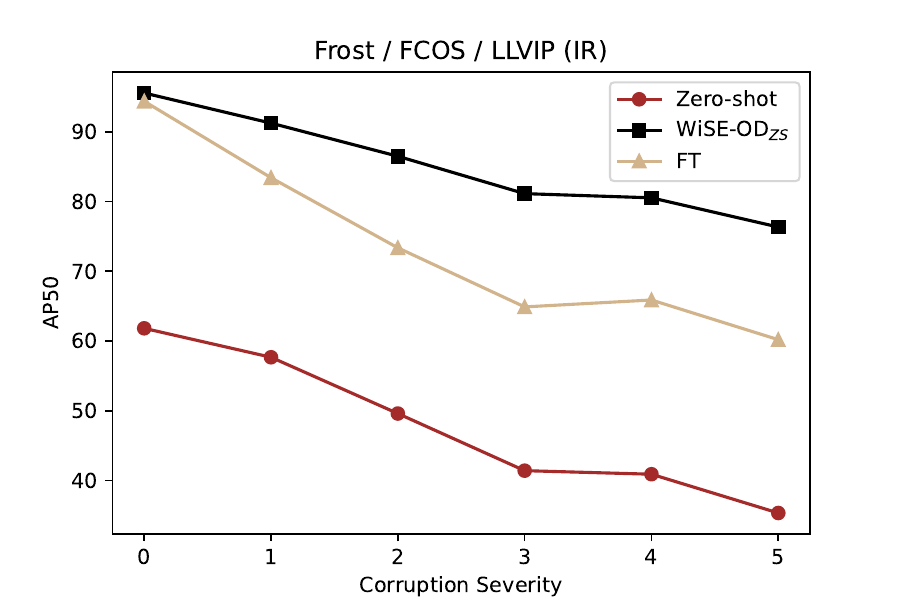}
\end{subfigure}
\begin{subfigure}[t]{0.60\columnwidth}
    
    \includegraphics[width=\columnwidth]{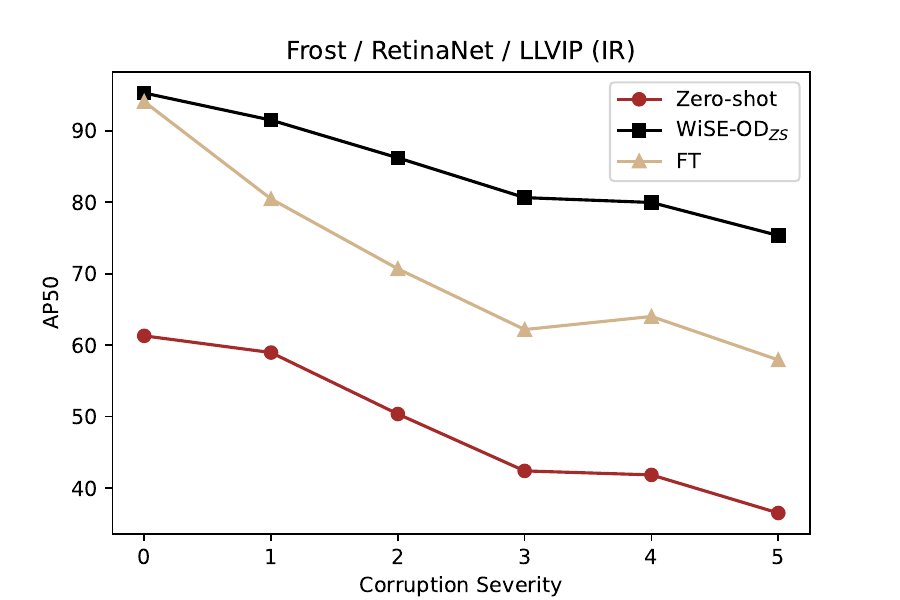}
\end{subfigure}

\begin{subfigure}[t]{0.60\columnwidth}
    
    \includegraphics[width=\columnwidth]{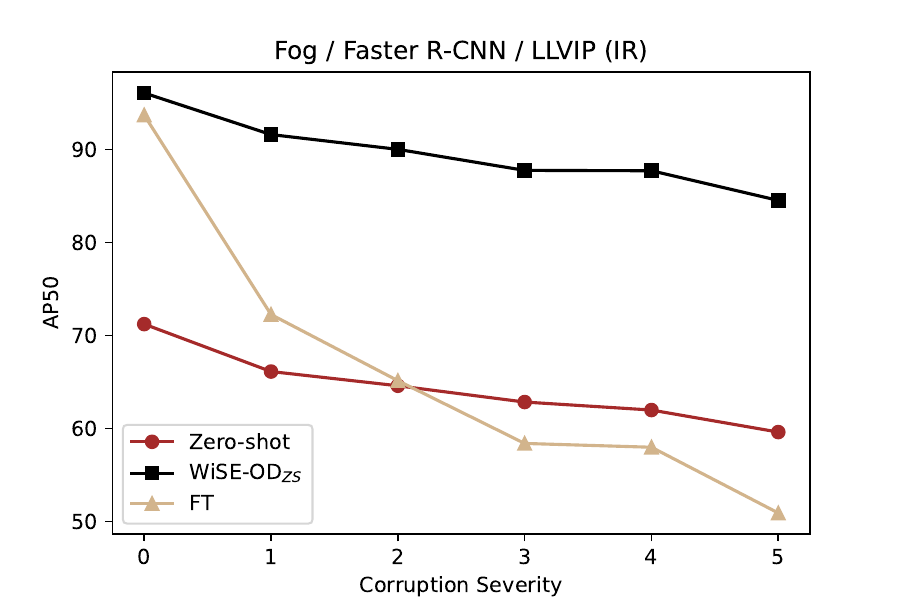}
\end{subfigure}
\begin{subfigure}[t]{0.60\columnwidth}
    
    \includegraphics[width=\columnwidth]{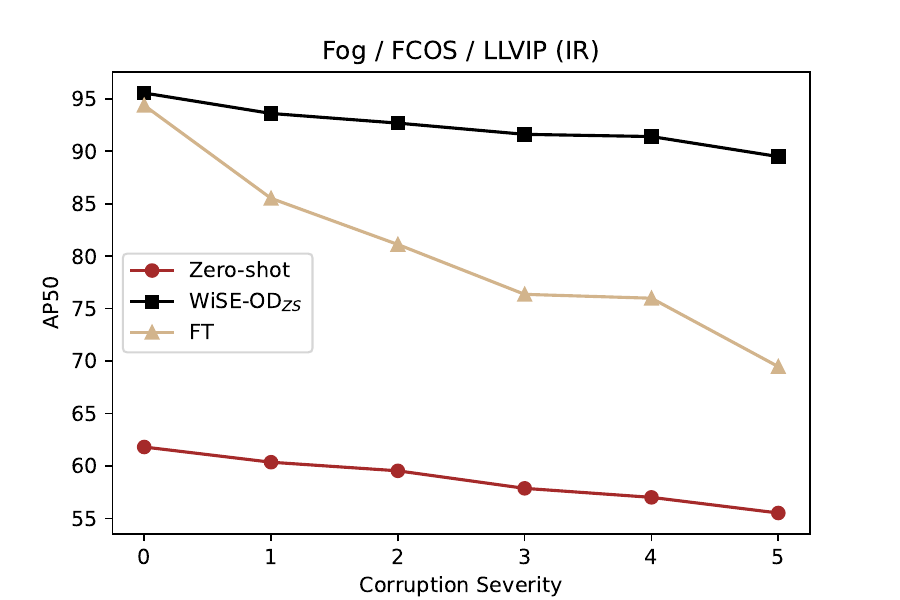}
\end{subfigure}
\begin{subfigure}[t]{0.60\columnwidth}
    
    \includegraphics[width=\columnwidth]{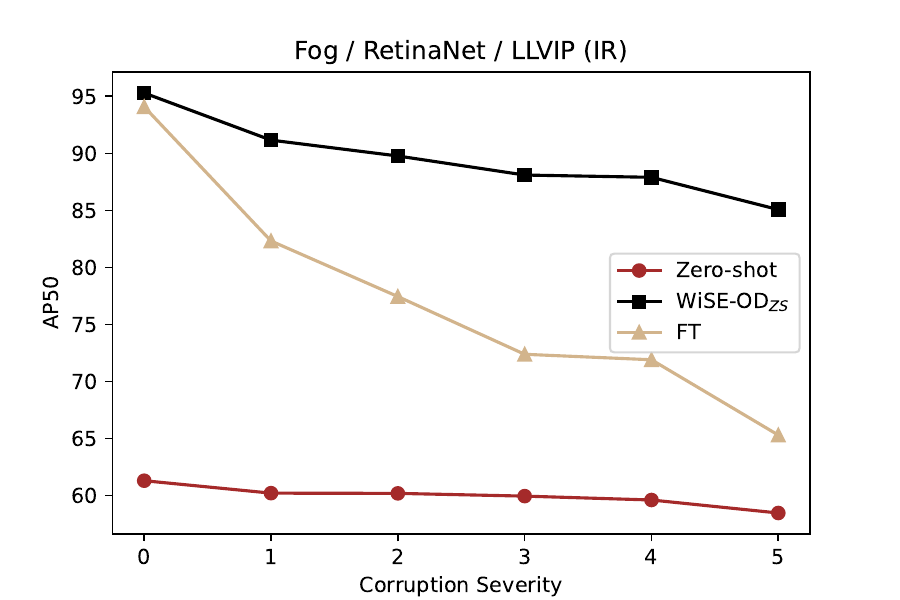}
\end{subfigure}


\begin{subfigure}[t]{0.60\columnwidth}
    
    \includegraphics[width=\columnwidth]{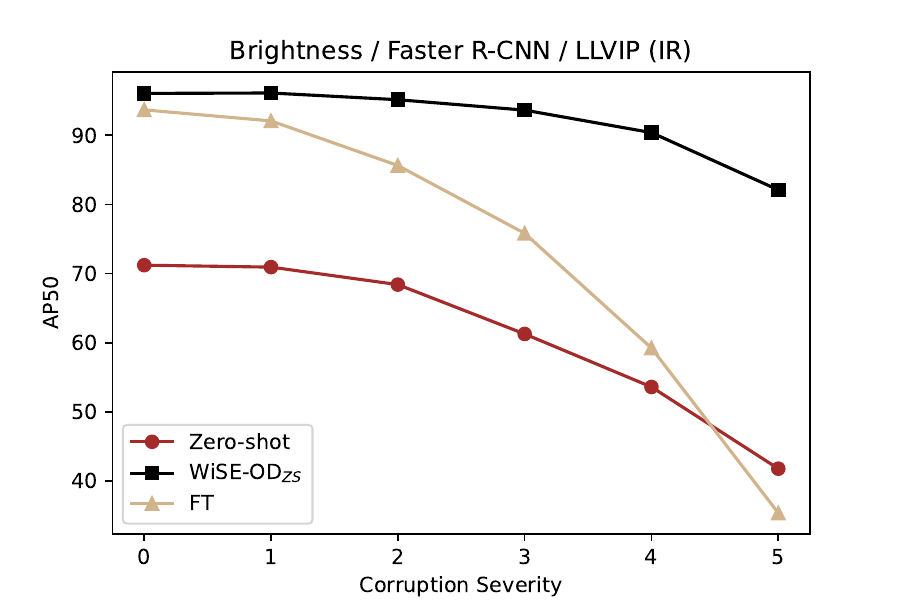}
\end{subfigure}
\begin{subfigure}[t]{0.60\columnwidth}
    
    \includegraphics[width=\columnwidth]{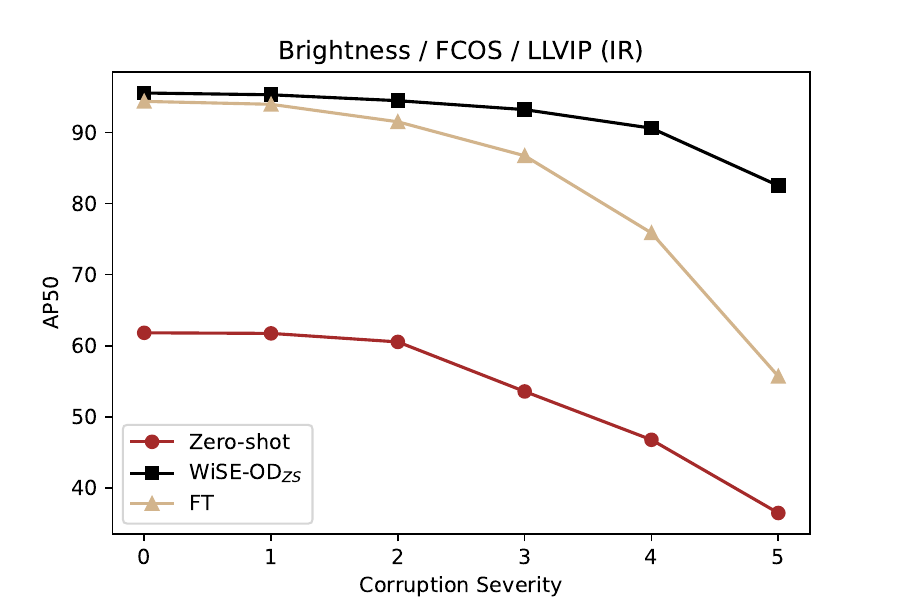}
\end{subfigure}
\begin{subfigure}[t]{0.60\columnwidth}
    
    \includegraphics[width=\columnwidth]{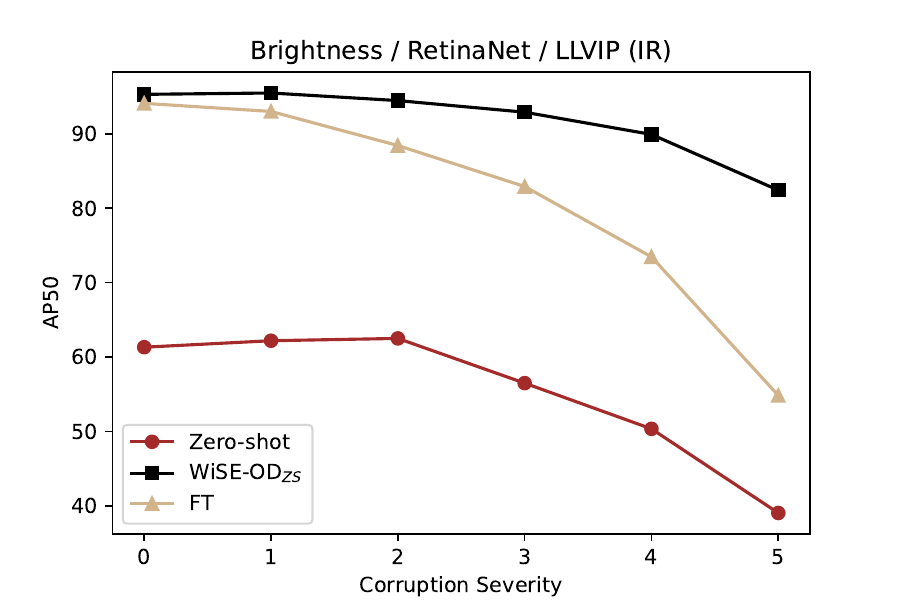}
\end{subfigure}

\begin{subfigure}[t]{0.60\columnwidth}
    
    \includegraphics[width=\columnwidth]{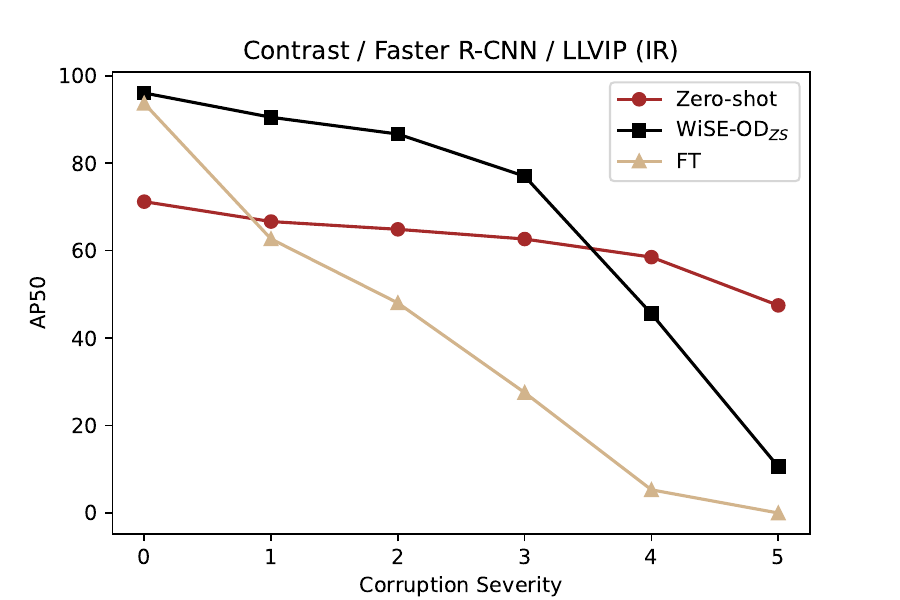}
\end{subfigure}
\begin{subfigure}[t]{0.60\columnwidth}
    
    \includegraphics[width=\columnwidth]{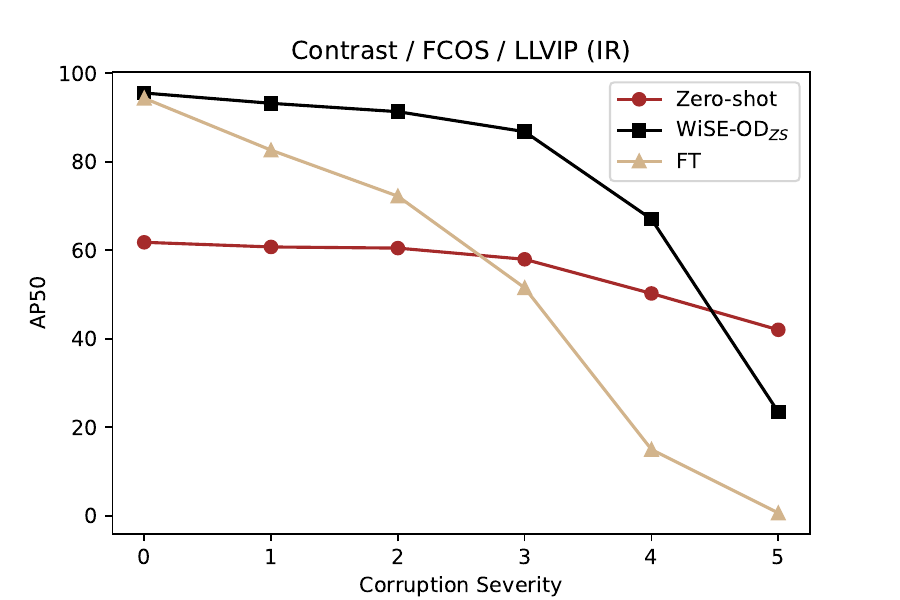}
\end{subfigure}
\begin{subfigure}[t]{0.60\columnwidth}
    
    \includegraphics[width=\columnwidth]{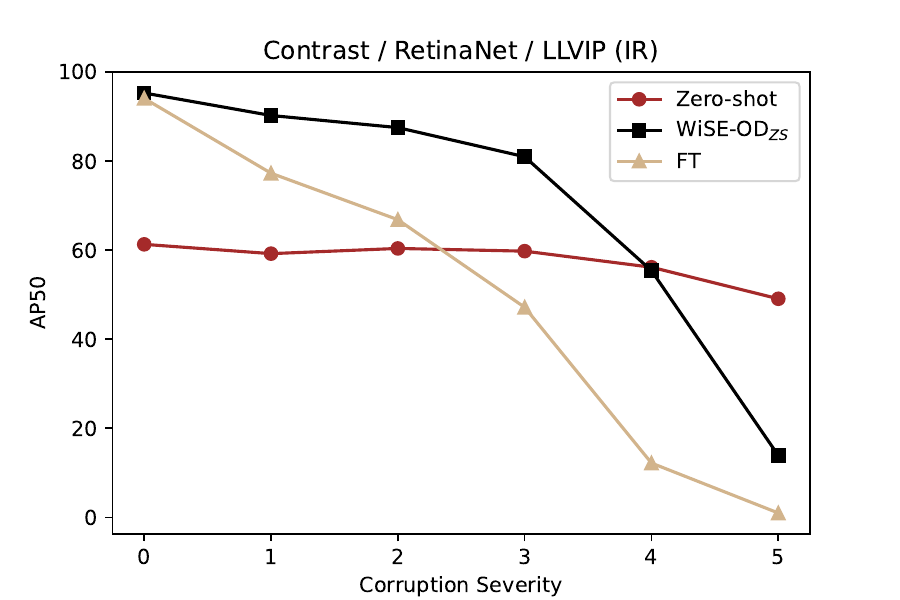}
\end{subfigure}

\begin{subfigure}[t]{0.60\columnwidth}
    
    \includegraphics[width=\columnwidth]{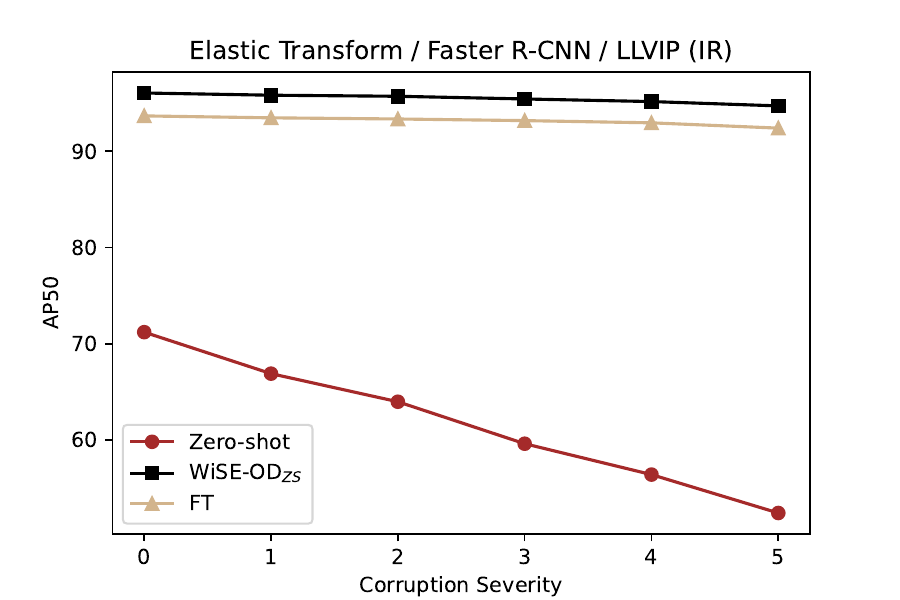}
\end{subfigure}
\begin{subfigure}[t]{0.60\columnwidth}
    
    \includegraphics[width=\columnwidth]{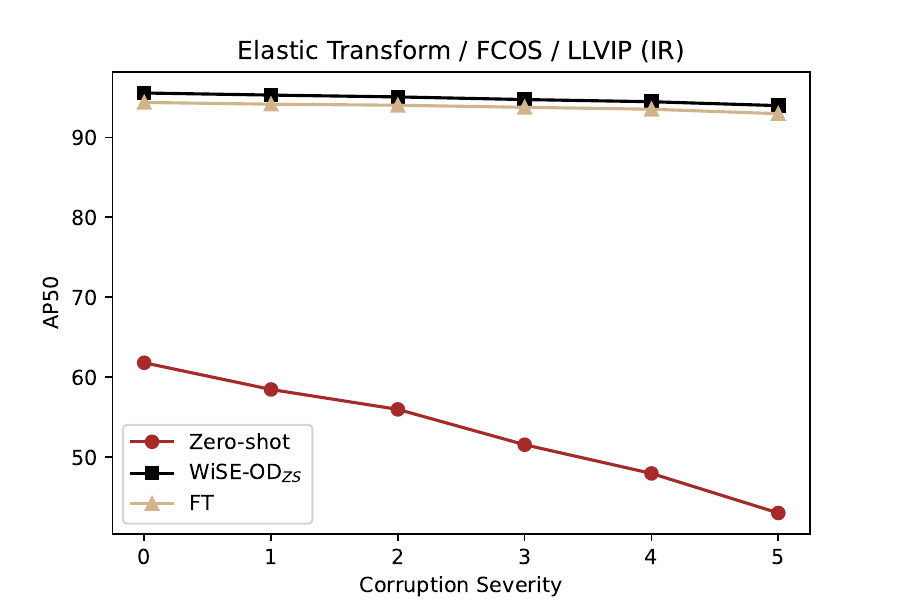}
\end{subfigure}
\begin{subfigure}[t]{0.60\columnwidth}
    
    \includegraphics[width=\columnwidth]{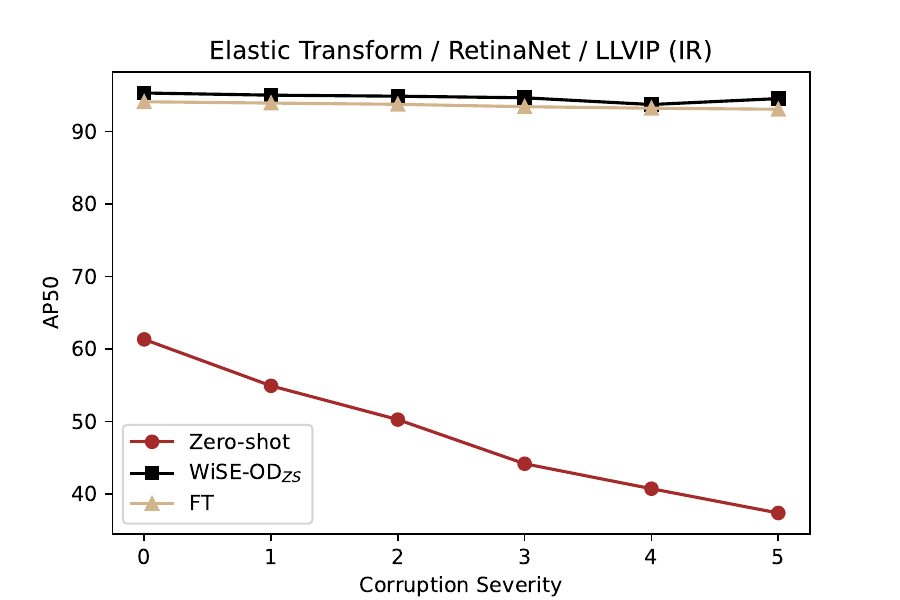}
\end{subfigure}

\caption{\textbf{AP$_{50}$ performance for all detectors over different corruption severity levels for Snow, Frost, Fog, Brightness, Contrast, Elastic Transform.} For each perturbation, we evaluated different levels of corruption for the Zero-Shot, WiSE-OD$_{ZS}$, and FT models for LLVIP-C.}
\label{fig:plots_llvip_part2}

\end{figure*}

\begin{figure*}
\captionsetup[subfigure]{labelformat=empty}
\centering

\begin{subfigure}[t]{0.60\columnwidth}
    
    \includegraphics[width=\columnwidth]{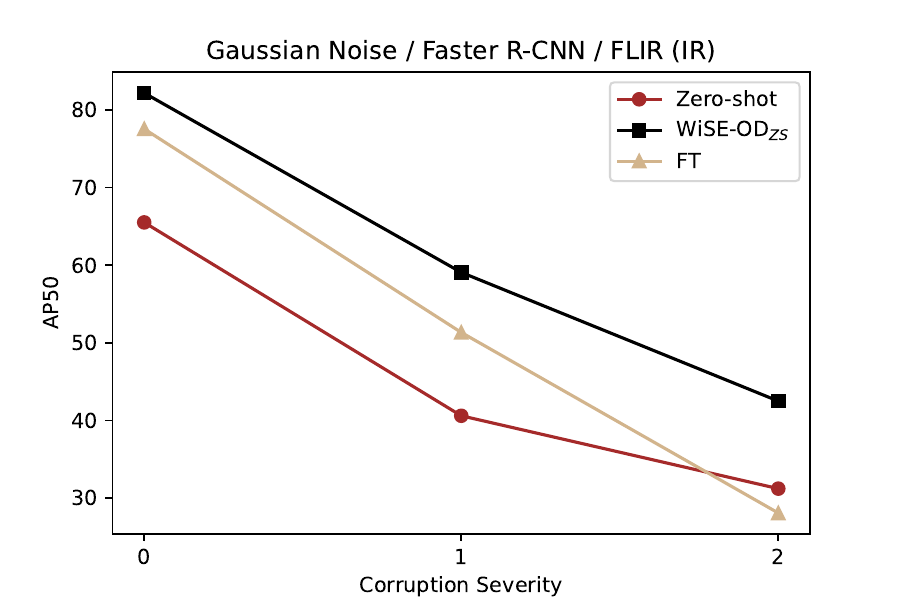}
\end{subfigure}
\begin{subfigure}[t]{0.60\columnwidth}
    
    \includegraphics[width=\columnwidth]{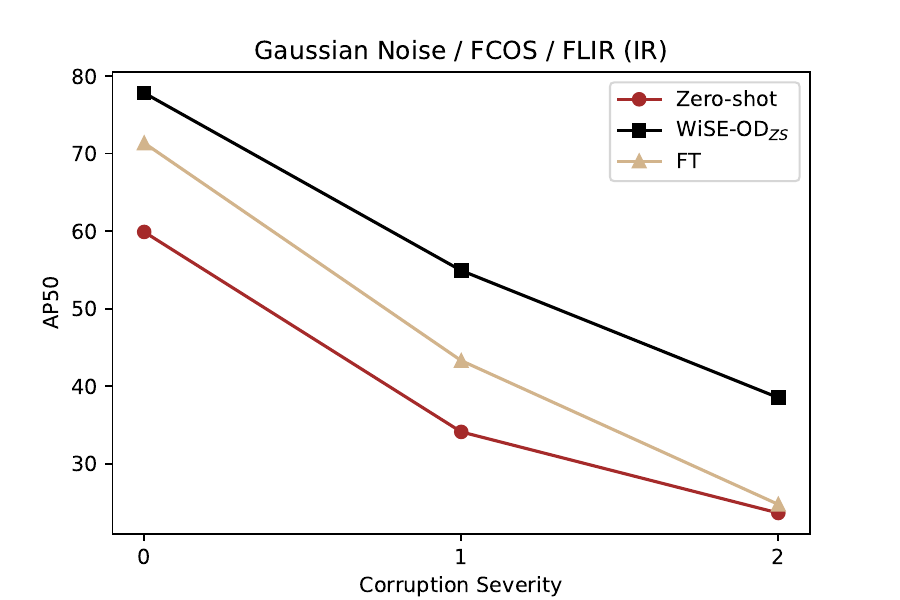}
\end{subfigure}
\begin{subfigure}[t]{0.60\columnwidth}
    
    \includegraphics[width=\columnwidth]{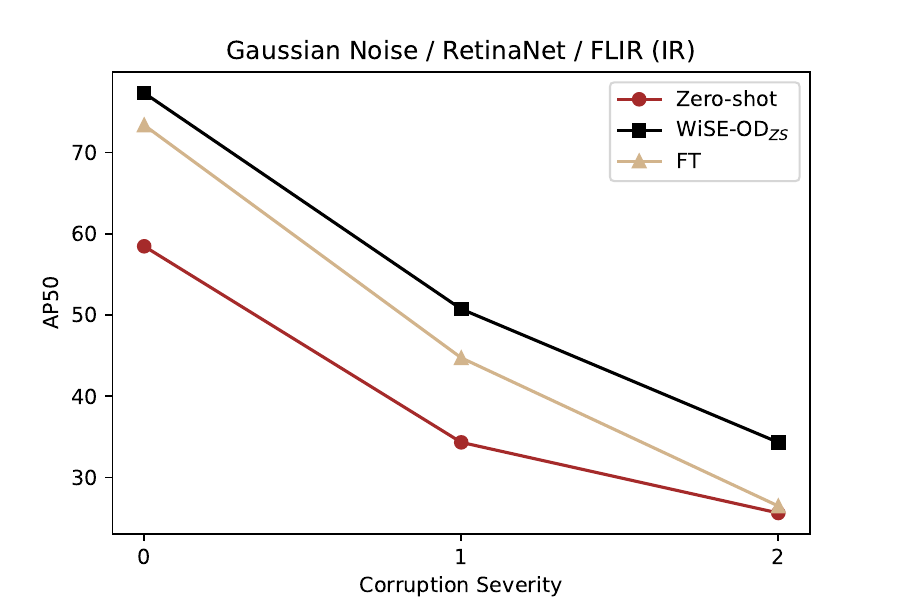}
\end{subfigure}

\begin{subfigure}[t]{0.60\columnwidth}
    
    \includegraphics[width=\columnwidth]{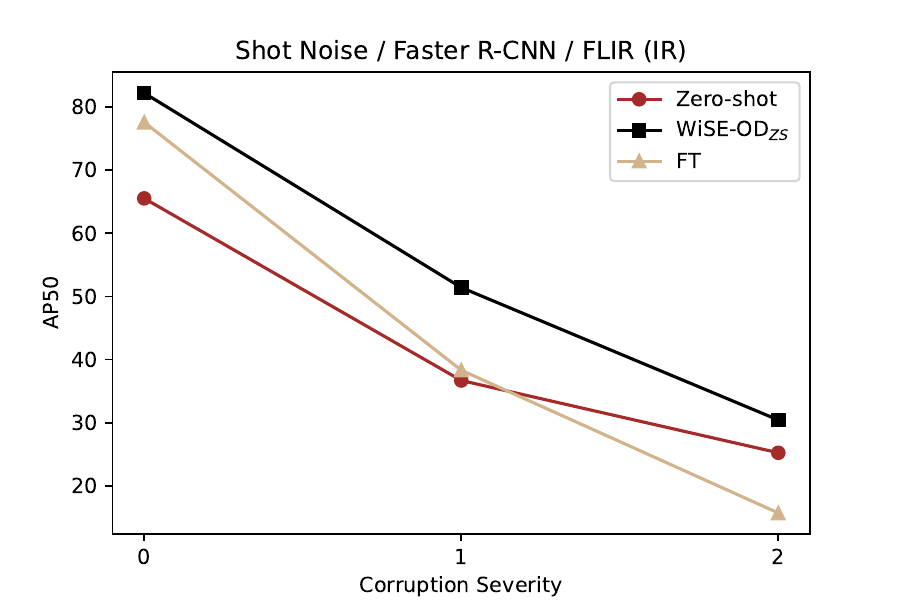}
\end{subfigure}
\begin{subfigure}[t]{0.60\columnwidth}
    
    \includegraphics[width=\columnwidth]{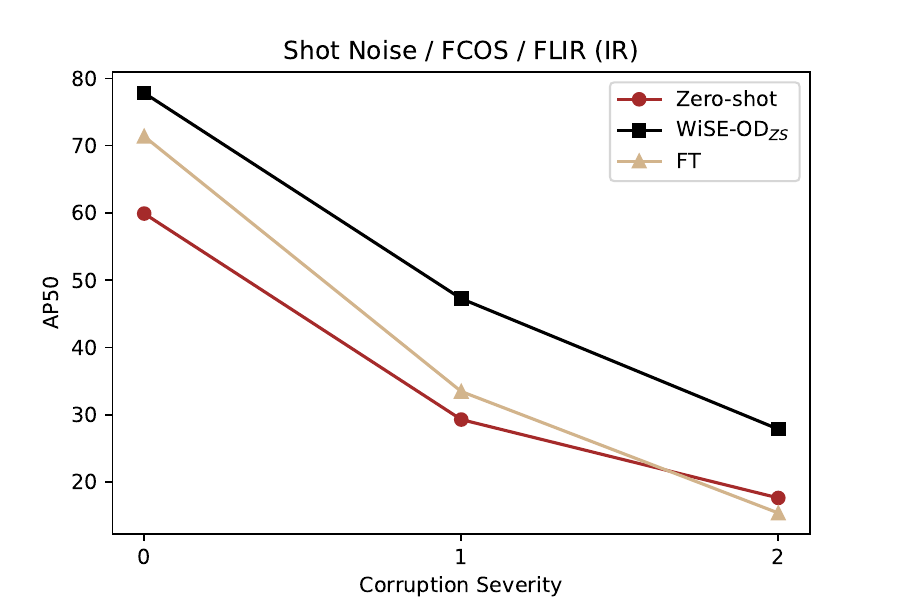}
\end{subfigure}
\begin{subfigure}[t]{0.60\columnwidth}
    
    \includegraphics[width=\columnwidth]{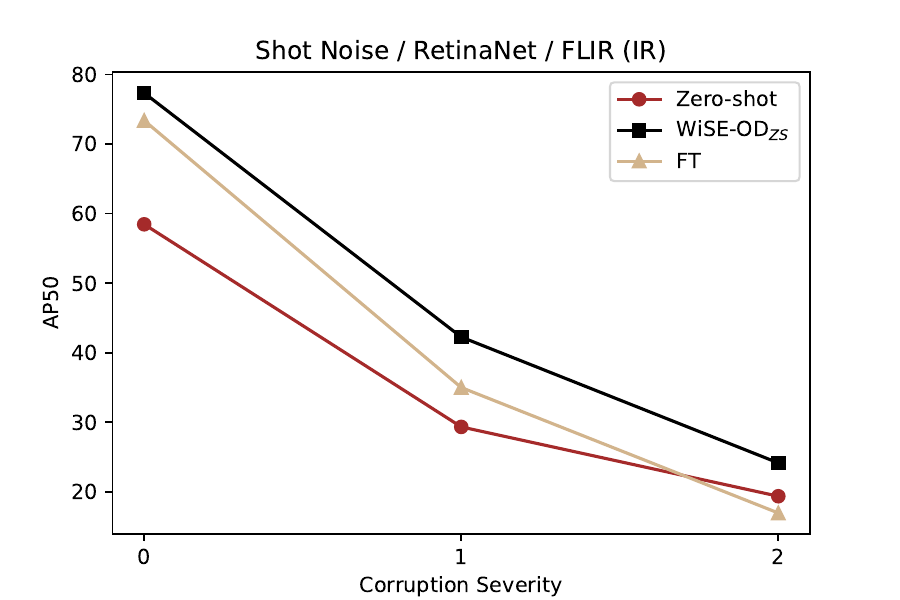}
\end{subfigure}

\begin{subfigure}[t]{0.60\columnwidth}
    
    \includegraphics[width=\columnwidth]{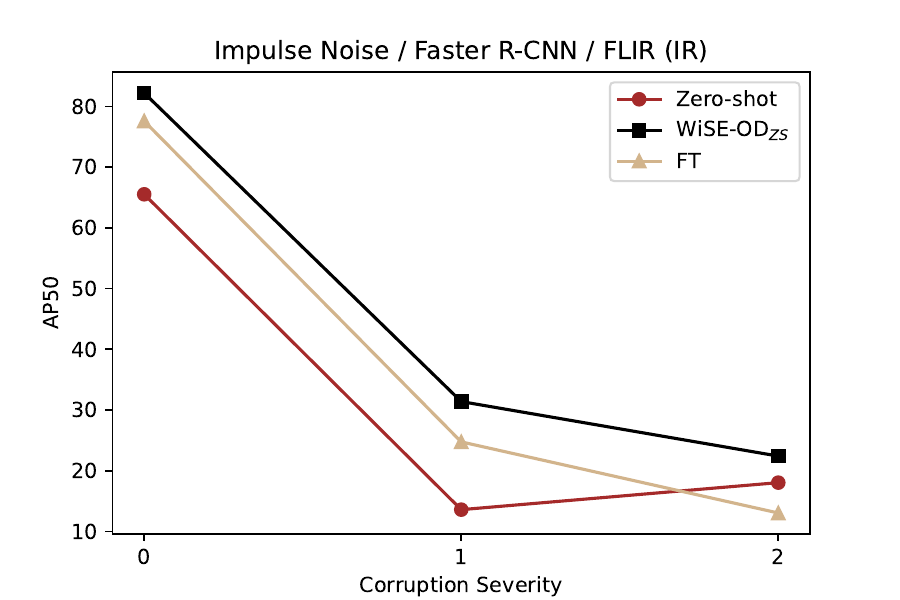}
\end{subfigure}
\begin{subfigure}[t]{0.60\columnwidth}
    
    \includegraphics[width=\columnwidth]{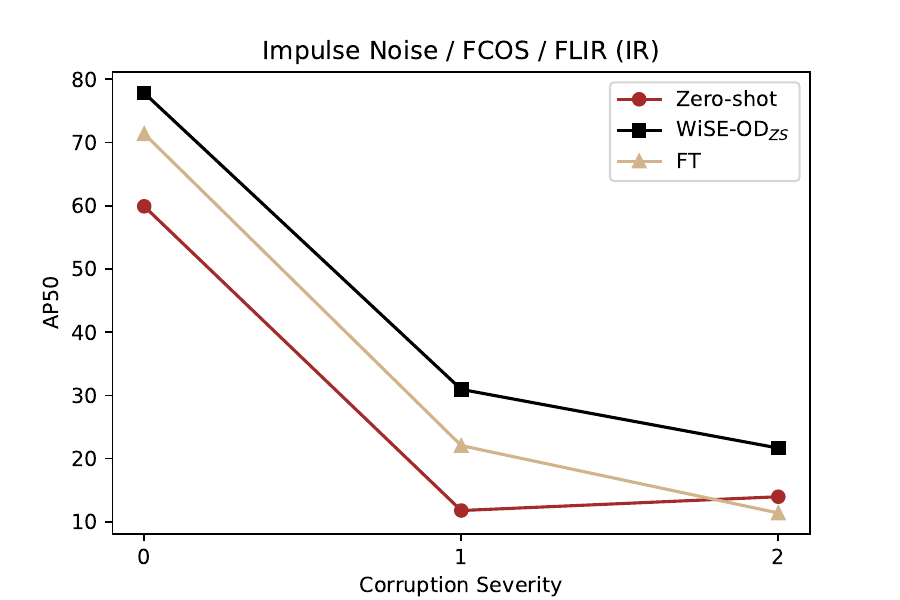}
\end{subfigure}
\begin{subfigure}[t]{0.60\columnwidth}
    
    \includegraphics[width=\columnwidth]{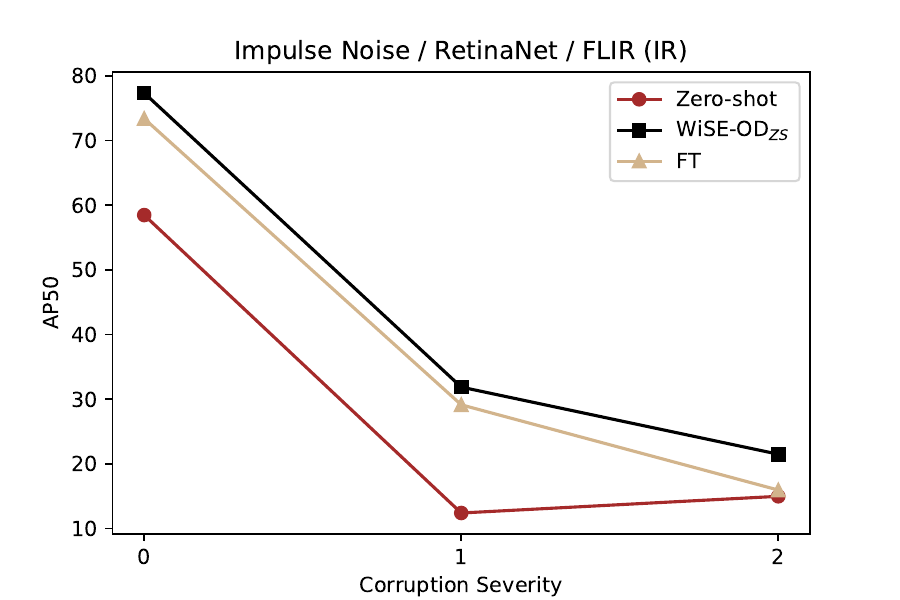}
\end{subfigure}

\begin{subfigure}[t]{0.60\columnwidth}
    
    \includegraphics[width=\columnwidth]{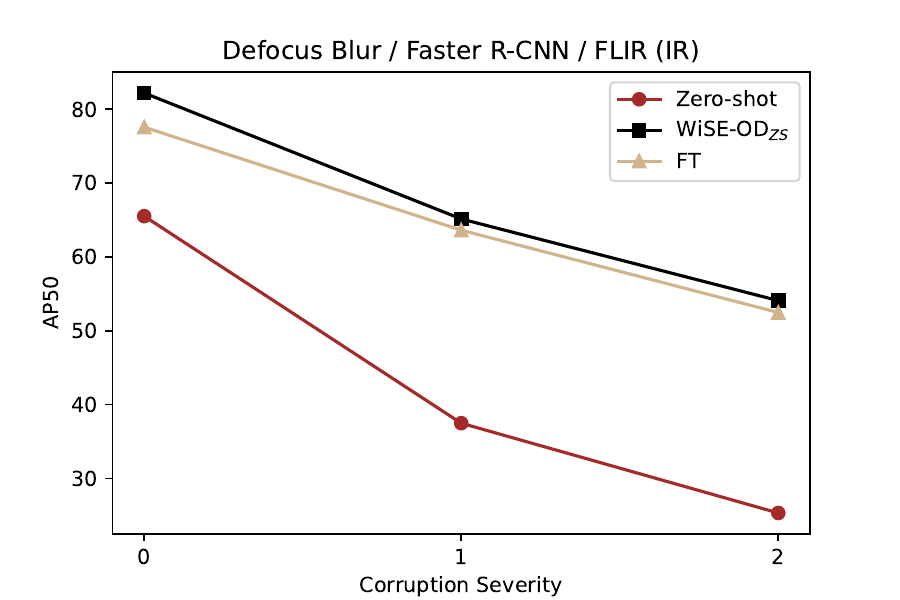}
\end{subfigure}
\begin{subfigure}[t]{0.60\columnwidth}
    
    \includegraphics[width=\columnwidth]{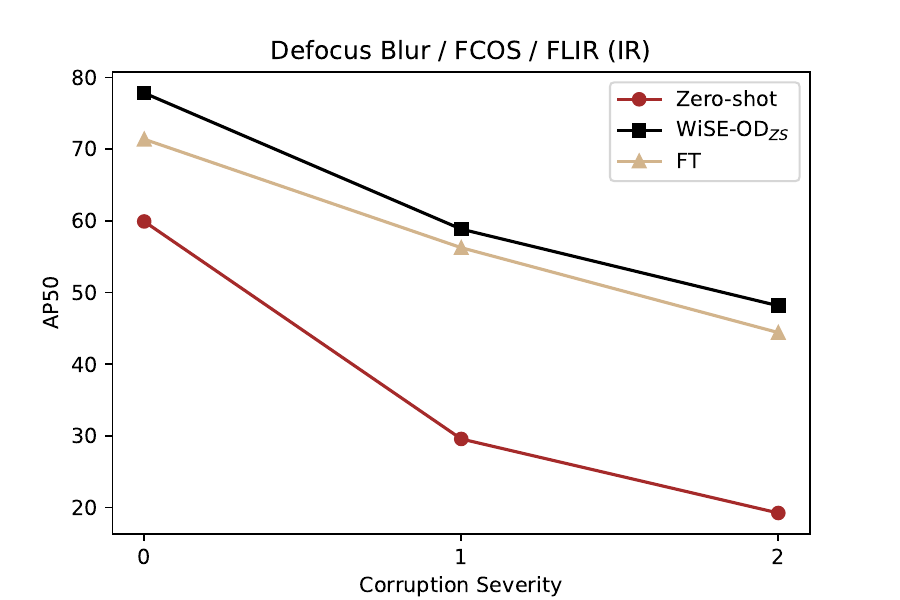}
\end{subfigure}
\begin{subfigure}[t]{0.60\columnwidth}
    
    \includegraphics[width=\columnwidth]{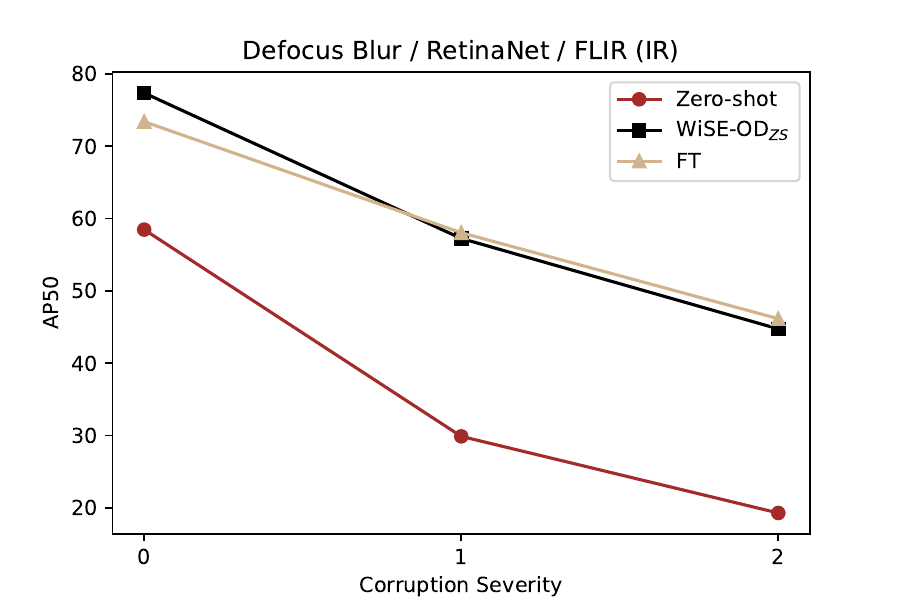}
\end{subfigure}

\begin{subfigure}[t]{0.60\columnwidth}
    
    \includegraphics[width=\columnwidth]{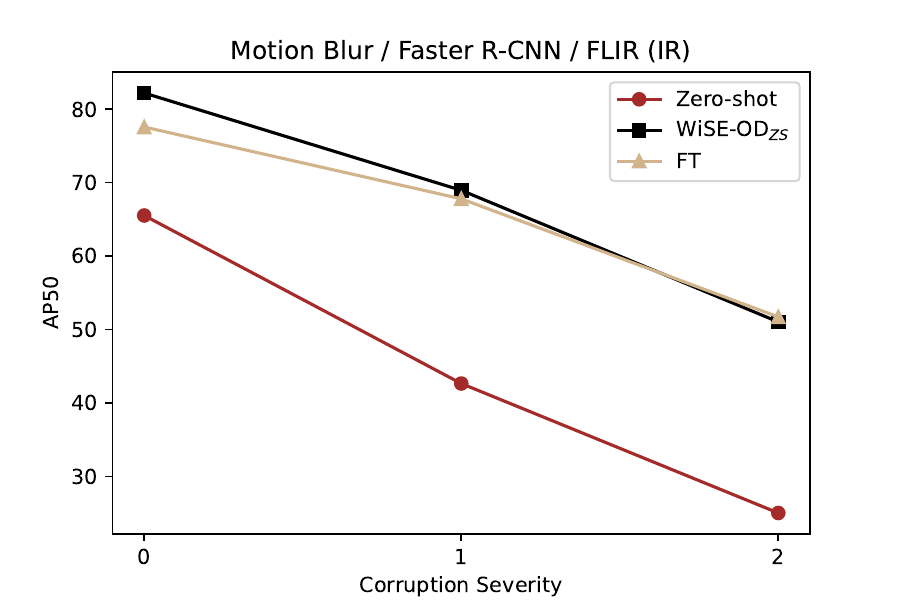}
\end{subfigure}
\begin{subfigure}[t]{0.60\columnwidth}
    
    \includegraphics[width=\columnwidth]{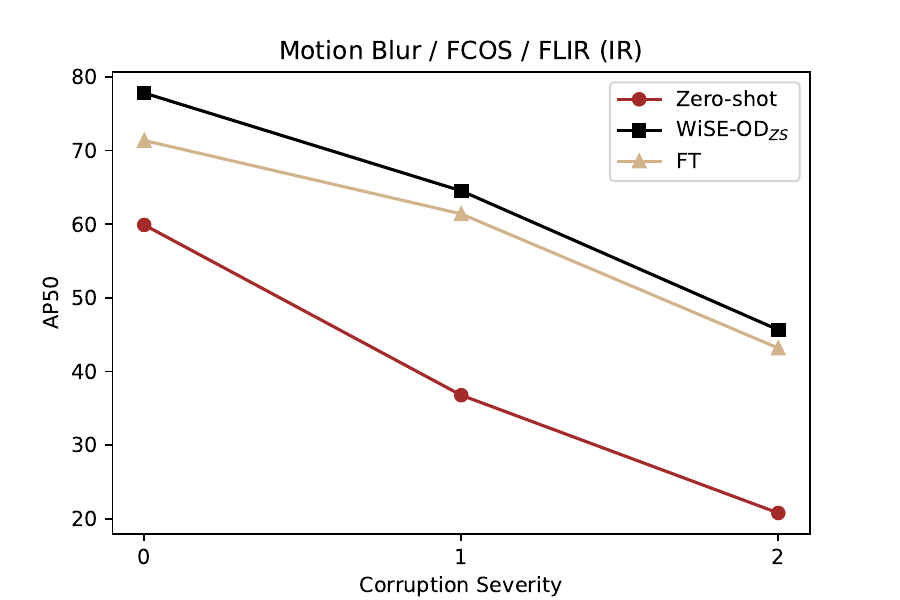}
\end{subfigure}
\begin{subfigure}[t]{0.60\columnwidth}
    
    \includegraphics[width=\columnwidth]{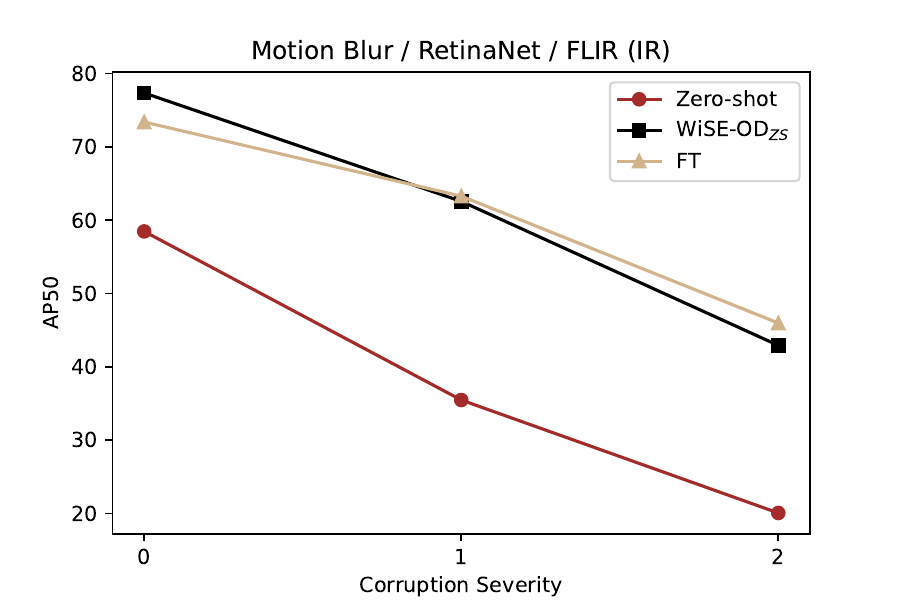}
\end{subfigure}

\caption{\textbf{AP$_{50}$ performance for all detectors over different corruption severity levels for Gaussian Noise, Shot Noise, Impulse Noise, Defocus Blur, Motion Blur.} For each perturbation, we evaluated different levels of corruption for the Zero-Shot, WiSE-OD$_{ZS}$, and FT models for FLIR-C.}
\label{fig:plots_flir_part1}

\end{figure*}

\begin{figure*}
\captionsetup[subfigure]{labelformat=empty}
\centering

\begin{subfigure}[t]{0.60\columnwidth}
    
    \includegraphics[width=\columnwidth]{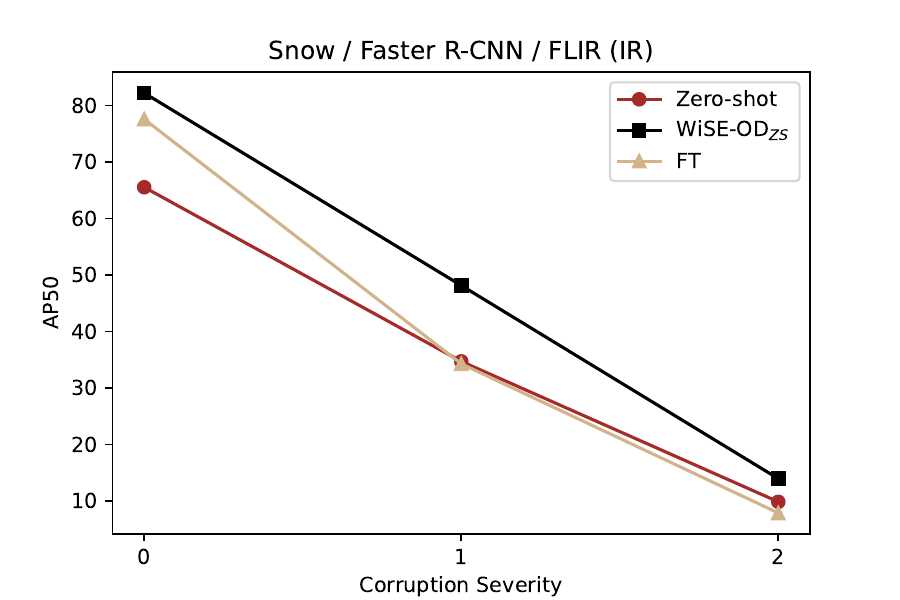}
\end{subfigure}
\begin{subfigure}[t]{0.60\columnwidth}
    
    \includegraphics[width=\columnwidth]{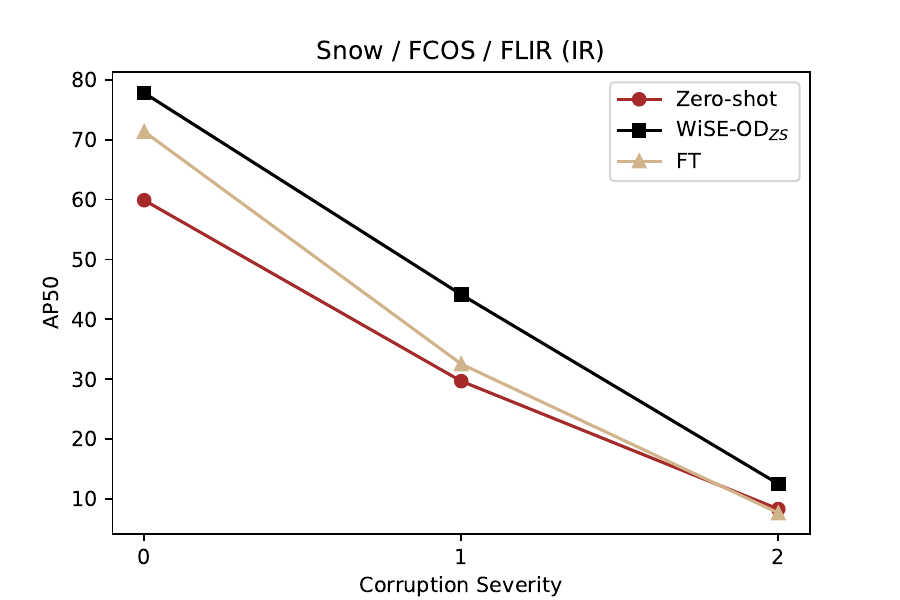}
\end{subfigure}
\begin{subfigure}[t]{0.60\columnwidth}
    
    \includegraphics[width=\columnwidth]{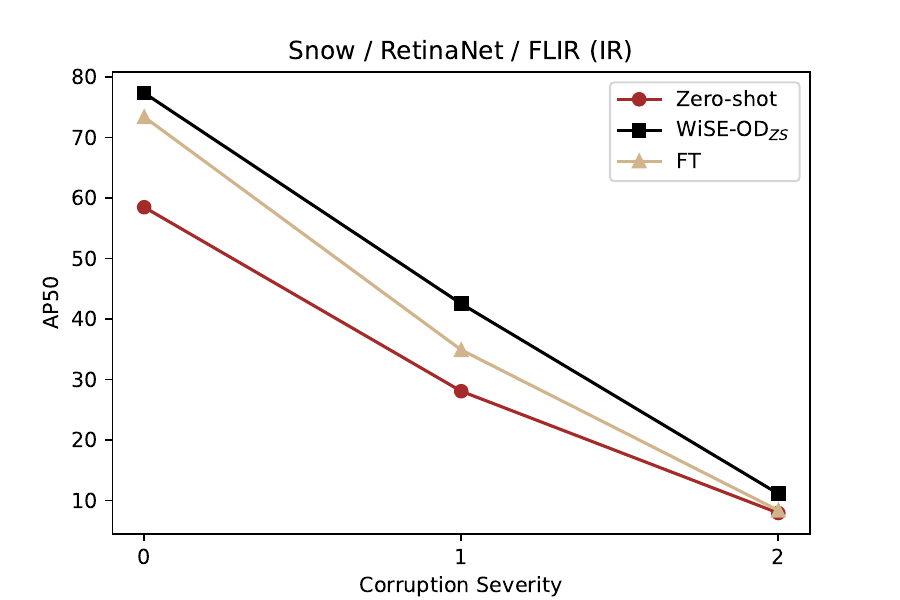}
\end{subfigure}

\begin{subfigure}[t]{0.60\columnwidth}
    
    \includegraphics[width=\columnwidth]{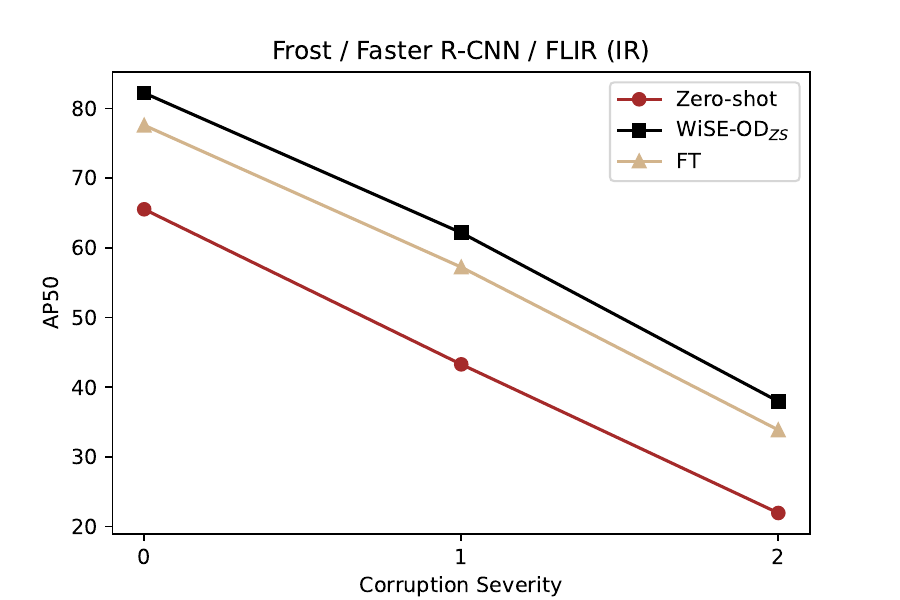}
\end{subfigure}
\begin{subfigure}[t]{0.60\columnwidth}
    
    \includegraphics[width=\columnwidth]{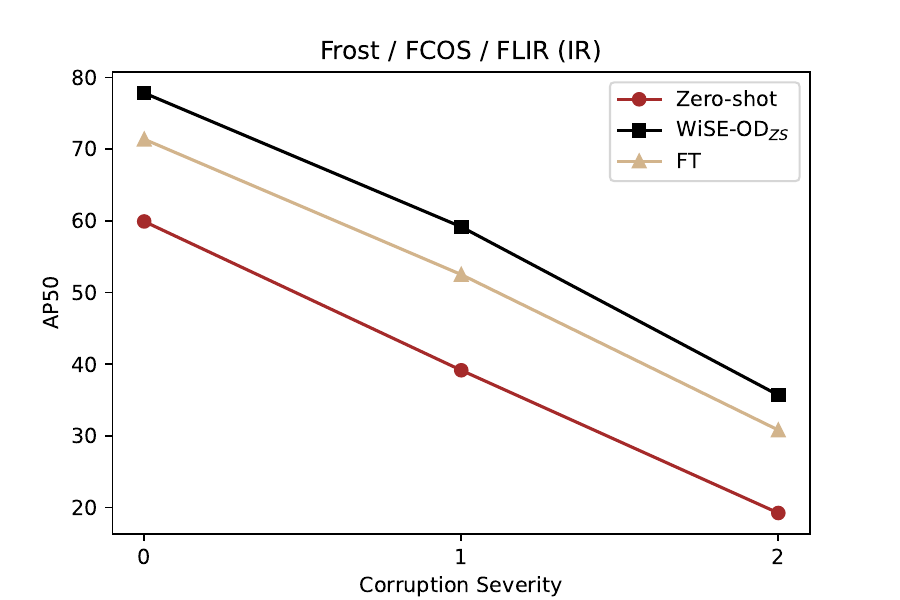}
\end{subfigure}
\begin{subfigure}[t]{0.60\columnwidth}
    
    \includegraphics[width=\columnwidth]{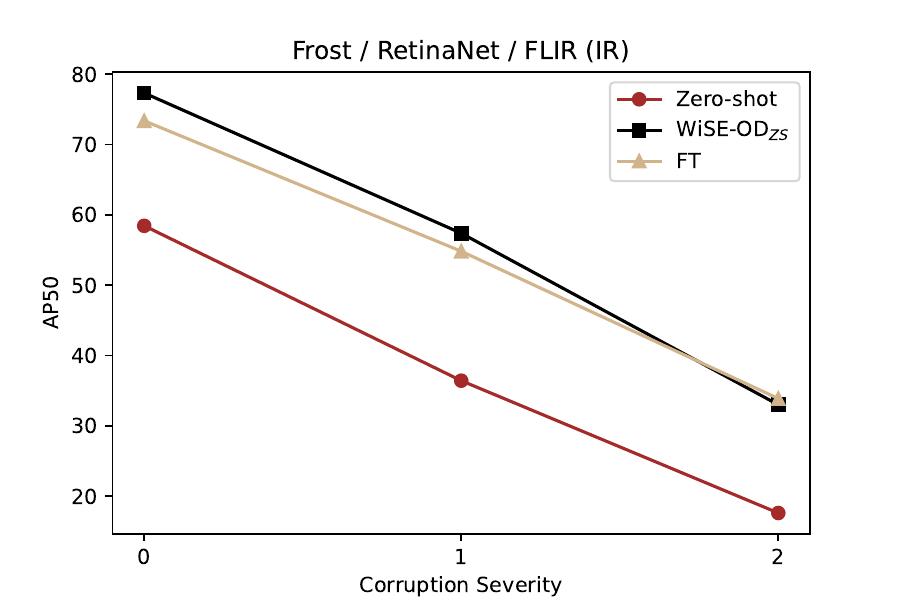}
\end{subfigure}

\begin{subfigure}[t]{0.60\columnwidth}
    
    \includegraphics[width=\columnwidth]{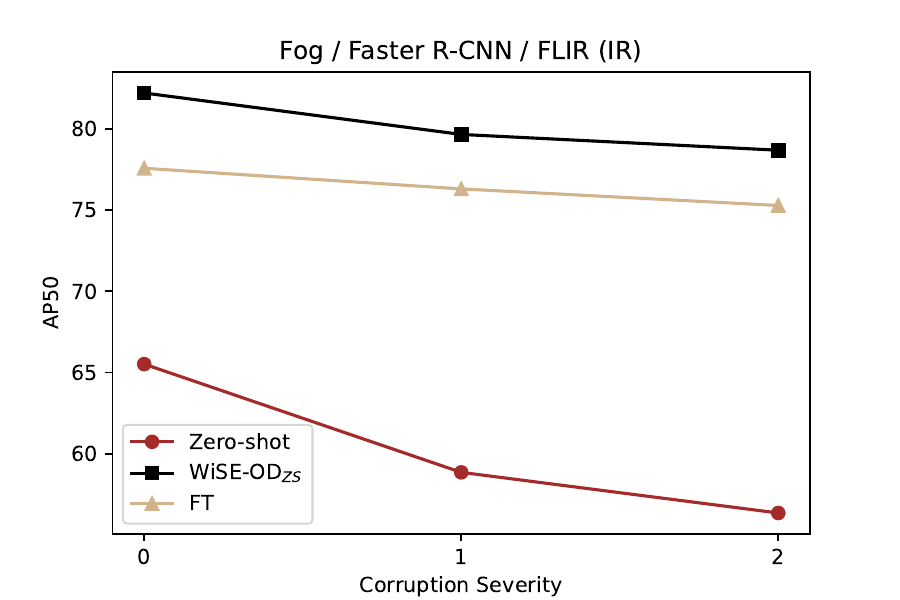}
\end{subfigure}
\begin{subfigure}[t]{0.60\columnwidth}
    
    \includegraphics[width=\columnwidth]{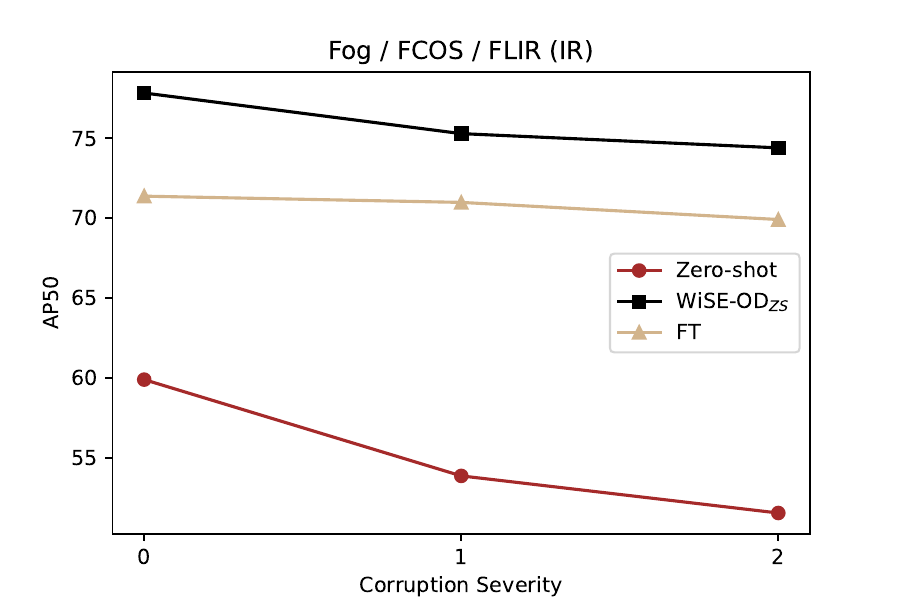}
\end{subfigure}
\begin{subfigure}[t]{0.60\columnwidth}
    
    \includegraphics[width=\columnwidth]{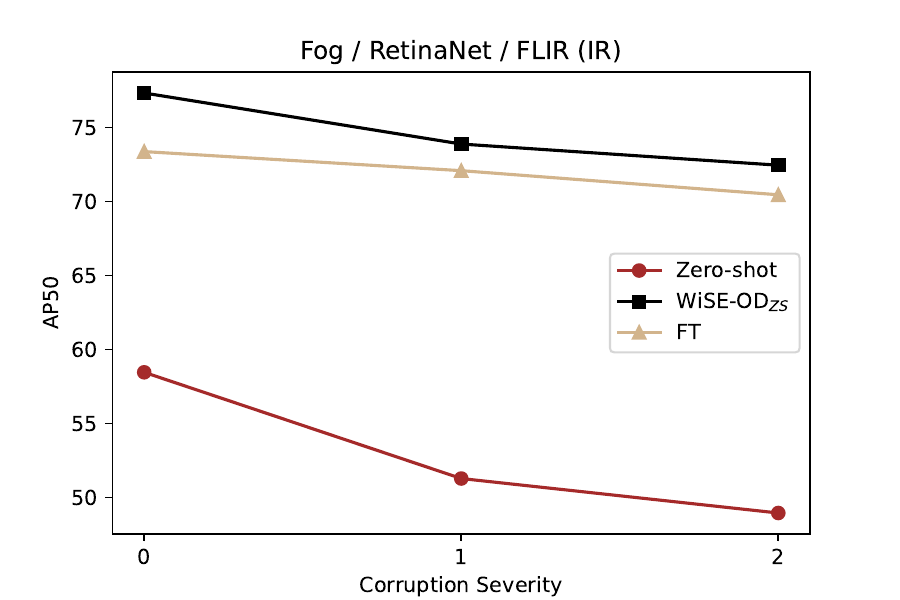}
\end{subfigure}

\caption{\textbf{AP$_{50}$ performance for all detectors over different corruption severity levels for Snow, Frost and Fog.} For each perturbation, we evaluated different levels of corruption for the Zero-Shot, WiSE-OD$_{ZS}$, and FT models for FLIR-C.}
\label{fig:plots_flir_part2}

\end{figure*}

In this section, we measured the per AP$_{50}$ performance for Faster R-CNN, FCOS, and RetinaNet over different corruption severity levels for the benchmark. Here, we focus on Zero-Shot, FT and WiSE-OD$_{ZS}$ for most of the corruptions on LLVIP-C illustrated in~\fref{fig:plots_llvip_part1} and~\fref{fig:plots_llvip_part2}, and for FLIR-C in~\fref{fig:plots_flir_part1} and~\fref{fig:plots_flir_part2}.


\clearpage

{
    \small
    \bibliographystyle{ieeenat_fullname}
    \bibliography{main}
}

\end{document}